\documentclass[lettersize,journal]{IEEEtran}
\usepackage{amsmath}
\usepackage{amsfonts}
\usepackage{algorithmic}
\usepackage{algorithm}
\usepackage{array}
\usepackage{textcomp}
\usepackage{stfloats}
\usepackage{url}
\usepackage{verbatim}
\usepackage{graphicx}
\usepackage{cite}
\usepackage{hyperref}
\usepackage{graphbox}
\usepackage{multirow}
\usepackage{makecell}
\usepackage{booktabs}
\usepackage{bm}
\usepackage{caption}
\usepackage{subcaption}
\usepackage{placeins}
\usepackage{amsthm}
\usepackage{amssymb}
\usepackage{enumitem}

\hypersetup{hidelinks}

\newtheorem{definition}{\textup{Definition}}

\newcommand{\mymodelend}{$\text{Dis-H}^2\text{GCN}$}
\newcommand{\mymodel}{$\text{Dis-H}^2\text{GCN}$~}

\begin{document}

\title{Disentangled Hyperbolic Representation Learning \\ for Heterogeneous Graphs}

\author{Qijie Bai, Changli Nie, Haiwei Zhang, Zhicheng Dou~\IEEEmembership{Member,~IEEE,}, Xiaojie Yuan
        % <-this % stops a space
\thanks{Q. Bai, C. Nie, H. Zhang and X. Yuan are with the College of Computer Science, Tianjin Key Laboratory of Network and Data Security Technology, Nankai University, Tianjin, China. E-mail: \{qijie.bai, nie\_cl\}@mail.nankai.edu.cn, \{zhhaiwei, yuanxj\}@nankai.edu.cn.}% <-this % stops a space
\thanks{Z. Dou is with Gaoling School of Artificial Intelligence, Renmin University of China, Beijing, China. E-mail: dou@ruc.edu.cn.}
}

% {Bai \MakeLowercase{\textit{et al.}}: Disentangled Hyperbolic Representation Learning for Heterogeneous Graphs}


\maketitle

\begin{abstract}
Heterogeneous graphs have attracted a lot of research interests recently due to the success for representing complex real-world systems. 
However, existing methods have two pain points in embedding them into low-dimensional spaces: the mixing of structural and semantic information, and the distributional mismatch between data and embedding spaces. 
These two challenges require representation methods to consider the global and partial data distributions while unmixing the information. 
Therefore, in this paper, we propose \mymodelend, a Disentangled Hyperbolic Heterogeneous Graph Convolutional Network. 
On the one hand, we leverage the mutual information minimization and discrimination maximization constraints to disentangle the semantic features from comprehensively learned representations by independent message propagation for each edge type, away from the pure structural features. 
On the other hand, the entire model is constructed upon the hyperbolic geometry to narrow the gap between data distributions and representing spaces. 
We evaluate our proposed \mymodel on five real-world heterogeneous graph datasets across two downstream tasks: node classification and link prediction. 
The results demonstrate its superiority over state-of-the-art methods, showcasing the effectiveness of our method in disentangling and representing heterogeneous graph data in hyperbolic spaces.
\end{abstract}


\section{Introduction}  \label{sec::intro}

\IEEEPARstart{G}{raph} data, which can be abstracted from a lot of real-world systems (e.g. social networks~\cite{yuEnhancingSocialRecommendation2020}, academic networks~\cite{tangArnetMinerExtractionMining2008}, recommendation systems~\cite{xueMultiplexBipartiteNetwork2021} and molecular biology~\cite{huberGraphsMolecularBiology2007}), is widely used to model the complex relationships between entities.
Driven by the requirements of these realistic scenarios and the characteristics of ease for calculation, recently graph representation learning has attracted great attention as a general operation for graph data analysis, and has achieved outstanding performances on diverse downstream tasks, ranging from node clustering~\cite{boStructuralDeepClustering2020, jinAutomatedSelfsupervisedLearning2022}, node classification~\cite{wangAMGCNAdaptiveMultichannel2020}, link prediction~\cite{maJointMultilabelLearning2022} to community detection~\cite{kimDenselyConnectedUser2020}.
As a special subset with node/edge types, heterogeneous graphs maintain cooperative information on both graph structure and type semantics simultaneously~\cite{baiH2TNETemporalHeterogeneous2022}, and have become a research hotspot in the past years~\cite{changHeterogeneousNetworkEmbedding2015, shiSurveyHeterogeneousInformation2017, zhaoHeterogeneousGraphStructure2021}.

Existing heterogeneous graph representation learning methods can be categorized into two technical routes for capturing interactions between node pairs of different types.
On the one hand, sequential sampling strategies provide an efficient way to conversing non-structural graph data to structural sequences.
The sampled sequences can keep the topological and semantic correlations by adding extra type constraints, such as meta-path~\cite{sunPathSimMetaPathbased2011} and JUST~\cite{husseinAreMetaPathsNecessary2018}, and then general sequential modeling methods could be applied~\cite{dongMetapath2vecScalableRepresentation2017}.
On the other hand, with the rapid development of graph neural networks, researchers begin to consider using different aggregating functions or processes to directly capture structural and semantic information~\cite{schlichtkrullModelingRelationalData2018, zhangHeterogeneousGraphNeural2019}.
Some methods combine both of the above two categories by first utilizing meta-path to reconstruct the structural and semantic correlations and subsequently applying graph neural networks to learn their representations~\cite{fuMAGNNMetapathAggregated2020, zhaoHeterogeneousGraphStructure2021}.

\begin{figure}
    \centering
    \subfloat[Homogeneous graph.]{
        \hspace{5pt}
        \includegraphics[width=0.248\linewidth]{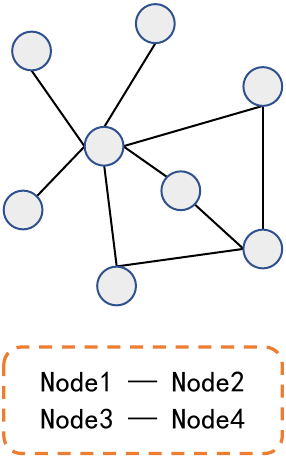}
        \hspace{5pt}
        \label{fig::intro:information:homogeneous}
    } \quad
    \subfloat[Heterogeneous graph.]{
        \hspace{5pt}
        \includegraphics[width=0.273\linewidth]{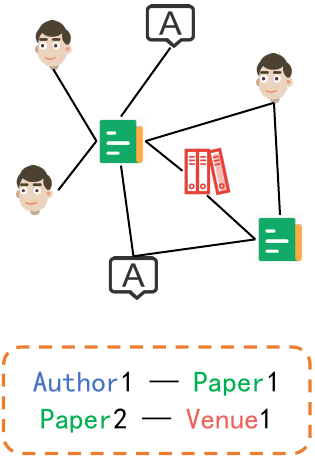}
        \hspace{5pt}
        \label{fig::intro:information:heterogeneous}
    }
    \caption{(a) Homogeneous graphs have only structural information which refers to the connections between node pairs. (b) Heterogeneous graphs have the fusion information of structure and semantics which refers to the types of nodes and edges.}
    \label{fig::intro:information}
    \vspace{-1em}
\end{figure}

However, all these methods fall into a paradigm that learns a mixed representation for each node, ignoring the distinguishing influences from different aspects of characteristics, e.g. structural and semantic.
Take Fig.~\ref{fig::intro:information} as an example, the homogeneous graph and heterogeneous graph have the same topological structure, and keep the same structural information ideally.
Meanwhile, the semantics of the heterogeneous graph tell us that the two nodes corresponding to node 2 and node 3 in the homogeneous graph are all of the type \textit{Paper}, and are supposed to keep more related in some contents than others.
Mixing the two aspects of information into one unified representation may cause a performance decrease in real-world scenarios due to the correlation of various downstream tasks with different information~\cite{cebiricSummarizingSemanticGraphs2019}.
Moreover, the mixed joint distribution is far more complex than two relatively simple marginal distributions, and significantly increases the difficulty of learning appropriate representations for heterogeneous graphs.
In the past few years, disentangling representation learning has drawn a great deal of interests~\cite {maDisentangledGraphConvolutional2019}.
It provides a common idea to learn data with mixed features of various aspects, e.g. different domains in cross-domain problems~\cite{guoDisentangledRepresentationsLearning2023}, time-invariant and time-varying information in temporal modeling~\cite{zhangDyTedDisentangledRepresentation2023}, etc.
Inspired by the success of these models on data with mixed information, in this paper, we attempt to disentangle the structural and semantic features in heterogeneous graphs, expecting to obtain better representations.

\begin{figure}
    \centering
    \subfloat[]{
        \includegraphics[width=0.28\linewidth]{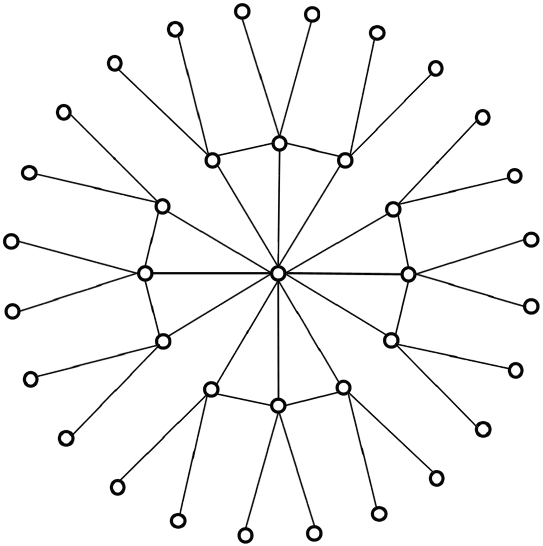}
        \label{fig::intro:distribution:hierarchy}
    }
    \subfloat[]{
        \includegraphics[width=0.28\linewidth]{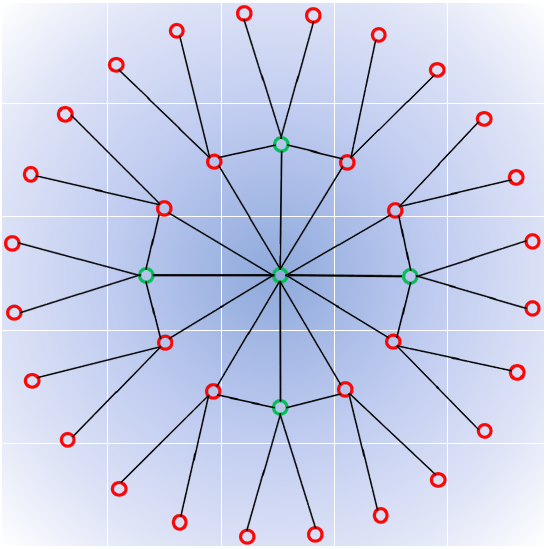}
        \label{fig::intro:distribution:hierarchy_euclidean}
    }
    \subfloat[]{
        \includegraphics[width=0.28\linewidth]{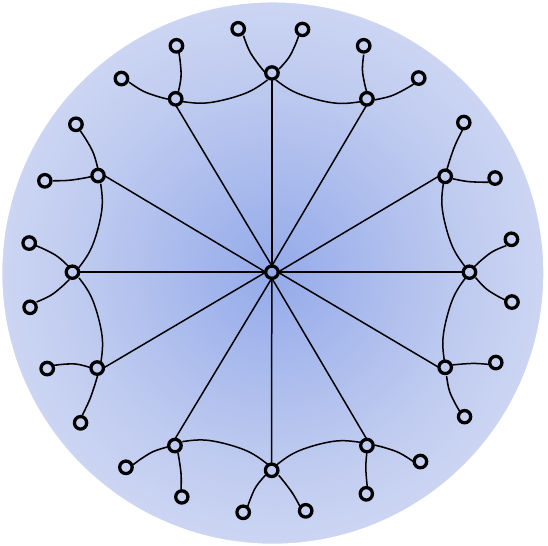}
        \label{fig::intro:distribution:hierarchy_hyperbolic}
    }
    \vspace{5pt}
    \\
    \subfloat[]{
        \includegraphics[width=0.45\linewidth]{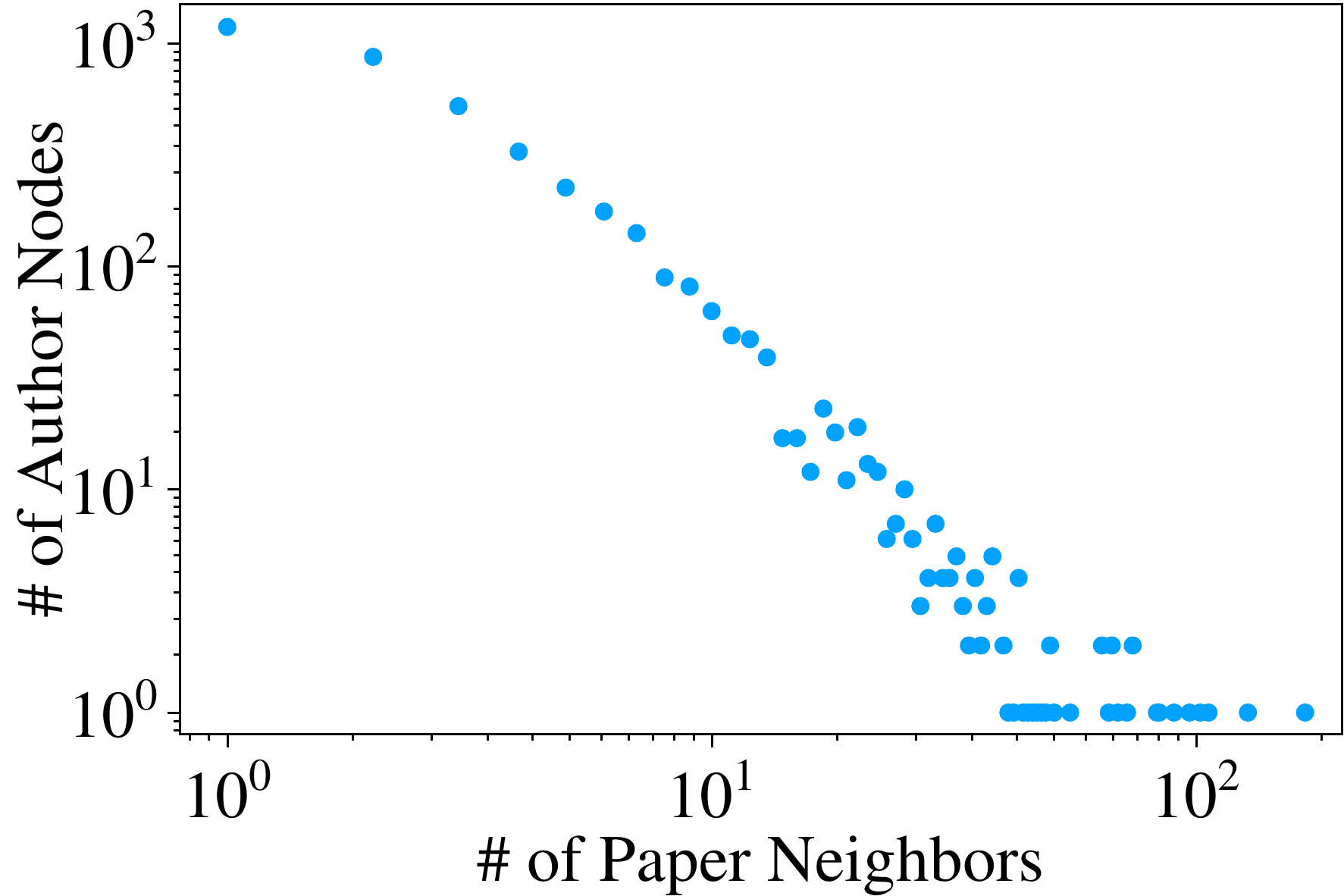}
        \label{fig::intro:distribution:distribution_ap}
    }
    \subfloat[]{
        \includegraphics[width=0.45\linewidth]{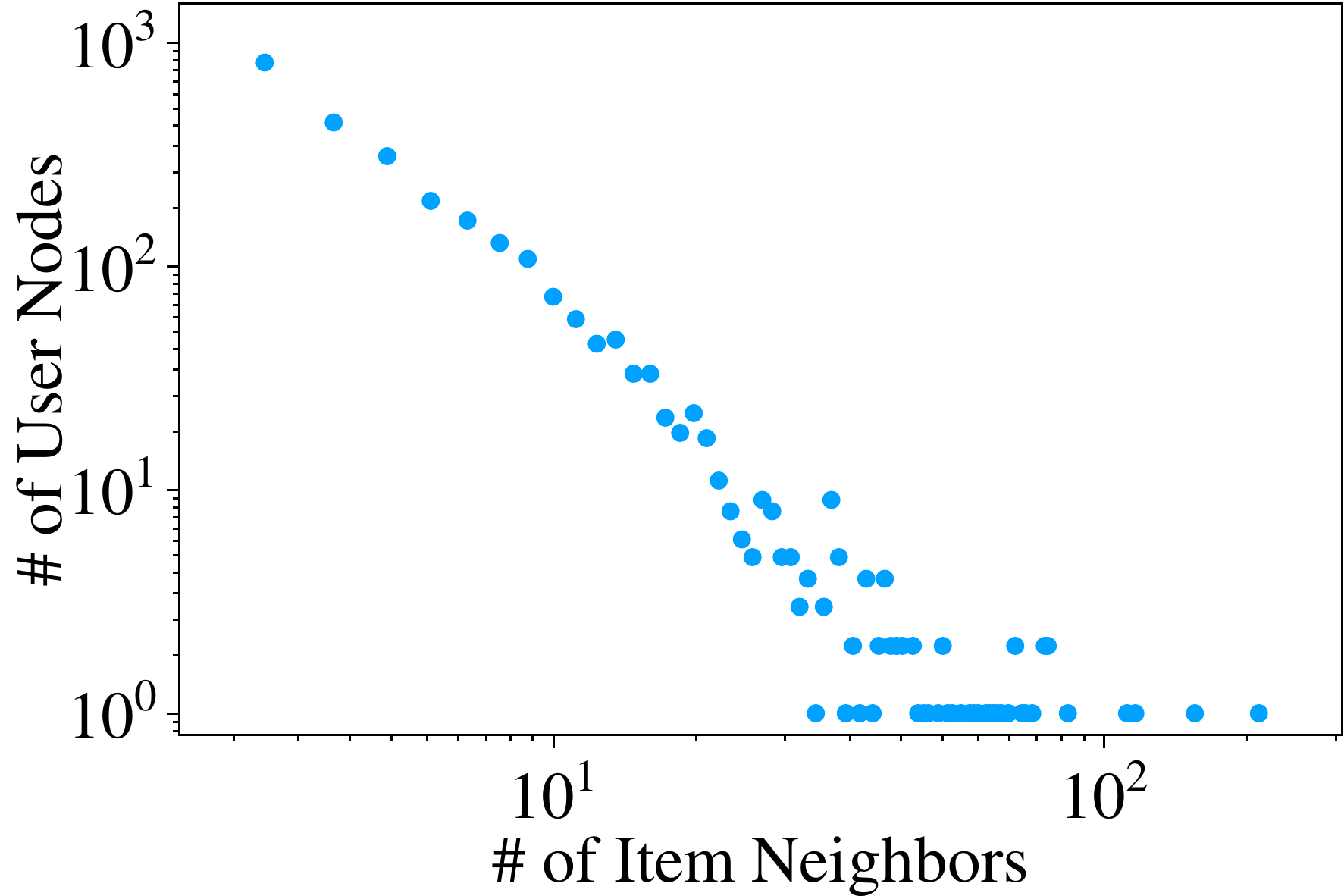}
        \label{fig::intro:distribution:distribution_ui}
    }
    \caption{(a) The hierarchical structure of graphs. (b) Embedding the graph into 2-D Euclidean space. The red nodes refer to the conflicts. (c) Embedding the graph into 2-D hyperbolic space (Poincaré disk model). The measures of distance between all node pairs connected with each other are equal. (d) The distribution of the relation \textit{Author} $\longleftrightarrow$ \textit{Paper} in dataset DBLP.  (e) The distribution of the relation \textit{User} $\stackrel{click}{\longleftrightarrow}$ \textit{Item} in dataset Alibaba. Best viewed in color.}
    \label{fig::intro:distribution}
    \vspace{-1em}
\end{figure}

In addition, some recent studies have noticed that most real-world graphs (e.g. social networks and e-economy networks) show explicit or implicit hierarchical structures (usually manifested as power-law distributions)~\cite{krioukovHyperbolicGeometryComplex2010, bronsteinGeometricDeepLearning2017, xiongPseudoriemannianGraphConvolutional2022}, as shown in Fig.~\ref{fig::intro:distribution}(\subref{fig::intro:distribution:hierarchy}).
This observation implies an approximately exponential expansion rate of neighbors, while existing graph embedding methods are mostly built upon polynomially expanded Euclidean spaces.
The mismatches between data and embedding spaces may lead to severe limitations on model performances~\cite{nickelPoincareEmbeddingsLearning, liuHyperbolicGraphNeural, chamiHyperbolicGraphConvolutional}.
Fig.~\ref{fig::intro:distribution}(\subref{fig::intro:distribution:hierarchy_euclidean}) shows an example of representing a hierarchical graph into the Euclidean space.
To ensure the distinguishability of nodes, ideally, only one node is allowed to be embedded in each small square.
But with the exponential expansion of nodes, conflicts appear quickly.
Increasing the dimension of the Euclidean space can temporarily alleviate the conflicts, but cannot radically solve this problem.
A better choice is to find an alternative that can fit the special data distributions natively, i.e., the hyperbolic space.
Hyperbolic spaces are a class of geometric spaces with constant negative curvatures~\cite{pengHyperbolicDeepNeural2021}, and could be seen as the continuous version of tree spaces as their exponential expansion property~\cite{yangHyperbolicTemporalNetwork2022}, which fits the power-law graph data very well.
Recently, there have been many methods that attempt to represent graph data into hyperbolic spaces~\cite{chamiHyperbolicGraphConvolutional, wangHyperbolicHeterogeneousInformation2019, zhangHyperbolicGraphAttention2021, yangKHGCNTreelikenessModeling2023, baiHGWaveNetHyperbolicGraph2023}.
Fig.~\ref{fig::intro:distribution}(\subref{fig::intro:distribution:hierarchy_hyperbolic}) shows how to embed the hierarchical graph into a Poincaré disk (a mathematically equivalent model of the hyperbolic space).
%
% To verify the feasibility of hyperbolic spaces in heterogeneous graphs, we use two typical real-world datasets DBLP and Alibaba to check whether the power-law distributions exist in various types of edges.
%
The power-law characteristics are not only reflected in the global topology of a graph, but also in the distributions for single edge types in heterogeneous graphs.
As shown in Fig.~\ref{fig::intro:distribution}(\subref{fig::intro:distribution:distribution_ap}), for DBLP, we count the number of \textit{Paper} neighbors for each \textit{Author} node, and the result shows that the relation \textit{Author} $\longleftrightarrow$ \textit{Paper} obeys power-law distribution.
The same applies to the the relation \textit{User} $\stackrel{click}{\longleftrightarrow}$ \textit{Item} for Alibaba (shown in Fig.~\ref{fig::intro:distribution}(\subref{fig::intro:distribution:distribution_ui})).
This fact motivates us to try to model the respective edge types in hyperbolic spaces when representing heterogeneous graphs.

In this paper, we introduce \mymodelend, a \textbf{\underline{Dis}}entangled \textbf{\underline{H}}yperbolic \textbf{\underline{H}}eterogeneous \textbf{\underline{G}}raph \textbf{\underline{C}}onvolutional \textbf{\underline{N}}etwork, for representation learning, taking the distributions of both the entire graphs and single edges into account and meanwhile disentangling the various aspects of information.
Without loss of generality, we assume that the heterogeneous graph is a chimera of structural information and semantic information.
In order to disentangle the two aspects of information, firstly we collapse the graph into a homogeneous one so that we can eliminate semantics thoroughly and learn a pure structural representation.
%
% Simultaneously, a type-based message propagation mechanism is applied to fully mine all the information in the heterogeneous graph, cooperated with contrastive learning.
%
Simultaneously we design a hyperbolic type-based message propagation mechanism, which keeps a group of exclusive parameters for each edge type in the information aggregation to capture its unique features based on its power-law distribution in real-world scenarios.
For the comprehensive learning of endogenous information in the graph, contrastive learning is applied by constructing a perturbed view and maximizing the mutual information between representations of the original graph and perturbed view.
To disentangle the semantic part from learned endogenous information, we leverage the mutual information minimization constraint and discrimination maximization constraint during the training process to ensure the independence and difference from the structural representation.
To narrow the gap between data distribution and the representing space, all above modules are built upon the hyperbolic spaces.
Experiments on five real-world heterogeneous graph datasets show the significant superiority of our proposed \mymodelend, which fuses the disentangled structural and semantic representations via a simple Möbius addition operation, over state-of-the-art baselines on both node classification and link prediction tasks.
%
% We also experimentally quantify the level of disentanglement and study the benefits from it.
%

The contributions of this paper are summarized as follows.
\begin{itemize}[leftmargin=0em, itemindent=2em]
    \item We propose a novel heterogeneous graph representation model named \mymodelend, which tries to disentangle the structural and semantic information in hyperbolic spaces.
    \item Based on the distributions of single edge types in real-world heterogeneous graphs, we design a new message propagation and aggregation mechanism in hyperbolic spaces for each edge type to capture the differences of message patterns among various edge types.
    \item We present an effective strategy for disentangling the structural and semantic information, which simultaneously uses the mutual information constraint to keep them independent and the trainable discriminator to keep the disentangled parts distinguishing.
    \item Extensive experiments are conducted on five real-world heterogeneous graphs, demonstrating the superior performance of \mymodel on node classification and link prediction tasks.
\end{itemize}

% The remainder of the paper is structured as follows.
% %
% In Section~\ref{sec::related}, we provide a concise yet systematic review of relevant studies.
% %
% Section~\ref{sec::preli} presents necessary definitions and critical preliminaries for this paper.
% %
% Our proposed \mymodel is comprehensively introduced, covering key modules, overall framework, training strategy, and complexity analysis in Section~\ref{sec::method}.
% %
% In Section~\ref{sec::exper}, we detail the experiments and further analyze the results.
% %
% Finally, we conclude this paper in Section~\ref{sec::conlu}.
% %

\section{Related Work}  \label{sec::related}

In this section, we systematically review the relevant studies.

\subsection{Heterogeneous Graph Representation Learning}  \label{sec::related:heterogeneous}

Representation learning for heterogeneous graphs aims to learn low-dimension representation vectors for nodes, preserving structural and semantic information of the graphs~\cite{baiH2TNETemporalHeterogeneous2022}.
%
% Early methods try to extend homogeneous graph models to the domain of heterogeneous graphs by adding extra processes for semantics like PTE~\cite{tangPTEPredictiveText2015}.
%
In the current literature, meta-path~\cite{sunPathSimMetaPathbased2011} has become a general sampling approach due to its ability to express both structure and semantics simultaneously~\cite{dongMetapath2vecScalableRepresentation2017, fuHIN2VecExploreMetapaths2017, fuMAGNNMetapathAggregated2020, wangSelfsupervisedHeterogeneousGraph2021}.
Metapath2vec~\cite{dongMetapath2vecScalableRepresentation2017} and HIN2vec~\cite{fuHIN2VecExploreMetapaths2017} employ meta-paths as the fundamental strategy for random walks, followed by utilizing skip-gram like models to learn node representations.
%
% A similar node sequence sampling strategy is applied on HIN2vec~\cite{fuHIN2VecExploreMetapaths2017}.
%
Another way to use meta-path is to generalize the neighborhood, such as in MAGNN~\cite{fuMAGNNMetapathAggregated2020} and Heco~\cite{wangSelfsupervisedHeterogeneousGraph2021}, where the generalized neighborhood serves for the subsequent graph neural networks.
However, there comes a problem on how to select proper meta-paths.
It is proved that different meta-paths can make significant performance gaps~\cite{husseinAreMetaPathsNecessary2018}, while the selection of meta-paths requires manual participation. 
To overcome the above problem, JUST~\cite{husseinAreMetaPathsNecessary2018} introduces a novel heterogeneous random walk strategy that does not rely on meta-paths.
Likewise, SR-RSC~\cite{zhangSimpleMetapathfreeFramework2022} learns the heterogeneous information by automatically discovering the multi-hop relations or meta-paths.

With the advancement of graph deep learning, methods directly using graph neural networks to learn the heterogeneous relations have emerged.
R-GCN~\cite{schlichtkrullModelingRelationalData2018} learns a convolution matrix for each edge type.
%
% GEM~\cite{liuHeterogeneousGraphNeural2018} presents an attention mechanism for discerning the significance of individual node types.
%
HetGNN~\cite{zhangHeterogeneousGraphNeural2019} incorporates the influences of various types of neighboring nodes during the aggregation phase.
% , thereby encoding both structure and semantics in a single step.
%
HGT~\cite{huHeterogeneousGraphTransformer2020} introduces a transformer-like model and devises node/edge-type dependent parameters to express heterogeneous attention across each edge.
SHCF~\cite{liSequenceawareHeterogeneousGraph2021} incorporates high-order heterogeneous signals with dual-level attention.
%
% For heterogeneous hypergraph learning, HWNN~\cite{sunHeterogeneousHypergraphEmbedding2021} performs the localized hypergraph convolution with wavelet basis, while HeteHG-VAE~\cite{fanHeterogeneousHypergraphVariational2021} models the multi-level relations in heterogeneous settings with variational autoencoder architecture.
%
FAME~\cite{liuFastAttributedMultiplex2020}, DualGCN~\cite{xueMultiplexBipartiteNetwork2021}, MHGCN~\cite{yuMultiplexHeterogeneousGraph2022} and BPHGNN~\cite{fuMultiplexHeterogeneousGraph2023} study multiplex heterogeneous graphs, in which there are multiple types of edges between node pairs.
All of these methods learn mixed representations of structural and semantic information and ignore their different influences. 
In this paper, we propose to disentangle the two aspects of features and learn the separate representation for each.

\subsection{Disentangling Representation Learning}  \label{sec::related:disentangling}

Recently, disentangling representation learning has become a research hotspot and been used to factorize the unobserved structural factors from data~\cite{locatelloChallengingCommonAssumptions2019, wangDisentangledGraphCollaborative2020, wangDisenCTRDynamicGraphbased2022, zhangDyTedDisentangledRepresentation2023}.
%
% Methods based on this idea have achieved great success in many fields such as computer vision~\cite{tranDisentangledRepresentationLearning2017a, higginsVVAELEARNINGBASIC2017},  natural language processing~\cite{johnDisentangledRepresentationLearning2019}.
%
In the context of graph data, traditional GCNs~\cite{kipfSemiSupervisedClassificationGraph2017} overlook the interconnection of latent factors, potentially diminishing model robustness and rendering results less interpretable~\cite{maDisentangledGraphConvolutional2019}.
DisGNN~\cite{zhaoExploringEdgeDisentanglement2022} notices the specific semantics of edges in real world and uses a self-supervised manner for edge disentangling on node classification.
DyTed~\cite{zhangDyTedDisentangledRepresentation2023} focuses on the time-invariant and time-varying factor for discrete dynamic graphs representation.
%
% For downstream recommendation systems, MacridVAE~\cite{maLearningDisentangledRepresentations2019} represents each low-level factor of users and items into one isolated dimension, while DGCF~\cite{wangDisentangledGraphCollaborative2020} disentangles the fine granularity of user intents for each user-item interaction.
%
To enhance cross-domain recommendation performance, DisenCDR~\cite{caoDisenCDRLearningDisentangled2022} aims to disentangle domain-shared and domain-specific information, while \cite{wangDisenHANDisentangledHeterogeneous2020} decomposes high-order connectivity in user-item pairs into various meta-relations.
CurCoDis~\cite{wangCurriculumCodisentangledRepresentation2023} is designed to disentangle the latent factors across consuming environment and social environment when making decision in recommender scenarios.
%
% To sum up, it can be seen that there is no disentangling representation learning model specifically for heterogeneous graphs at present, except for the preliminary and inadequate discussion in~\cite{zhaoExploringEdgeDisentanglement2022}.
%
To sum up, it can be seen that there is no method structurally and semantically disentangling the features in heterogeneous graphs at present, except for the preliminary and inadequate discussion in~\cite{zhaoExploringEdgeDisentanglement2022}.

\subsection{Hyperbolic Graph Representation Learning}  \label{sec::related:hyperbolic}

Due to the explicit or implicit hierarchical structures in real-world graphs~\cite{krioukovHyperbolicGeometryComplex2010, bronsteinGeometricDeepLearning2017}, representation learning in hyperbolic spaces has attracted great attention in the past few years.
Similar to the development trend of Euclidean manifold, early hyperbolic approaches use shallow models for representation learning~\cite{nickelPoincareEmbeddingsLearning, nickelLearningContinuousHierarchies}, but achieve outstanding performance compared with corresponding Euclidean methods.
The significant improvement motivates researchers to extend various graph neural networks to hyperbolic geometry (equivalent models, e.g. Poincaré ball and Lorentz model)~\cite{liuHyperbolicGraphNeural, chamiHyperbolicGraphConvolutional, zhangHyperbolicGraphAttention2021, yangKHGCNTreelikenessModeling2023}.

With regard to heterogeneous graphs, hyperbolic models also make encouraging progress.
HHNE~\cite{wangHyperbolicHeterogeneousInformation2019} uses the meta-path guided random walk strategy followed with a shallow representation model, and HHNE++~\cite{zhangEmbeddingHeterogeneousInformation} further improves it by embedding different relations in separate spaces.
H$^2$TNE~\cite{baiH2TNETemporalHeterogeneous2022} imposes the temporal information and avoids the selection of meta-paths through specially designed constraints.
ConE~\cite{baiModelingHeterogeneousHierarchies2021} models the multiple heterogeneous hierarchies in knowledge graphs with cone containment constraints in different hyperbolic subspaces.
In recommendation systems, HGCC~\cite{zhangHGCCEnhancingHyperbolic2023} puts forward a hyperbolic GCN collaborative filtering method to enhance the learning for user-item interactions.
Summarily, researchers have made great efforts to adapt the power-law distribution of heterogeneous information in graphs, but there is no hyperbolic method of disentangling different factors influencing the graph construction.

\section{Preliminaries}  \label{sec::preli}

In this section, we introduce some formalized definitions in heterogeneous graph representation learning, as well as some critical concepts about hyperbolic geometry.

\subsection{Problem Definition}  \label{sec::preli:definition}

This paper focuses on the representation learning of heterogeneous graphs. 
Heterogeneous graphs are defined as~\cite{zhangEmbeddingHeterogeneousInformation}.
\begin{definition}[Heterogeneous Graphs]  \label{def::preli:heterogeneous}
A heterogeneous graph can be formalized as $\mathcal{G} = \left( \mathcal{V}, \mathcal{E}, \mathcal{L}, \phi, \psi \right)$, in which $\mathcal{V}$ and $\mathcal{E}$ are the sets of nodes and edges. 
$\phi: \mathcal{V} \rightarrow \mathcal{L}_{\mathcal{V}}$ is the mapping function from nodes to corresponding types and $\psi: \mathcal{E} \rightarrow \mathcal{L}_{\mathcal{E}}$ is the type mapping function for edges.
$\mathcal{L}_{\mathcal{V}}$ and $\mathcal{L}_{\mathcal{E}}$ are subsets of $\mathcal{L}$, referring to the node type set and edge type set.
It is satisfied that $\mathcal{L}_{\mathcal{V}} \cup \mathcal{L}_{\mathcal{E}} = \mathcal{L}$, $\mathcal{L}_{\mathcal{V}} \cap \mathcal{L}_{\mathcal{E}} = \emptyset$ and $\left| \mathcal{L}_{\mathcal{V}} \right| + \left| \mathcal{L}_{\mathcal{E}} \right| > 2$.
% $\mathcal{L}_{\mathcal{V}}$ and $\mathcal{L}_{\mathcal{E}}$ denote the sets of node and edge types respectively and satisfy $\left| \mathcal{L}_{\mathcal{V}} \right| + \left| \mathcal{L}_{\mathcal{E}} \right| > 2$.
% %
% $\mathcal{L}$ is the conjunctive form of $\mathcal{L}_{\mathcal{V}}$ and $\mathcal{L}_{\mathcal{E}}$.
%
\end{definition}

Fig.~\ref{fig::intro:distribution}(\subref{fig::intro:information:heterogeneous}) is an example of heterogeneous graphs. 
Based on Definition~\ref{def::preli:heterogeneous}, representation learning on them is defined as:
\begin{definition}[Heterogeneous Graph Representation Learning]  \label{def::preli:representaion}
Given a heterogeneous graph $\mathcal{G} = \left( \mathcal{V}, \mathcal{E}, \mathcal{L}, \phi, \psi \right)$, heterogeneous graph representation learning (also called heterogeneous graph embedding) aims to output a node representation matrix $\mathbf{Z} \in \mathbb{R}^{\left| \mathcal{V} \right| \times d}$.
Each row of~$\mathbf{Z}$ is a representation vector corresponding to a node, and $d \ll \left| \mathcal{V} \right|$ is the number of vector dimensions.
The node representation matrix $\mathbf{Z}$ is required to preserve the original structure and semantic correlations.
\end{definition}

In this paper, we try to disentangle the structural and semantic information of heterogeneous graphs.
Disentangled heterogeneous graph representation learning is defined as:

\begin{definition}[Disentangled Heterogeneous Graph Representation Learning]  \label{def::preli:disentangled}
Given a heterogeneous graph $\mathcal{G} = \left( \mathcal{V}, \mathcal{E}, \mathcal{L}, \phi, \psi \right)$, disentangled heterogeneous graph representation learning aims to learn: {\rm (1)} a structural representation matrix $\mathbf{Z}_{st} \in \mathbb{R}^{\left| \mathcal{V} \right| \times d_{st}}$ corresponding to the structural information of $\mathcal{G}$; {\rm (2)} a semantic representation matrix $\mathbf{Z}_{se} \in \mathbb{R}^{\left| \mathcal{V} \right| \times d_{se}}$ corresponding to the semantic information of $\mathcal{G}$.
The final disentangled representation matrix $\mathbf{Z} \in \mathbb{R}^{\left| \mathcal{V} \right| \times d}$ is the fusion of above two types of representations: $\mathbf{Z} = \varphi \left( \mathbf{Z}_{st}, \mathbf{Z}_{se} \right)$, in which $\varphi (\cdot)$ is the fusion operation.
\end{definition}

\subsection{Hyperbolic Geometry}  \label{sec::preli:hyperbolic}

Next we will briefly introduce some critical concepts about hyperbolic geometry involved in this paper.
For systematic and comprehensive introduction, please refer to ~\cite{leeManifoldsDifferentialGeometry2022}.

\textit{Manifold and Tangent Space.}
Mathematically, a manifold $\mathbb{M}$ is a topological space locally resembling Euclidean space near each point.
In an n-dimensional manifold, each point has a neighborhood homeomorphic to an open subset of n-dimension Euclidean space.
For $\forall \textbf{x} \in \mathbb{M}$, the tangent space $\mathcal{T}_{\textbf{x}}\mathbb{M}$ is the linear space of all tangent vectors at $\textbf{x}$.
$\mathcal{T}_{\textbf{x}}\mathbb{M}$ is the first order approximation of $\mathbb{M}$ around $\textbf{x}$ and could be used to describe the local properties of $\mathbb{M}$ at this point.

\textit{Riemannian Metric and Riemannian Manifold.}
For a connected manifold $\mathbb{M}$, the Riemannian metric gives an inner product operation $g^{\mathbb{M}}: \mathcal{T}_{\textbf{x}}\mathbb{M} \times \mathcal{T}_{\textbf{x}}\mathbb{M} \rightarrow \mathbb{R}$, and the inner product smoothly depends on $\textbf{x}$.
Based on $g^{\mathbb{M}}$, the distance between arbitrary points $\textbf{x}, \textbf{y} \in \mathbb{M}$ could be defined as
\begin{equation}
    d(\textbf{x}, \textbf{y}) = \inf_{\gamma}  \ell(\gamma),
    \label{eq::preli:dis_on_manifold}
\end{equation}
where $\gamma$ takes all smooth curves connecting $\textbf{x}$ and $\textbf{y}$.
$l(\gamma)$ is the length integral of $\gamma$.
A Riemannian manifold is a differential manifold $\mathbb{M}$ with corresponding Riemannian metric.

\textit{Exponential Map and Logarithmic Map.}
To map points between the manifold $\mathbb{M}$ and its tangent space $\mathcal{T}_{\textbf{x}}\mathbb{M}$ at $\textbf{x}$, exponential map and logarithmic map are defined.
Exponential map $\exp_{\textbf{x}}^{\mathbb{M}}: \mathcal{T}_{\textbf{x}}\mathbb{M} \rightarrow \mathbb{M}$ maps points from the tangent space $\mathcal{T}_{\textbf{x}}\mathbb{M}$ to the manifold $\mathbb{M}$ itself, and logarithmic map $\log_{\textbf{x}}^{\mathbb{M}}: \mathbb{M} \rightarrow \mathcal{T}_{\textbf{x}}\mathbb{M}$ is the inverse of $\exp_{\textbf{x}}^{\mathbb{M}}$.

\textit{Hyperbolic Space and Poincaré Ball.}
The hyperbolic space $\left( \mathbb{H}_c^n, g^{c, \mathbb{H}} \right)$ is a simply connected n-dimensional complete Riemannian manifold $\mathbb{H}_c^n$ with a negative curvature $-c$.
$g^{c, \mathbb{H}}$ is the Riemannian metric.
Poincaré ball model is an extensively used isometric model of hyperbolic spaces.
Corresponding to $\left( \mathbb{H}_c^n, g^{c, \mathbb{H}} \right)$, the Poincaré ball model $\left( \mathbb{B}_c^n, g^{c} \right)$ is defined as
\begin{equation}
    \begin{split}
        & \mathbb{B}_c^n = \left\{ \textbf{x} \in \mathbb{R}^n: c \| \textbf{x} \|^2 < 1 \right\}, \\
        & g^{c}_{\textbf{x}} = \left( \lambda_{\textbf{x}}^c \right)^2 g^{E}, \quad \lambda_{\textbf{x}}^c = \frac{2}{1 - c \| \textbf{x} \|^2},
        \end{split}
    \label{eq::preli:poincare}
\end{equation}
in which $g^{E} = \textbf{I}_n$ is the Euclidean metric tensor. 
In this paper, we construct our method based on the Poincaré ball.

It can be seen that the Poincaré ball manifold $\mathbb{B}_c^n$ is an open ball with the radius $1/\sqrt{c}$.
Induced from Eq.~(\ref{eq::preli:dis_on_manifold}), the distance between two points $\textbf{x}, \textbf{y} \in \mathbb{B}_c^n$ could be calculated as
\begin{equation}
    d_c^{\mathbb{B}}(\textbf{x}, \textbf{y}) = \frac{2}{\sqrt{c}} \cdot \tanh^{-1} \left( \sqrt{c} \left\| - \textbf{x} \oplus_c \textbf{y} \right\| \right),
    \label{eq::preli:dis_on_poincare}
\end{equation}
in which $\oplus_c$ is the Möbius addition operation
\begin{equation}
    \textbf{x} \oplus_c \textbf{y} = \frac{\left( 1 + 2 c \langle \textbf{x}, \textbf{y} \rangle + c \| \textbf{y} \|^2 \right) \textbf{x} + \left( 1 - c \| \textbf{x} \|^2 \right) \textbf{y}}{1 + 2 c \langle \textbf{x}, \textbf{y} \rangle + c^2 \| \textbf{x} \|^2 \| \textbf{y} \|^2}.
    \label{eq::preli:addition}
\end{equation}
The exponential map $\exp_{\textbf{x}}^{c}: \mathcal{T}_{\textbf{x}}\mathbb{B}_c^n \rightarrow \mathbb{B}_c^n$ and logarithmic map $\log_{\textbf{x}}^{c}: \mathbb{B}_c^n \rightarrow \mathcal{T}_{\textbf{x}}\mathbb{B}_c^n$ on $\mathbb{B}_c^n$ are calculated as
\begin{equation}
    \exp_{\textbf{x}}^{c} (\textbf{v}) = \textbf{x} \oplus_c \left( \tanh \left( \frac{\sqrt{c} \lambda_{\textbf{x}}^c \| \textbf{v} \|}{2} \right) \frac{\textbf{v}}{\sqrt{c} \| \textbf{v} \|} \right),
    \label{eq::preli:expmap}
\end{equation}
\begin{equation}
    \log_{\textbf{x}}^{c} (\textbf{y}) = d_c^{\mathbb{B}} \left( \textbf{x}, \textbf{y} \right) \frac{- \textbf{x} \oplus_c \textbf{y}}{\lambda_{\textbf{x}}^c \left\| - \textbf{x} \oplus_c \textbf{y} \right\|},
    \label{eq::preli:logmap}
\end{equation}
in which $\textbf{x}, \textbf{y} \in \mathbb{B}_c^n$, $\textbf{v} \in \mathcal{T}_{\textbf{x}}\mathbb{B}_c^n$, $\textbf{x} \neq \textbf{y}$ and $\textbf{v} \neq \textbf{0}$.
In this paper, we take the origin point $\textbf{o}$ as reference point $\textbf{x}$ to balance the directional errors unless otherwise specified.

\section{Methodology}  \label{sec::method}

\begin{figure*}
    \centering
    \includegraphics[width=0.8\linewidth]{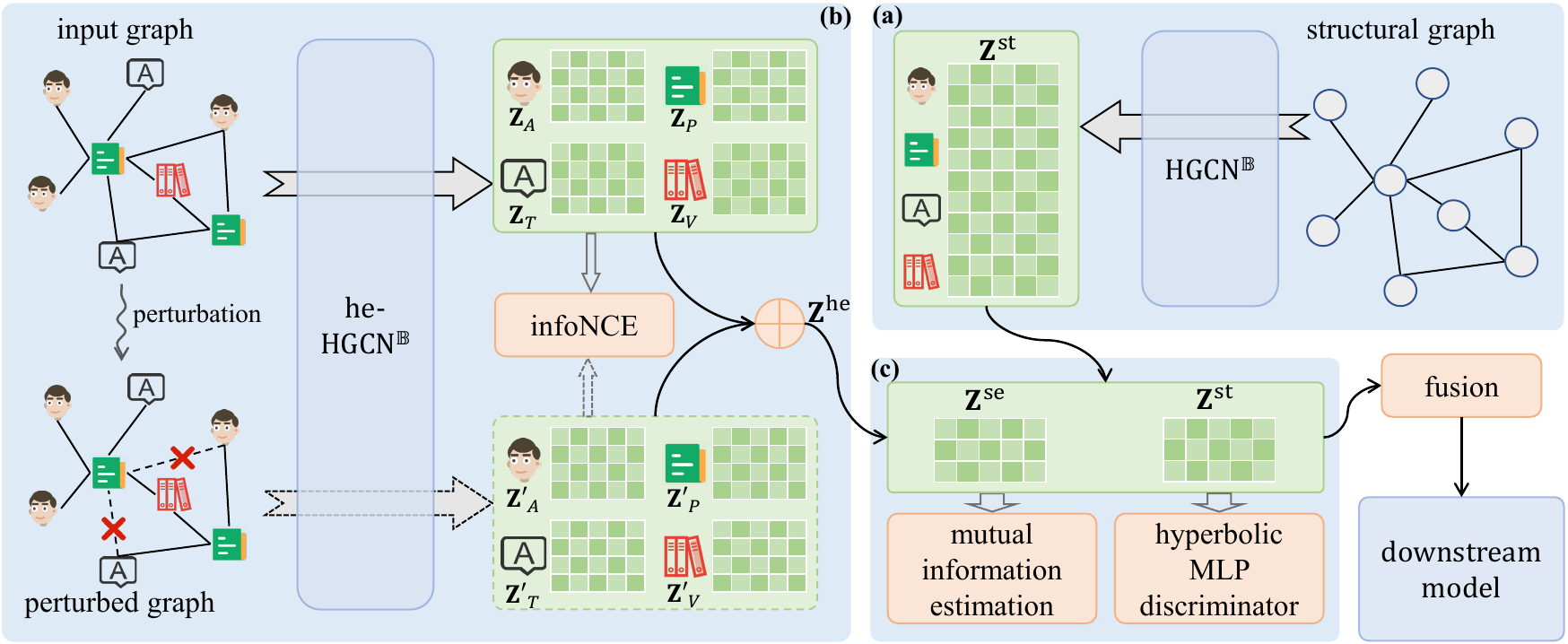}
    \caption{The overall framework of ~\mymodelend. Three main modules are: (a) structural information learning module, (b) heterogeneous graph contrastive learning module and (c) disentangling module.}
    \label{fig::method:dis-h2gcn}
    \vspace{-1em}
\end{figure*}

This section presents our proposed model \mymodelend.
As shown in Fig.~\ref{fig::method:dis-h2gcn}, \mymodel mainly consists of three modules: (1) structural information learning module, which eliminates semantics thoroughly and learns the pure structural embeddings; (2) heterogeneous graph contrastive learning module, which aims to mine all the information contained in heterogeneous graphs with the type-based message propagation mechanism and contrastive training strategy; (3) disentangling module, which tries to disentangle the semantic and structural aspects of information effectively.
%
% The three modules are detailedly described in Section~\ref{sec::method:structural},~\ref{sec::method:heterogeneous} and~\ref{sec::method:disentangle}, separately.
% %
% Section~\ref{sec::method:framework} summarizes the overall framework of \mymodel and model training.
% %
% % Section~\ref{sec::method:training} introduces the model training, and 
% %
% Section~\ref{sec::method:complexity} further analyses the time complexity.
%

\subsection{Structural Information Learning Module}  \label{sec::method:structural}

In order to learn the pure structural embeddings from heterogeneous information, a natural idea is to remove semantics from the source of input data.
Structural information learning module implements this by eliminating all node types and edge types of the input graph $\mathcal{G}$ into a homogeneous graph $\mathcal{G}_{\text{st}}$, called structural graph.
A hyperbolic graph convolutional network upon Poincaré ball (HGCN$^{\mathbb{B}}$) is then applied, constructed in a manner similar to HGCN~\cite{chamiHyperbolicGraphConvolutional}.

Before the structural graph $\mathcal{G}_{\text{st}}$ is fed into HGCN$^{\mathbb{B}}$, an initialization process for node feature matrix $\mathbf{X}$ is performed.
On the one hand, the original node features need to be aligned to a uniform dimension.
On the other hand, only node features mapped into the hyperbolic space can participate in subsequent operations.
The initialization process is calculated as
\begin{equation}
    \mathbf{X}^{\mathbb{B}} = \exp_{\textbf{o}}^{c^0} \left( \mathbf{X} \mathbf{W}^0 \right),
    \label{eq::method:initialization}
\end{equation}
where $\mathbf{W}^0 \in \mathbb{R}^{d^{\prime} \times d_0}$ is the trainable parameter, $d^{\prime}$ is the original feature dimension and $d_0$ is the aligned dimension.
The initialized node feature matrix $\mathbf{X}^{\mathbb{B}}$ is on the $d_0$-dimension Poincaré ball $\mathbb{B}_{c^0}^{d_0}$ with the curvature $-c^0$, and is to be input into HGCN$^{\mathbb{B}}$ together with the adjacent matrix $\mathbf{A}$.

Analogous to traditional GCNs, HGCN$^{\mathbb{B}}$ is built upon the hyperbolic space.
A single HGCN$^{\mathbb{B}}$ layer often consists of three key operations: hyperbolic linear transformation, hyperbolic feature aggregation and hyperbolic non-linear activation.
Take the $l$-th layer of HGCN$^{\mathbb{B}}$ as an example.
Hyperbolic linear transformation aims to extract informative features for better message propagation, which is expressed as
\begin{equation}
    \textbf{h}_i^{l} = \mathbf{W}^l \otimes_{c^{l-1}} \textbf{x}_i^{l-1} \oplus_{c^{l-1}} \textbf{b}^l,
    \label{eq::method:transformation}
\end{equation}
where $\mathbf{W}^l$ and $\textbf{b}^l$ are trainable parameters of layer $l$.
$\textbf{x}_i^{l-1}$ is the hyperbolic feature representation vector of node $i$ at layer $l-1$ and $\textbf{h}_i^{l}$ is the transformed feature vector of node $i$ at layer $l$, with curvature $-c^{l-1}$.
For the first layer $l=1$, $\textbf{x}_i^{0}$ is the row corresponding to node $i$ in matrix $\mathbf{X}^{\mathbb{B}}$.
The matrix-vector multiplication $\otimes_c$ is defined as~\cite{yangDiscretetimeTemporalNetwork2021}
\begin{equation}
    \mathbf{M} \otimes_c \textbf{x} = \exp_{\textbf{o}}^{c} \left( \mathbf{M} \log_{\textbf{o}}^{c} ( \textbf{x} ) \right), \quad \mathbf{M} \in \mathbb{R}^{n \times n}, \textbf{x} \in \mathbb{B}_c^n.
    \label{eq::method:multiplication}
\end{equation}
Weighted summation is operated on transformed representation vectors as hyperbolic feature aggregation process by
\begin{equation}
\begin{split}
    \textbf{y}_i^{l} &= \mathop{\text{Agg}_{c^l}}\limits_{j \in \mathcal{N}(i)} \left( \textbf{h}_j^{l} \right) \\
    &= \exp_{\textbf{o}}^{c^l} \left( \sum_{j \in \mathcal{N}(i)} \omega_{ij} \cdot \log_{\textbf{o}}^{c^{l-1}} \left( \textbf{h}_j^{l} \right) \right),
    \label{eq::method:aggregation}
\end{split}
\end{equation}
where $\mathcal{N}(i)$ is the neighbor set of node $i$ indicated by the adjacent matrix $\mathbf{A}$. 
It is worth noting that the curvature of $\textbf{h}_j^{l}$ is $-c^{l-1}$ while that of $\textbf{y}_i^{l}$ is changed to $-c^l$ through the exponential map.
% in order to keep the operations on correct manifolds.
%
$\omega_{ij}$ is the weight of node $j$ for aggregation of node $i$, defined as~\cite{yangHyperbolicTemporalNetwork2022}
\begin{equation}
    \omega_{ij} = \frac{1}{\sqrt{|\mathcal{N}(i)| \cdot |\mathcal{N}(j)|}}.
    \label{eq::method:weight}
\end{equation}
This definition satisfies a plain idea that the greater the degree of one node, the less affected it is by a single neighbor node.
Hyperbolic non-linear activation is achieved by
\begin{equation}
    \textbf{x}_i^{l} = \exp_{\textbf{o}}^{c^l} \left( \sigma \left( \log_{\textbf{o}}^{c^l} \left( \textbf{y}_i^{l} \right) \right) \right),
    \label{eq::method:activation}
\end{equation}
where $\sigma (\cdot)$ is the Euclidean non-linear activation function and $\textbf{x}_i^{l}$ is the representation vector of node $i$ at layer $l$. 

By stacking $L$ above layers, the HGCN$^{\mathbb{B}}$ is constructed.
The final structural embedding matrix $\mathbf{Z}^{\text{st}}$ is exactly a permutation of vector set $\left\{ \textbf{x}_i^{L} | i \in \mathcal{V} \right\}$, the output when $l=L$ in HGCN$^{\mathbb{B}}$.

\subsection{Heterogeneous Graph Contrastive Learning Module}  \label{sec::method:heterogeneous}

Nodes in heterogeneous graphs have various neighbors connected by various types of edges with various semantics. 
Using shared parameters for different edge types in the message propagation of GCN-like models would fail to learn the semantic differences.
%
% In this part, we design a heterogeneous hyperbolic graph convolutional network (he-HGCN$^{\mathbb{B}}$) architecture, which applies a type-based message propagation process to address this issue.
%
Consequently, researchers begin to consider aggregating information by the types of neighbors, to learn the semantics implicated in the connection patterns~\cite{schlichtkrullModelingRelationalData2018}.
In this paper, we generalize this concept into hyperbolic spaces and design a hyperbolic type-based message propagation process based on the observation that distributions of single edge types in real-word heterogeneous graphs obey the power-law.
%
% We also apply the contrastive learning strategy in this module to further improve the model's performance and robustness.
%
In order to comprehensively mine the endogenous information of the input heterogeneous graph and further improve model's performance, we also apply contrastive learning in this module by constructing two different graph views and maximizing the mutual information between representations of the two views.

\subsubsection{Heterogeneous Hyperbolic Graph Convolutional Networks}  \label{sec::method:heterogeneous:he-hgcn}

\begin{figure*}
    \centering
    \includegraphics[width=\linewidth]{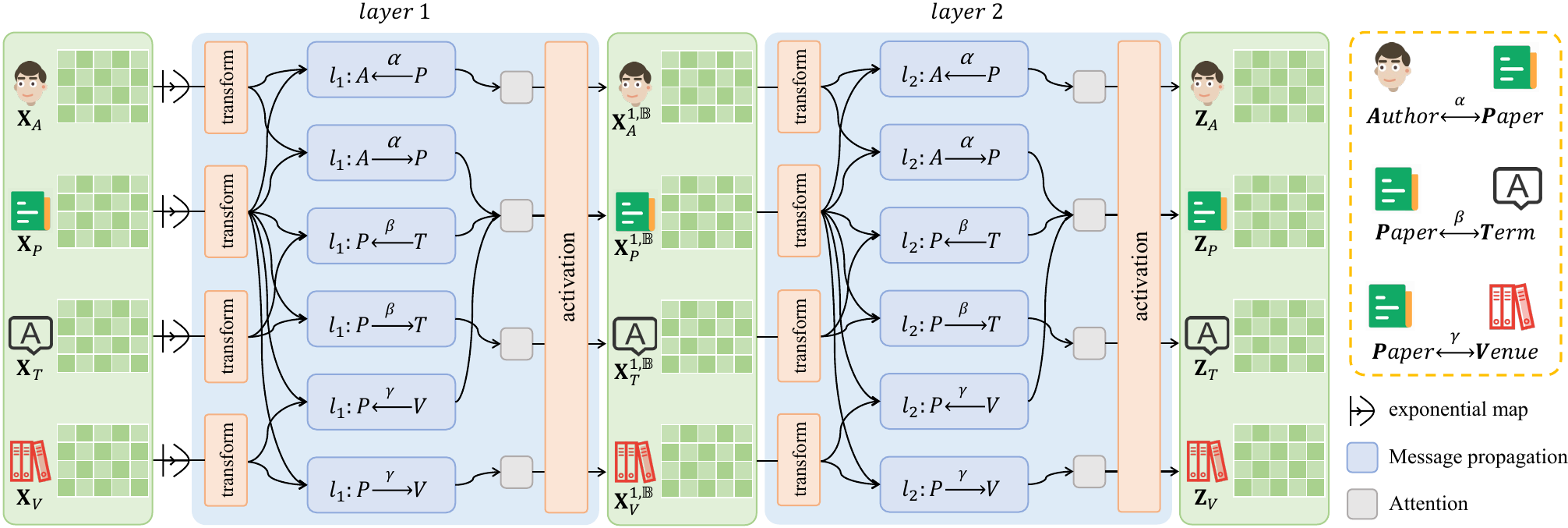}
    \caption{An example of he-HGCN$^{\mathbb{B}}$ architecture upon the toy academic network. Best viewed in color.}
    \label{fig::method:he-hgcn}
    \vspace{-1em}
\end{figure*}

Fig.~\ref{fig::method:he-hgcn} shows an example of he-HGCN$^{\mathbb{B}}$ upon a toy academic network.
In this network, there are four node types (\textit{\textbf{A}uthor}, \textit{\textbf{P}aper}, \textit{\textbf{T}erm} and \textit{\textbf{V}enue}) and three bi-directional edge types (\textit{A} $\stackrel{\alpha}{\longleftrightarrow}$ \textit{P}, \textit{P} $\stackrel{\beta}{\longleftrightarrow}$ \textit{T} and \textit{P} $\stackrel{\gamma}{\longleftrightarrow}$ \textit{V}).
To be compatible with directed graphs, we treat bi-directional edges as two edges with different types and opposite directions.
Consequently, this example is subsequently handled as containing six edge types.

In he-HGCN$^{\mathbb{B}}$, the embedding vectors are calculated separately for different node types, denoted as $\mathbf{X}_{l_v}$ in which $l_v \in \mathcal{L}_{\mathcal{V}}$ refers to the edge type.
Firstly, similar to Eq.~(\ref{eq::method:initialization}) in structural information learning module, for all $l_v$ in $\mathcal{L}_{\mathcal{V}}$, the initialization of feature matrix $\mathbf{X}_{l_v}$ is 
\begin{equation}
    \mathbf{X}_{l_v}^{\mathbb{B}} = \exp_{\textbf{o}}^{c^0} \left( \mathbf{X}_{l_v} \mathbf{W}_{\text{he}}^0 \right),
    \label{eq::method:he-initialization}
\end{equation}
where $\mathbf{W}_{\text{he}}^0$ is the trainable parameter and $\mathbf{X}_{l_v}^{\mathbb{B}}$ is the initialized feature matrix with the curvature $-c^0$.
Then $\{ \mathbf{X}_{l_v}^{\mathbb{B}} | l_v \in \mathcal{L}_{\mathcal{V}} \}$ and the type specific adjacent matrix set $\{ \mathbf{A}_{l_e} | l_e \in \mathcal{L}_{\mathcal{E}} \}$ are fed into stacked he-HGCN$^{\mathbb{B}}$ layers. 
$\mathbf{A}_{l_e}$ is the adjacency under the edge type $l_e$.
Independent hyperbolic linear transformation for each node type $l_v$ at layer $l$ is performed as
\begin{equation}
    \textbf{h}_{i(l_v)}^{l} = \exp_{\textbf{o}}^{c_{l_v}^{l}} \left( \log_{\textbf{o}}^{c^{l-1}} \left( \mathbf{W}_{l_v}^l \otimes_{c^{l-1}} \textbf{x}_{i(l_v)}^{l-1} \oplus_{c^{l-1}} \textbf{b}_{l_v}^l \right) \right),
    \label{eq::method:he-transformation}
\end{equation}
where $\phi(i) = l_v$ and other notations are analogous to Eq.~(\ref{eq::method:transformation}). The transformed features of nodes with type $l_v$ are projected into the hyperbolic space with curvature $-c_{l_v}^l$ from that with curvature $-c^{l-1}$ by logarithmic map and exponential map.

The main difference between he-HGCN$^{\mathbb{B}}$ and HGCN$^{\mathbb{B}}$ lies on the type-based message propagation in feature aggregation.
To better learn the various semantics embodied by different edge types, he-HGCN$^{\mathbb{B}}$ designs a two-step feature aggregation operation.
(1) \textit{Inner-type aggregation}, which can be seen as type-specialized Eq.~(\ref{eq::method:aggregation}).
Take the message propagation block $l_1: \textit{A} \stackrel{\alpha}{\longrightarrow} \textit{P}$ in Fig.~\ref{fig::method:he-hgcn} as an example.
The edge type \textit{A} $\stackrel{\alpha}{\longrightarrow}$ \textit{P} is with the message propagation direction from node type \textit{A} to \textit{P}, and its corresponding inner-type aggregation follows
\begin{equation}
\begin{split}
    \textbf{y}_{i(\textit{P}) \sim \alpha}^{1} &= \mathop{\text{Agg}_{c_{A}^1}^{\textit{A} \stackrel{\alpha}{\longrightarrow} \textit{P}}}\limits_{j \in \mathcal{N}_{\alpha}(i)} \left( \textbf{h}_{j(\textit{A})}^{1} \right) \\
    &= \exp_{\textbf{o}}^{c_{A}^1} \left( \sum_{j \in \mathcal{N}_{\alpha}(i)} \omega_{ij}^{\alpha} \cdot \log_{\textbf{o}}^{c_{A}^{1}} \left( \textbf{h}_{j(\textit{A})}^{1} \right) \right),
    \label{eq::method:he-inner-aggregation}
\end{split}
\end{equation}
where $\phi(i) = \textit{P}$, $\phi(j) = \textit{A}$, $\mathcal{N}_{\alpha} (\cdot)$ is the set of neighbors connected by edges with type \textit{A} $\stackrel{\alpha}{\longrightarrow}$ \textit{P}, $\omega_{ij}^{\alpha}$ is the weight of node $j$ for node $i$ in this inner-type aggregation, and $\textbf{y}_{i(\textit{P}) \sim \alpha}^{1}$ refers to the output of block $l_1: \textit{A} \stackrel{\alpha}{\longrightarrow} \textit{P}$.
$\omega_{ij}^{\alpha}$ is
\begin{equation}
    \omega_{ij}^{\alpha} = \frac{1}{\sqrt{|\mathcal{N}_{\alpha} (i)| \cdot |\mathcal{N}_{\alpha} (j)|}}.
    \label{eq::method:he-inner-weight}
\end{equation}
(2) \textit{Inter-type aggregation}.
After performing inner-type aggregation for all edge types, several separate representations are obtained for each node type.
For example in Fig.~\ref{fig::method:he-hgcn}, node type $P$ in layer 1 involves three output representations of message propagation blocks $l_1: \textit{A} \stackrel{\alpha}{\longrightarrow} \textit{P}$, $l_1: \textit{P} \stackrel{\beta}{\longleftarrow} \textit{T}$ and $l_1: \textit{P} \stackrel{\beta}{\longleftarrow} \textit{V}$.
Since different edge types make different contributions for node type \textit{P}, the hyperbolic attention mechanism is adopted in inter-type aggregation to adaptive learn the importance of each edge type, expressed as
\begin{equation}
\begin{split}
    \textbf{y}_{i(\textit{P})}^{1} &= \mathop{\text{Agg}_{c_{P}^1}}\limits_{l_e \in \mathcal{L}_{\mathcal{E}}^{\textit{P}}} \left( \textbf{y}_{i(\textit{P}) \sim l_e}^{1} \right) \\
    &= \exp_{\textbf{o}}^{c_{P}^1} \left( \sum_{l_e \in \mathcal{L}_{\mathcal{E}}^{\textit{P}}} \rho_{l_e} \cdot \log_{\textbf{o}}^{c_{l_v}^1} \left( \textbf{y}_{i(\textit{P}) \sim l_e}^{1} \right) \right),
    \label{eq::method:he-inter-aggregation}
\end{split}
\end{equation}
where $\phi(i) = \textit{P}$, $l_v$ is the other node type except $\textit{P}$ in the ends of edges with type $l_e$, $\mathcal{L}_{\mathcal{E}}^{\textit{P}}$ is the set of edge types related to node type \textit{P}, and $\textbf{y}_{i(\textit{P})}^{1}$ is the aggregated feature representation for node $i$.
The attention weight $\rho_{l_e}$ is given by
\begin{equation}
\begin{split}
    \rho_{l_e} &= \frac{\exp (r_{l_e})}{\sum_{l_e^{\prime} \in \mathcal{L}_{\mathcal{E}}^P} \exp(r_{l_e^{\prime}})}, \\
    r_{l_e} &= \iota \left( \textbf{u} \left( \log_{\textbf{o}}^{c_{P}^1} \left( \tilde{\textbf{y}}_{i(\textit{P})}^{1} \right) \| \log_{\textbf{o}}^{c_{l_v}^1} \left( \textbf{y}_{i(\textit{P}) \sim l_e}^{1} \right) \right) \right),
    \label{eq::method:he-inter-weight}
\end{split}
\end{equation}
where $\textbf{u}$ is a trainable row vector, $\|$ denotes vector concatenation, $\iota$ is the LeakyReLU function, and $\tilde{\textbf{y}}_{i(\textit{P})}^{1}$ is the aggregated feature representation in the last training epoch.

Subsequently, the shared non-linear activation acts on representations of each node type $l_v$, calculated as
\begin{equation}
    \textbf{x}_{i(l_v)}^{l} = \exp_{\textbf{o}}^{c^l} \left( \sigma \left( \log_{\textbf{o}}^{c_{l_v}^l} \left( \textbf{y}_{i(l_v)}^{l} \right) \right) \right),
    \label{eq::method:he-activation}
\end{equation}
in which $\textbf{x}_{i(l_v)}^{l}$ is the representation vector of node $i$ at the $l$-th he-HGCN$^{\mathbb{B}}$ layer.
For a complete he-HGCN$^{\mathbb{B}}$, the last layer outputs the final feature matrix set $\left\{ \mathbf{Z}_{l_v} | l_v \in \mathcal{L}_{\mathcal{V}} \right\}$.

\subsubsection{Contrastive Learning}  \label{sec::method:heterogeneous:contrastive}

Contrastive learning has been proved to have great superiority in many unsupervised or self-supervised graph learning tasks~\cite{wangSelfsupervisedHeterogeneousGraph2021}.
Some researches also apply it into supervised tasks and gain great improvement, like~\cite{fuMultiplexHeterogeneousGraph2023}.
Inspired by that, we design a contrastive learning strategy in this module to mine the endogenous information contained in heterogeneous graph data itself.

We regard the original input graph as one view for contrastive learning, and the other view is constructed by putting random perturbations on the edges of input graph.
The perturbations are equally proportional for various edge types.
Both views are fed into the same he-HGCN$^{\mathbb{B}}$.
For different node types in $\mathcal{L}_{\mathcal{V}}$, two feature matrix sets $\left\{ \mathbf{Z}_{l_v} | l_v \in \mathcal{L}_{\mathcal{V}} \right\}$ and $\left\{ \mathbf{Z}_{l_v}^{\prime} | l_v \in \mathcal{L}_{\mathcal{V}} \right\}$ are obtained, corresponding to input graph view and perturbed graph view separately.
In the learning process, we treat the same nodes in both views as positive samples and different nodes as negative samples.
With the node embeddings of both views, we use the InfoNCE~\cite{oordRepresentationLearningContrastive2019} as cross-view contrastive loss, calculated as
\begin{equation}
    \mathcal{L}_{cl} = - \sum_{i \in \mathcal{V}} \log \frac{\exp \left( s \left( \textbf{z}_{i(\phi (i))}, \textbf{z}_{i(\phi (i))}^{\prime} \right) / \tau \right)}{\sum_{j \in \mathcal{V}} \exp \left( s \left( \textbf{z}_{i(\phi (i))}, \textbf{z}_{j(\phi (j))}^{\prime} \right) / \tau \right)},
    \label{eq::method:cl_loss}
\end{equation}
where $\textbf{z}_{i(\phi (i))}$ and $\textbf{z}_{i(\phi (i))}^{\prime}$ are embeddings of node $i$ from input graph view and perturbed graph view, $\tau$ is a temperature parameter, and $s(\cdot, \cdot)$ is the cosine similarity function.
Benefited from the conformal property of Poincaré ball model, the angles between vectors upon the manifold are identical to them in the Euclidean space and cosine similarity can be utilized directly without logarithmic map.
The contrastive learning makes a constraint between the input graph view and perturbed graph view, and allows them to collaboratively supervise each other, improving the robustness of the learning process.

For entire heterogeneous graph contrastive learning module, the output is the combination for both views, defined as
\begin{equation}
    \mathbf{Z}^{\text{he}} = \mathop{\|}\limits_{l_v \in \mathcal{L}_{\mathcal{V}}} \left( \mathbf{Z}_{l_v} \oplus_{c^{L}} \mathbf{Z}_{l_v}^{\prime} \right),
    \label{eq::method:contrastive_combine}
\end{equation}
where $c^{L}$ is the curvature of last layer in he-HGCN$^{\mathbb{B}}$, and $\|$ refers to matrix stack.

\subsection{Disentangling Module}  \label{sec::method:disentangle}

From the structural learning module and heterogeneous graph contrastive learning module, we get two embeddings of the original graph, in which $\mathbf{Z}^{\text{st}}$ only represents the structural information while $\mathbf{Z}^{\text{he}}$ also contains the semantics.
In order to separate the semantic representations from $\mathbf{Z}^{\text{he}}$, we propose the disentangling module to enhance the independence and difference between the two embeddings.

Mutual information is a widely used concept in information theory for the quantification of independence modeling.
For two discrete random variables $\mathbf{X}$ and $\mathbf{Y}$, it is defined as
\begin{equation}
    I (\mathbf{X}, \mathbf{Y}) = \sum_{x \in \mathbf{X}} \sum_{y \in \mathbf{Y}} p(x, y) \log \frac{p(x, y)}{p(x) \cdot p(y)},
    \label{eq::method:mutual_info}
\end{equation}
where $p(x)$ and $p(y)$ are the marginal probability distributions, and $p(x, y)$ is the joint probability distribution of $\mathbf{X}$ and $\mathbf{Y}$.
However, Eq.~(\ref{eq::method:mutual_info}) is incalculable in our task because the involved distributions are unavailable.
Alternatively, we adopt CLUB~\cite{chengCLUBContrastiveLogratio2020} as the upper bound estimation of mutual information.
CLUB is calculated as
\begin{equation}
    I_{C} (\mathbf{X}, \mathbf{Y}) = \frac{1}{N} \sum_{i=1}^N \left( \log p(y_i | x_i) - \frac{1}{N} \sum_{j=1}^N \log p(y_j | x_i) \right),
    \label{eq::method:club_estimation}
\end{equation}
where $\{(x_i, y_i)\}_{i=1}^N$ is the set of sampled pairs from $\mathbf{X}$ and $\mathbf{Y}$.
Eq.~(\ref{eq::method:club_estimation}) relies on the condition distribution $p(y | x)$, which is also unavailable, but fortunately, a neural network based variational distribution $q_{\theta} (y | x)$ has been proved to be substitutable for $p(y | x)$.
Specific to our model, a smaller information overlap is expected between $\mathbf{Z}^{\text{st}}$ and $\mathbf{Z}^{\text{he}}$, meaning that the upper bound estimation of mutual information should be constrained as
\begin{equation}
    \mathcal{L}_{mi} = \frac{1}{|\mathcal{V}|} \sum_{i \in \mathcal{V}} \left( \log q_{\theta} \left( \textbf{z}_i^{\text{he}} | \textbf{z}_i^{\text{st}} \right) - \frac{1}{|\mathcal{V}|} \sum_{j \in \mathcal{V}} \log q_{\theta} \left( \textbf{z}_j^{\text{he}} | \textbf{z}_i^{\text{st}} \right) \right),
    \label{eq::method:mi_loss}
\end{equation}
where $\textbf{z}_i^{\text{st}}$ and $\textbf{z}_i^{\text{he}}$ denote the embeddings of node $i$ in $\mathbf{Z}^{\text{st}}$ and $\mathbf{Z}^{\text{he}}$, separately.
We use $\mathcal{L}_{mi}$ to quantify the independence between $\mathbf{Z}^{\text{st}}$ and $\mathbf{Z}^{\text{he}}$.

Apart from independence, difference is another reflection of disentanglement.
Since our model is built upon the Poincaré ball, $\mathbf{Z}^{\text{st}}$ and $\mathbf{Z}^{\text{he}}$ are both numerically limited to be less than $\sqrt{1/c^L}$ (for ease of calculation, we set the curvatures of last layers in both HGCN$^{\mathbb{B}}$ and he-HGCN$^{\mathbb{B}}$ to be $c^L$) and difficult to distinguish by Euclidean distance.
Even though we use the Poincaré ball distance defined in Eq.~(\ref{eq::preli:dis_on_poincare}), it requires a large amount of computation to achieve the macro difference of the entire embeddings.
In this part, we leverage a simple but effective measurement: classification.
The embeddings of nodes in $\mathbf{Z}^{\text{st}}$ and $\mathbf{Z}^{\text{he}}$ are fed into a hyperbolic MLP discriminator and classified into two groups according to whether they are from $\mathbf{Z}^{\text{st}}$ or $\mathbf{Z}^{\text{he}}$.
The binary cross entropy loss is applied for the quantification of difference, denoted as $\mathcal{L}_{df}$.

The mutual information estimation and hyperbolic MLP discriminator make up the disentangling module.
This module does not directly participate in the data flow, but provides critical constraints in the training process for independence and difference between two embedding matrices.
We summarize the total objective function of disentangling module as
\begin{equation}
    \mathcal{L}_{dis} = \mathcal{L}_{mi} + \lambda_{dis} \cdot \mathcal{L}_{df},
    \label{eq::method:dis_loss}
\end{equation}
where $\lambda_{dis}$ is used to trade off the two components.
With these constraints, the disentanglement between $\mathbf{Z}^{\text{st}}$ and $\mathbf{Z}^{\text{he}}$ is expected to be enhanced.
$\mathbf{Z}^{\text{he}}$ is expected to be the pure semantic embedding matrix of input graph, renamed as $\mathbf{Z}^{\text{se}}$.

\subsection{Overall Framework and Model Training}  \label{sec::method:framework}

With above modules, the overall framework of our model \mymodel in Fig.~\ref{fig::method:dis-h2gcn} is introduced in this part.
For an input heterogeneous graph, we first eliminate the types of nodes and edges in structural information learning module.
The input node features are mapped into the hyperbolic space, and the HGCN$^{\mathbb{B}}$ is applied to obtain the pure structural embeddings $\mathbf{Z}^{\text{st}}$.
Simultaneously, heterogeneous graph contrastive learning module tries to mine all the information contained in the input graph and learns the heterogeneous embeddings $\mathbf{Z}^{\text{he}}$ through the contrastive training strategy based on he-HGCN$^{\mathbb{B}}$.
Then we disentangle the heterogeneous embeddings $\mathbf{Z}^{\text{he}}$ and structural embeddings $\mathbf{Z}^{\text{st}}$ in disentangling module to enhance the independence and difference between them, achieving the semantic embeddings $\mathbf{Z}^{\text{se}}$.
The final structural embeddings $\mathbf{Z}^{\text{st}}$ and semantic embeddings $\mathbf{Z}^{\text{se}}$ can either be used separately in downstream tasks, or collaborate through a simple fusion operation.
We adopt Möbius addition in Eq.~(\ref{eq::preli:addition}) for fusion as
\begin{equation}
    \mathbf{Z} = \mathbf{Z}^{\text{st}} \oplus_{c^L} \mathbf{Z}^{\text{se}},
    \label{eq::method:fusion}
\end{equation}
where $\mathbf{Z}^{\text{st}}$ and $\mathbf{Z}^{\text{se}}$ are with the curvature $- c^L$.
%

% \subsection{Model Training}  \label{sec::method:training}

\mymodel is trained with an end-to-end manner.
We take node classification and link prediction as downstream tasks.
For the former, the cross entropy loss function is used, as
\begin{equation}
    \mathcal{L}_{nc} (\mathbf{Z}^{*}) = \mathop{\text{Avg}}\limits_{i \in \mathcal{V}} \left( y_i \cdot \log (\hat{y}_i) \right),
    \label{eq::method:nc_loss}
\end{equation}
where $y_i$ and $\hat{y}_i$ are ground truth and predicted labels of node $i$, $\mathbf{Z}^{*}$ is the input embedding matrix.
For link prediction, to maximize the probability of connected nodes and minimize that of unconnected ones, the loss function is
\begin{equation}
    \mathcal{L}_{lp} (\mathbf{Z}^{*}) = \mathop{\text{Avg}}\limits_{(i, j) \in \mathcal{E}} \left( - \log (p_{i, j}) \right) + \mathop{\text{Avg}}\limits_{(i, j^{\prime}) \notin \mathcal{E}} \left( - \log (1 - p_{i, j^{\prime}}) \right),
    \label{eq::method:lp_loss}
\end{equation}
where $p_{i, j}$ is the predicted probability for the connection between node $i$ and $j$, and $(i, j^{\prime})$ is the negative sample used to accelerate training and prevent over-smoothing.
Taking all modules into account, the overall loss function is 
\begin{equation}
    \mathcal{L} = \mathcal{L}_{dt} (\mathbf{Z}) + \lambda_1 \cdot \left( \mathcal{L}_{dt} (\mathbf{Z}^{\text{st}}) + \mathcal{L}_{dt} (\mathbf{Z}^{\text{se}}) \right) + \lambda_2 \cdot \mathcal{L}_{dis} + \lambda_3 \cdot \mathcal{L}_{cl},
    \label{eq::method:loss}
\end{equation}
where $\lambda_1$, $\lambda_2$ and $\lambda_3$ are hyper-parameters.
$\mathcal{L}_{dt}$ switches between $\mathcal{L}_{nc}$ and $\mathcal{L}_{lp}$ with specific downstream task.
To improve the performance of each individual aspect of embeddings, we allow the structural embeddings $\mathbf{Z}^{\text{st}}$ and semantic embeddings $\mathbf{Z}^{\text{se}}$ to participate directly in the end-to-end training of downstream tasks, reflected in the second component of Eq.~(\ref{eq::method:loss}).

\subsection{Complexity Analysis}  \label{sec::method:complexity}

We analyse the time complexity of \mymodel by module in this section.
Regarding the structural information learning module, the computation complexity is $\mathcal{O} ( |\mathcal{V}|dd^{\prime} + |\mathcal{E}|d )$, where $d^{\prime}$ and $d$ are the dimensions of input features and output embeddings.
For heterogeneous graph contrastive learning module, the time complexity is $\mathcal{O} ( |\mathcal{V}|dd^{\prime} + |\mathcal{E}|d + |\mathcal{V}| |\mathcal{L}_{\mathcal{E}}|d + |\mathcal{V}|nd )$, where $n$ represents the number of negative samples for each positive pair in InfoNCE.
Practically edge type number $|\mathcal{L}_{\mathcal{E}}|$ is far smaller than the embedding dimension $d$, so the complexity can be reduced to $\mathcal{O} ( |\mathcal{V}|dd^{\prime} + |\mathcal{E}|d + |\mathcal{V}|nd )$.
The disentangling module takes the complexity of $\mathcal{O} ( |\mathcal{V}|n^{\prime}d^2 )$, where $n^{\prime}$ is the number of negative samples for each node when estimating the upper bound of mutual information in Eq.~(\ref{eq::method:mi_loss}).
But the disentangling module only participates in the model training, not in the inference process.

\section{Experiments}  \label{sec::exper}

In this section, we conduct extensive experiments on five real-world heterogeneous graphs to address the following research questions (\textbf{RQ}s).
\begin{itemize}[leftmargin=0em, itemindent=2em]
    \item \textbf{RQ1}: Can \mymodel achieve good performance on downstream tasks?
    \item \textbf{RQ2}: What do modules of \mymodel and the framework bring?
    \item \textbf{RQ3}: Are the structural and semantic features really disentangled by \mymodelend?
    \item \textbf{RQ4}: Are there any other benefits of disentanglement?
    \item \textbf{RQ5}: How do the key hyper-parameters influence the performance of \mymodelend?
\end{itemize}

\subsection{Experimental Setup}  \label{sec::exper:setup}

\subsubsection{Datasets}  \label{sec::exper:setup:dataset}

\begin{table}[tbp!]
    \centering
    \setlength{\tabcolsep}{2pt}
    \setlength{\abovecaptionskip}{0pt}
    \setlength{\belowcaptionskip}{0pt}
    \caption{Statistics of datasets.}
    \resizebox{\linewidth}{!}{
        \begin{tabular}{c|c|c|c|c|c|c|c|c}
        \toprule
        \multirow{2.5}{*}{\textbf{Dataset}} & \multicolumn{2}{c|}{\textbf{\#Nodes}} & \multicolumn{3}{c|}{\textbf{\#Edges}} & \multirow{2.5}{*}{\textbf{\#Features}} & \multirow{2.5}{*}{\textbf{\#Labels}} & \multirow{2.5}{*}{\textbf{$\delta$$^*$}} \\
        \cmidrule(lr){2-6}
        & \textbf{Type} & \textbf{\#} & \textbf{Type} & \textbf{\#} & \textbf{$\delta$$^*$} & & & \\
        \midrule
        IMDB & \makecell[c]{ \textbf{M}ovie \\ \textbf{D}irector \\ \textbf{A}ctor } & \makecell[c]{ 4,278 \\ 2,081 \\ 5,257 } & \makecell[c]{ M $\longleftrightarrow$ D \\ M $\longleftrightarrow$ A } & \makecell[c]{ 4,278 \\ 12,828 } & \makecell[c]{ - \\ 3.0 } & 1,524 & M: 3 & 3.0 \\
        \midrule
        DBLP & \makecell[c]{ \textbf{A}uthor \\ \textbf{P}aper \\ \textbf{T}erm \\ \textbf{V}enue } & \makecell[c]{ 4,057 \\ 14,328 \\ 7,723 \\ 20 } & \makecell[c]{ A $\longleftrightarrow$ P \\ P $\longleftrightarrow$ T \\ P $\longleftrightarrow$ V } & \makecell[c]{ 19,645 \\ 85,810 \\ 14,328 } & \makecell[c]{ 4.0 \\ 2.0 \\ 0.0 } & 4,635 & A: 4 & 2.0 \\
        \midrule
        Alibaba & \makecell[c]{ \textbf{U}ser \\ \textbf{I}tem } & \makecell[c]{ 6,054 \\ 16,595 } & \makecell[c]{ U $\stackrel{click}{\longleftrightarrow}$ I \\ U $\stackrel{enquiry}{\longleftrightarrow}$ I \\ U $\stackrel{contact}{\longleftrightarrow}$ I } & \makecell[c]{ 25,180 \\ 16,125 \\ 4,429 } & \makecell[c]{ 3.0 \\ 1.0 \\ 0.0} & 18 & I: 5 & 3.0 \\
        \midrule
        Alibaba-s & \makecell[c]{ \textbf{U}ser \\ \textbf{I}tem } & \makecell[c]{ 1,869 \\ 13,349 } & \makecell[c]{ U $\stackrel{click}{\longleftrightarrow}$ I \\ U $\stackrel{enquiry}{\longleftrightarrow}$ I \\ U $\stackrel{contact}{\longleftrightarrow}$ I } & \makecell[c]{ 15,404 \\ 9,475 \\ 2,157 } & \makecell[c]{ 5.0 \\ 3.0 \\ 0.0} & - & I: 5 & 4.0 \\
        \midrule
        Amazon & \makecell[c]{ \textbf{U} \\ \textbf{V} } & \makecell[c]{ 3,781 \\ 5,749 } & \makecell[c]{ U $\stackrel{type 0}{\longleftrightarrow}$ V \\ U $\stackrel{type 1}{\longleftrightarrow}$ V } & \makecell[c]{ 31,378 \\ 29,280 } & \makecell[c]{ 3.0 \\ 2.0 } & 4 & - & 2.0 \\
        \bottomrule
        \end{tabular}
    }
    \scriptsize
    \flushleft{{$^*$ $\delta$ refers to Gromov's hyperbolicity~\cite{gromovHyperbolicGroups}, employed to assess the tree-likeness and hierarchical properties of graphs. A lower $\delta$ indicates a more tree-like structure, with $\delta=0$ denoting a pure tree. All datasets exhibit implicit hierarchical structures and distinct power-law distributions over both the entire graphs and single edge types.}}
    \label{tab::expr:datasets}
    \vspace{-1em}
\end{table}

Five widely used real-world heterogeneous graphs are employed to verify the effectiveness of the proposals.
The statistics are given in TABLE~\ref{tab::expr:datasets}.
\textbf{IMDB}\footnote{\url{https://github.com/RuixZh/SR-RSC/tree/main/dataset/imdb}}~\cite{zhangSimpleMetapathfreeFramework2022} is a graph constructed with movie metadata, containing three node types (\textit{\textbf{M}ovie}, \textit{\textbf{D}irector} and \textit{\textbf{A}ctor}) and two edge types (\textit{M} $\longleftrightarrow$ \textit{D} and \textit{M} $\longleftrightarrow$ \textit{A}). 
The movies are categorized into three classes (\textit{action}, \textit{comedy} and \textit{drama}) based on their genres. 
Movie features are derived from the bag-of-words representation of their plots.
\textbf{DBLP}\footnote{\url{https://www.dropbox.com/s/yh4grpeks87ugr2/DBLP_processed.zip?dl=0}}~\cite{wangHeterogeneousGraphAttention2019} is an academic network, which contains four node types (\textit{\textbf{A}uthor}, \textit{\textbf{P}aper}, \textit{\textbf{T}erm}, \textit{\textbf{V}enue}) and three edge types (\textit{A} $\longleftrightarrow$ \textit{P}, \textit{P} $\longleftrightarrow$ \textit{T} and \textit{P} $\longleftrightarrow$ \textit{V}). 
The authors are categorized into four classes (\textit{database}, \textit{data mining}, \textit{machine learning} and \textit{information retrieval}) based on their research areas. 
Author features are derived from the bag-of-words representation of keywords.
\textbf{Alibaba} and \textbf{Alibaba-s}\footnote{\url{https://github.com/xuehansheng/DualHGCN/tree/main/data/Alibaba}}~\cite{xueMultiplexBipartiteNetwork2021} consist of behavior logs of users and items from the e-commerce. 
Both of them contain two node types (\textit{\textbf{U}ser} and \textit{\textbf{I}tem}) and three edge types (\textit{U} $\stackrel{click}{\longleftrightarrow}$ \textit{I}, \textit{U} $\stackrel{enquiry}{\longleftrightarrow}$ \textit{I} and \textit{U} $\stackrel{contact}{\longleftrightarrow}$ \textit{I}). 
The items are divided into five classes (\textit{women’s clothing}, \textit{men’s clothing}, \textit{etc.}) according to their categories. 
In Alibaba, the features are from the basic attributes of users and items. Alibaba-s is a smaller unfeatured dataset from the same source.
\textbf{Amazon}\footnote{\url{https://github.com/THUDM/GATNE/tree/master/data/amazon}}~\cite{cenRepresentationLearningAttributed2019} is constructed with product metadata of \textit{electronics} category and transformed into a bipartite graph. 
It has two node types and two edge types. The features are from the basic attributes of products.

\subsubsection{Baselines}  \label{sec::exper:setup:baseline}

\newcommand{\yes}{\raisebox{0.6ex}{\scalebox{0.7}{$\sqrt{}$}}}
\newcommand{\no}{\scalebox{0.85}[1]{$\times$}}
\newcommand{\halfyes}{\raisebox{0.6ex}{\scalebox{0.7}{$\sqrt{}\mkern-9mu{\smallsetminus}$}}}

\begin{table}[tbp!]
    \centering
    \setlength{\abovecaptionskip}{0pt}
    \setlength{\belowcaptionskip}{0pt}
    \caption{Baselines.}
    \resizebox{0.7\linewidth}{!}{
        \begin{tabular}{c|cc}
        \toprule
        \textbf{Methods} & \textbf{Heterogeneous} & \textbf{Hyperbolic} \\
        \midrule
        \textbf{Node2vec}~\cite{groverNode2vecScalableFeature2016} & \no & \no \\
        \textbf{GCN}~\cite{kipfSemiSupervisedClassificationGraph2017} & \no & \no \\
        \textbf{AM-GCN}~\cite{wangAMGCNAdaptiveMultichannel2020} & \no & \no \\
        \midrule
        \textbf{Metapath2vec}~\cite{dongMetapath2vecScalableRepresentation2017} & \yes & \no \\
        \textbf{DualHGCN}~\cite{xueMultiplexBipartiteNetwork2021} & \yes & \no \\
        \textbf{MHGCN}~\cite{yuMultiplexHeterogeneousGraph2022} & \yes & \no \\
        \textbf{SR-RSC}~\cite{zhangSimpleMetapathfreeFramework2022} & \yes & \no \\
        \textbf{BPHGNN}~\cite{fuMultiplexHeterogeneousGraph2023} & \yes & \no \\
        \midrule
        \textbf{Poincaré}~\cite{nickelPoincareEmbeddingsLearning} & \no & \yes \\
        \textbf{HGCN}~\cite{chamiHyperbolicGraphConvolutional} & \no & \yes \\
        \textbf{HHNE}~\cite{wangHyperbolicHeterogeneousInformation2019} & \yes & \yes \\
        \textbf{HAT}~\cite{zhangHyperbolicGraphAttention2021} & \no & \yes \\
        \textbf{QGCN}~\cite{xiongPseudoriemannianGraphConvolutional2022} & \no & \halfyes \\
        \bottomrule
        \end{tabular}
    }
    \label{tab::expr:baselines}
    \vspace{-1em}
\end{table}

We compare our model \mymodel with three homogeneous graph embedding methods, five heterogeneous graph embedding methods and five hyperbolic graph embedding methods, as shown in TABLE~\ref{tab::expr:baselines}. 
\textbf{Node2vec}~\cite{groverNode2vecScalableFeature2016} performs biased second-order random walks on graphs and then the skip-gram model is applied.
\textbf{GCN}~\cite{kipfSemiSupervisedClassificationGraph2017} is a semi-supervised graph convolutional network for node embedding in Euclidean spaces.
\textbf{AM-GCN}~\cite{wangAMGCNAdaptiveMultichannel2020} is an adaptive multi-channel graph convolutional networks, learning the correlations between topological structures and node features.
\textbf{Metapath2vec}~\cite{dongMetapath2vecScalableRepresentation2017} uses meta-paths to guide random walks and the skip-gram model to learn node embeddings.
\textbf{DualHGCN}~\cite{xueMultiplexBipartiteNetwork2021} models bipartite heterogeneous graphs with dual hypergraph convolutional networks.
\textbf{MHGCN}~\cite{yuMultiplexHeterogeneousGraph2022} integrates multi-relation structural signals and attribute semantics into embeddings for multiplex heterogeneous graphs.
\textbf{SR-RSC}~\cite{zhangSimpleMetapathfreeFramework2022} is a meta-path-free framework for heterogeneous graphs, which adopts multi-hop message passing to avoid the dependency on pre-set meta-paths.
\textbf{BPHGNN}~\cite{fuMultiplexHeterogeneousGraph2023} captures both local and global relevant information through depth and breadth behavior pattern aggregations on multiplex heterogeneous graphs.
\textbf{Poincaré}~\cite{nickelPoincareEmbeddingsLearning} preserves the proximity of node pairs via a hyperbolic version of skip-gram like model in the Poincaré ball.
\textbf{HGCN}~\cite{chamiHyperbolicGraphConvolutional} constructs the graph convolutional network in Lorentz model, another isometric manifold of hyperbolic spaces.
\textbf{HHNE}~\cite{wangHyperbolicHeterogeneousInformation2019} samples the heterogeneous graphs with meta-paths and optimizes the learning object by Riemannian stochastic gradient descent.
\textbf{HAT}~\cite{zhangHyperbolicGraphAttention2021} proposes the hyperbolic multi-head attention mechanism and redesigns the graph attention network in hyperbolic spaces.
\textbf{QGCN}~\cite{xiongPseudoriemannianGraphConvolutional2022} generalizes the graph convolutional network to pseudo-Riemannian manifolds of constant nonzero curvatures.

\begin{table*}[tbp!]
    \centering
    \setlength{\tabcolsep}{8pt}
    \setlength{\abovecaptionskip}{0pt}
    \setlength{\belowcaptionskip}{0pt}
    \caption{The performance of \mymodel and baselines for node classification (\%).}
    \resizebox{\linewidth}{!}{
        \begin{tabular}{c|cc|cc|cc|cc}
        \toprule
        \textbf{Dataset} & \multicolumn{2}{c|}{\textbf{IMDB}} & \multicolumn{2}{c|}{\textbf{DBLP}} & \multicolumn{2}{c|}{\textbf{Alibaba}} & \multicolumn{2}{c}{\textbf{Alibaba-s}} \\
        \textbf{Metric} & \textbf{Macro-F1} & \textbf{Micro-F1} & \textbf{Macro-F1} & \textbf{Micro-F1} & \textbf{Macro-F1} & \textbf{Micro-F1} & \textbf{Macro-F1} & \textbf{Micro-F1} \\
        \midrule
        \textbf{Node2vec} & $35.71 \pm 1.03$ & $37.33 \pm 1.21$ & $26.59 \pm 0.59$ & $29.03 \pm 1.10$ & $19.16 \pm 0.64$ & $27.22 \pm 0.40$ & $31.44 \pm 1.26$ & $33.80 \pm 1.08$ \\
        \textbf{GCN} & $58.39 \pm 0.82$ & $58.50 \pm 0.76$ & $76.12 \pm 0.83$ & $76.69 \pm 0.72$ & $24.02 \pm 0.93$ & $35.08 \pm 0.80$ & $29.25 \pm 1.60$ & $35.29 \pm 1.43$ \\
        \textbf{AM-GCN} & $63.80 \pm 0.78$ & $64.18 \pm 0.72$ & $91.79 \pm 0.50$ & $92.12 \pm 0.49$ & $57.54 \pm 0.88$ & $58.37 \pm 0.87$ & $63.01 \pm 0.37$ & $63.43 \pm 0.36$ \\
        \midrule
        \textbf{Metapath2vec} & $46.51 \pm 0.97$ & $47.74 \pm 0.83$ & $73.90 \pm 0.27$ & $74.07 \pm 0.27$ & $41.85 \pm 0.72$ & $43.44 \pm 0.44$ & $49.62 \pm 0.79$ & $50.61 \pm 0.69$ \\
        \textbf{DualHGNN} & - & - & - & - & $46.60 \pm 0.75$ & $39.61 \pm 1.49$ & $44.68 \pm 3.58$ & $42.79 \pm 3.89$ \\
        \textbf{MHGCN} & $55.88 \pm 1.14$ & $56.41 \pm 1.13$ & $73.75 \pm 0.99$ & $73.58 \pm 0.91$ & $18.27 \pm 0.69$ & $27.30 \pm 0.71$ & $44.06 \pm 0.89$ & $46.60 \pm 0.69$ \\
        \textbf{SR-RSC} & $54.96 \pm 1.72$ & $55.46 \pm 1.59$ & $73.04 \pm 0.87$ & $72.88 \pm 0.88$ & $24.69 \pm 3.25$ & $33.35 \pm 2.90$ & $39.76 \pm 2.15$ & $42.40 \pm 2.20$ \\
        \textbf{BPHGNN} & $56.30 \pm 1.04$ & $56.56 \pm 1.00$ & $74.28 \pm 0.70$ & $74.16 \pm 0.68$ & $11.64 \pm 2.55$ & $26.28 \pm 0.10$ & $42.20 \pm 1.20$ & $44.61 \pm 1.12$ \\
        \midrule
        \textbf{Poincare} & $46.84 \pm 0.76$ & $47.87 \pm 0.90$ & $80.78 \pm 0.44$ & $81.05 \pm 0.41$ & $40.91 \pm 1.87$ & $43.22 \pm 1.09$ & $52.79 \pm 0.81$ & $53.62 \pm 0.73$ \\
        \textbf{HGCN} & $64.64 \pm 0.86$ & $65.12 \pm 0.78$ & $80.93 \pm 1.09$ & $81.88 \pm 1.03$ & $35.19 \pm 0.96$ & $43.23 \pm 1.27$ & $36.09 \pm 1.06$ & $41.85 \pm 1.45$ \\
        \textbf{HHNE} & $53.60 \pm 1.07$ & $54.87 \pm 1.04$ & $85.20 \pm 0.69$ & $85.82 \pm 0.64$ & $53.27 \pm 0.92$ & $54.49 \pm 0.71$ & $56.75 \pm 1.14$ & $57.95 \pm 1.04$ \\
        \textbf{HAT} & $62.52 \pm 0.74$ & $62.72 \pm 0.80$ & $82.86 \pm 0.99$ & $83.22 \pm 1.03$ & $33.72 \pm 1.27$ & $42.36 \pm 0.76$ & $38.04 \pm 1.26$ & $44.01 \pm 0.75$ \\
        \textbf{QGCN} & $60.38 \pm 1.11$ & $60.74 \pm 0.95$ & $78.17 \pm 1.77$ & $78.84 \pm 1.59$ & $20.21 \pm 0.61$ & $30.45 \pm 0.85$ & $18.33 \pm 1.02$ & $27.94 \pm 1.26$ \\
        \midrule
        \textbf{\mymodel} & \bm{$70.07 \pm 0.36$} & \bm{$70.05 \pm 0.33$} & \bm{$96.33 \pm 0.45$} & \bm{$96.45 \pm 0.41$} & \bm{$62.66 \pm 1.44$} & \bm{$65.09 \pm 1.63$} & \bm{$73.46 \pm 0.49$} & \bm{$74.85 \pm 0.41$} \\
        \textbf{Gain(\%)} & \textbf{+8.39} & \textbf{+7.57} & \textbf{+4.94} & \textbf{+4.70} & \textbf{+8.89} & \textbf{+11.52} & \textbf{+16.59} & \textbf{+18.01} \\
        \textbf{-structural} & $65.32 \pm 0.38$ & $65.73 \pm 0.35$ & $75.53 \pm 9.22$ & $81.47 \pm 8.24$ & $52.60 \pm 1.32$ & $55.19 \pm 1.12$ & $29.76 \pm 0.42$ & $37.14 \pm 0.15$ \\
        \textbf{-semantic} & $67.35 \pm 0.37$ & $67.55 \pm 0.36$ & $95.34 \pm 0.27$ & $95.53 \pm 0.24$ & $60.72 \pm 1.75$ & $63.11 \pm 1.59$ & $67.25 \pm 0.36$ & $67.91 \pm 0.49$ \\
        \bottomrule
        \end{tabular}
    }
    \label{tab::expr:nc}
    \vspace{-0.5em}
\end{table*}

\begin{table*}[tbp!]
    \centering
    \setlength{\tabcolsep}{1.5pt}
    \setlength{\abovecaptionskip}{0pt}
    \setlength{\belowcaptionskip}{0pt}
    \caption{The performance of \mymodel and baselines for link prediction (\%).}
    \resizebox{\linewidth}{!}{
        \begin{tabular}{c|cc|cc|cc|cc|cc}
        \toprule
        \textbf{Dataset} & \multicolumn{2}{c|}{\textbf{IMDB}} & \multicolumn{2}{c|}{\textbf{DBLP}} & \multicolumn{2}{c|}{\textbf{Alibaba}} & \multicolumn{2}{c|}{\textbf{Alibaba-s}} & \multicolumn{2}{c}{\textbf{Amazon}} \\
        \textbf{Metric} & \textbf{AUC} & \textbf{AP} & \textbf{AUC} & \textbf{AP} & \textbf{AUC} & \textbf{AP} & \textbf{AUC} & \textbf{AP} & \textbf{AUC} & \textbf{AP} \\
        \midrule
        \textbf{Node2vec} & $48.70 \pm 0.47$ & $50.26 \pm 0.38$ & $51.18 \pm 0.13$ & $50.22 \pm 0.11$ & $55.01 \pm 0.15$ & $55.15 \pm 0.13$ & $62.54 \pm 0.34$ & $63.15 \pm 0.49$ & $47.56 \pm 0.19$ & $48.12 \pm 0.15$ \\
        \textbf{GCN} & $62.15 \pm 0.45$ & $64.73 \pm 0.32$ & $84.57 \pm 0.10$ & $88.03 \pm 0.04$ & $63.53 \pm 0.07$ & $69.91 \pm 0.40$ & $68.19 \pm 0.07$ & $72.89 \pm 0.71$ & $68.10 \pm 0.20$ & $72.67 \pm 0.40$ \\
        \textbf{AM-GCN} & $59.17 \pm 1.55$ & $64.10 \pm 0.43$ & $83.34 \pm 1.07$ & $86.91 \pm 0.68$ & $68.70 \pm 3.37$ & $73.56 \pm 1.88$ & $81.14 \pm 1.98$ & $79.26 \pm 2.13$ & $77.47 \pm 1.69$ & $77.18 \pm 1.38$ \\
        \midrule
        \textbf{Metapath2vec} & $74.40 \pm 0.24$ & $71.82 \pm 0.36$ & $68.40 \pm 0.11$ & $71.05 \pm 0.11$ & $72.43 \pm 0.09$ & $69.97 \pm 0.16$ & $73.50 \pm 0.18$ & $72.43 \pm 0.12$ & $77.28 \pm 0.03$ & $76.18 \pm 0.06$ \\
        \textbf{DualHGNN} & - & - & - & - & $73.57 \pm 1.24$ & $75.91 \pm 0.91$ & $66.67 \pm 4.04$ & $68.51 \pm 2.77$ & $84.16 \pm 2.26$ & $84.48 \pm 1.69$ \\
        \textbf{MHGCN} & $84.02 \pm 2.05$ & $81.62 \pm 2.41$ & $86.21 \pm 4.93$ & $88.96 \pm 3.13$ & $85.77 \pm 6.36$ & $80.10 \pm 4.83$ & $93.54 \pm 0.06$ & $92.78 \pm 0.08$ & $85.48 \pm 1.07$ & $78.35 \pm 1.77$ \\
        \textbf{SR-RSC} & $82.57 \pm 2.44$ & $80.18 \pm 2.71$ & $90.70 \pm 1.29$ & $92.57 \pm 1.35$ & $86.23 \pm 1.48$ & $80.02 \pm 1.04$ & $90.20 \pm 1.83$ & $89.30 \pm 2.15$ & $78.50 \pm 1.54$ & $70.99 \pm 1.56$ \\
        \textbf{BPHGNN} & $83.76 \pm 1.19$ & $80.83 \pm 1.48$ & $91.60 \pm 1.65$ & $92.65 \pm 1.14$ & $88.77 \pm 0.72$ & $82.41 \pm 1.04$ & $93.41 \pm 0.42$ & $92.60 \pm 0.41$ & $86.30 \pm 2.22$ & $79.74 \pm 3.01$ \\
        \midrule
        \textbf{Poincare} & $64.15 \pm 1.42$ & $63.56 \pm 1.46$ & $90.73 \pm 0.48$ & $91.01 \pm 0.42$ & $71.58 \pm 1.17$ & $71.60 \pm 1.11$ & $83.02 \pm 1.35$ & $80.55 \pm 1.45$ & $77.78 \pm 0.23$ & $77.40 \pm 0.26$ \\
        \textbf{HGCN} & $85.10 \pm 1.65$ & $81.36 \pm 2.57$ & $93.29 \pm 0.87$ & $92.22 \pm 0.78$ & $88.67 \pm 0.72$ & $85.66 \pm 1.08$ & $89.97 \pm 2.95$ & $81.53 \pm 4.77$ & $84.42 \pm 0.64$ & $73.14 \pm 1.09$ \\
        \textbf{HHNE} & $81.83 \pm 0.54$ & $83.79 \pm 0.43$ & $89.90 \pm 2.11$ & $90.42 \pm 1.96$ & $87.66 \pm 0.83$ & $88.00 \pm 0.85$ & $88.67 \pm 0.37$ & $88.78 \pm 0.35$ & $85.89 \pm 0.23$ & $84.45 \pm 0.44$ \\
        \textbf{HAT} & $83.62 \pm 0.49$ & $84.87 \pm 0.55$ & $91.69 \pm 0.12$ & $90.60 \pm 1.55$ & $81.89 \pm 1.18$ & $82.46 \pm 1.26$ & $83.58 \pm 1.00$ & $82.36 \pm 0.51$ & $72.51 \pm 2.23$ & $71.34 \pm 2.52$ \\
        \textbf{QGCN} & $64.46 \pm 2.06$ & $68.32 \pm 3.28$ & $89.59 \pm 1.21$ & $90.60 \pm 1.21$ & $81.03 \pm 2.04$ & $78.89 \pm 1.93$ & $85.07 \pm 5.51$ & $79.24 \pm 4.95$ & $87.79 \pm 3.75$ & $86.00 \pm 3.86$ \\
        \midrule
        \textbf{\mymodel} & \bm{$96.69 \pm 0.44$} & \bm{$94.42 \pm 0.52$} & \bm{$95.10 \pm 0.08$} & \bm{$94.92 \pm 0.08$} & \bm{$96.95 \pm 0.45$} & \bm{$94.18 \pm 0.69$} & \bm{$97.59 \pm 0.17$} & \bm{$96.88 \pm 0.34$} & \bm{$95.02 \pm 0.60$} & \bm{$92.95 \pm 0.44$} \\
        \textbf{Gain(\%)} & \textbf{+13.62} & \textbf{+11.25} & \textbf{+1.94} & \textbf{+2.46} & \textbf{+9.22} & \textbf{+7.02} & \textbf{+4.33} & \textbf{+4.42} & \textbf{+8.23} & \textbf{+8.08} \\
        \textbf{-structural} & $89.28 \pm 0.80$ & $85.64 \pm 1.79$ & $91.05 \pm 1.13$ & $91.20 \pm 0.92$ & $92.54 \pm 0.71$ & $89.97 \pm 0.36$ & $95.66 \pm 0.45$ & $94.83 \pm 0.60$ & $91.58 \pm 1.16$ & $88.74 \pm 0.73$ \\
        \textbf{-semantic} & $92.67 \pm 0.54$ & $89.87 \pm 0.44$ & $94.06 \pm 0.26$ & $93.16 \pm 0.42$ & $93.79 \pm 0.43$ & $91.64 \pm 0.64$ & $96.92 \pm 0.24$ & $96.18 \pm 0.22$ & $93.04 \pm 0.52$ & $89.92 \pm 0.86$ \\
        \bottomrule
        \end{tabular}
    }
    \label{tab::expr:lp}
    \vspace{-1em}
\end{table*}

\subsubsection{Experimental Details}  \label{sec::exper:setup:detail}

For our proposed \mymodelend, the hyper-parameters are set as follows: the layer number of HGCN$^{\mathbb{B}}$ and he-HGCN$^{\mathbb{B}}$ $L=2$, the number of negative samples in InfoNCE and mutual information estimation $n=n^{\prime}=10$, $\lambda_{dis}=0.5$, $\lambda_1=0.2$, $\lambda_2=0.5$ and $\lambda_3=0.05$.
All involved curvatures are initialized as -1 and keep trainable during the training process.
We take the recommended hyper-parameter settings of baselines unless otherwise noted.
For meta-path based baselines, we use the following meta-paths in each dataset: \{\textit{AMA}, \textit{MAM}, \textit{MDM}, \textit{AMDMA}\} in IMDB, \{\textit{APA}, \textit{PAP}, \textit{APVPA}, \textit{APTPA}, \textit{TPAPT}\} in DBLP, \{\textit{UIU}, \textit{IUI}\} in Alibaba and Alibaba-s, \{\textit{UVU}, \textit{VUV}\} in Amazon.
The method DualHGNN is proposed for bipartite graphs, so only experiments on Alibaba, Alibaba-s and Amazon are run on it.
The dataset Amazon is not applicable on node classification task, because it has no label on either node type.
In node classification and link prediction tasks, the embedding dimension is set to 32 for all methods including baselines and \mymodelend.
The classifier and predictor are both two-layer MLPs.
In node classification task, all the nodes and edges can be seen during the training process, and for the labels of classified nodes, we take 50\% as training set, 25\% as validation set and 25\% as test set.
In link prediction task, the edges are divided into training set, validation set and test set according to the ratio of 50\%, 25\% and 25\%, equally proportional for various edge types, and only the training set is seen in training.
All graph neural network models leverage early stop strategy via the validation set.
The experiments are conducted on a server with 2 $\times$ Intel Xeon Gold 6226R 16C 2.90 GHz CPUs, 256 GB Memory and 4 $\times$ GeForce RTX 3090 GPUs.

\subsection{Performance on Downstream Tasks \textup{(\textbf{RQ1})}}  \label{sec::exper:task}

We evaluate our proposed model \mymodel and baselines on two traditional downstream tasks node classification and link prediction to answer RQ1.
In node classification task, macro and micro averages of F1 scores (Macro-F1, Micro-F1) are used for the quantification of experimental performance.
In link prediction task, average precision (AP) and area under ROC curve (AUC) are used.
To avoid random errors, each experiment is repeated five times.
TABLE~\ref{tab::expr:nc} and~\ref{tab::expr:lp} report the average values and standard deviations for the experiments.

\subsubsection{Node Classification}  \label{sec::exper:task:nc}

The results of node classification are shown in TABLE~\ref{tab::expr:nc}. 
Obviously that \mymodel achieves the best performance in all cases. 
In addition, based on the observation of the results, we have the following analysis.
\begin{itemize}[leftmargin=0em, itemindent=2em]
    \item The results report the performance of disentangled structural and semantic embeddings. For most cases, the semantic embeddings keep a huge advantage over the structural ones, meaning that the semantic information is generally more effective for node classification in heterogeneous graphs. But this advantage can be offset by well-designed model framework, e.g. the homogeneous method AM-GCN, which achieves the second best results in this evaluation task.
    \item The overall performance of hyperbolic models is better than that of Euclidean models. TABLE~\ref{tab::expr:datasets} shows the hyperbolicity $\delta$ of entire graphs and edge types, and all used datasets obey relatively obvious power-law distributions. It turns out that embedding graphs into hyperbolic spaces does match the unbalanced distributions of them.
    \item Compared with Alibaba, the performance of most methods increases on Alibaba-s. Only a few of methods get worse results, including the structural embedding of \mymodelend. The main difference between the two datasets is that Alibaba-s has no node attributes. It suggests that the input node features are sometimes beneficial to node classification and sometimes not, regardless of the dataset size.
\end{itemize}

\subsubsection{Link Prediction}  \label{sec::exper:task:lp}

TABLE~\ref{tab::expr:lp} shows the results of link prediction task.
As we can see, \mymodel performs consistently better than the baselines.
Further analysis on the results draws the following conclusions.
\begin{itemize}[leftmargin=0em, itemindent=2em]
    \item The overall performance of heterogeneous methods is much better than that of homogeneous methods, which proves that the semantics of heterogeneous graphs is of great importance in this task. Partiality on topological structures will cause a great loss of effective information.
    \item Compared with node classification, the performance gap between the disentangled structural and semantic embeddings of \mymodel is significantly smaller in link prediction. The observation tells that pure structural embeddings can still make comparable results with appropriate constraints despite the importance of semantics.
    \item Similar to node classification, hyperbolic models performs better than Euclidean models in link prediction, reflected in the Euclidean and hyperbolic homogeneous models or congeneric heterogeneous models, e.g. Metapath2vec and HHNE. The superiority of hyperbolic geometry on power-law distributed data is verified again.
\end{itemize}

\subsection{Variant Study \textup{(\textbf{RQ2})}}  \label{sec::exper:variant}

For RQ2, we conduct the following variant study to asses the effectiveness modules of \mymodel and the framework.
\begin{itemize}[leftmargin=0em, itemindent=2em]
    \item \textbf{w/o st}: \mymodel without structural information learning module, accordingly not having disentangling module.
    \item \textbf{w/o he}: without heterogeneous graph contrastive learning module, accordingly not having disentangling module.
    \item \textbf{w/o cl}: \mymodel without contrastive learning in heterogeneous graph contrastive learning module.
    \item \textbf{w/o $\mathbb{B}$}: \mymodel without hyperbolic geometry, implemented by constructing the model in Euclidean spaces.
    \item \textbf{w/ cl$^{\prime}$}: a variant of \mymodelend, applying contrastive learning strategy on the output of HGCN$^{\mathbb{B}}$ and he-HGCN$^{\mathbb{B}}$.
\end{itemize}

As for the analysis of disentangling module, please refer to Section~\ref{sec::exper:eval_disentangle} and~\ref{sec::exper:benefit_disentangle}.

We take the same experimental setup with \mymodel on the variants, and the result are reported in TABLE~\ref{tab::expr:nc_variant} and~\ref{tab::expr:lp_variant}.
It can be seen that the performance of the variants drops noticeably on both tasks.
What is worthy noting is that the degradation for w/o he is much severer than that for w/o st, which indicates that the semantic information is more important in most cases, mapping the analysis in Section~\ref{sec::exper:task}.
The results for w/o cl prove the powerful ability of contrastive learning for mining the endogenous information in heterogeneous graphs themselves.
The descending of the performance for w/o $\mathbb{B}$ shows the advantage of hyperbolic geometry compared with Euclidean spaces in modeling data with power-law distributions.
The variant w/ cl$^{\prime}$ is used to verify the superiority of the framework.
It takes a different data flow in model, but keeps almost all the components of \mymodel except the disentangling module.
The results on it also indirectly demonstrate the necessity of disentangling module.

\begin{table*}[tbp!]
    \centering
    \setlength{\tabcolsep}{8.5pt}
    \setlength{\abovecaptionskip}{0pt}
    \setlength{\belowcaptionskip}{0pt}
    \caption{The performance of \mymodel and variants for node classification (\%).}
    \resizebox{\linewidth}{!}{
        \begin{tabular}{c|cc|cc|cc|cc}
        \toprule
        \textbf{Dataset} & \multicolumn{2}{c|}{\textbf{IMDB}} & \multicolumn{2}{c|}{\textbf{DBLP}} & \multicolumn{2}{c|}{\textbf{Alibaba}} & \multicolumn{2}{c}{\textbf{Alibaba-s}} \\
        \textbf{Metric} & \textbf{Macro-F1} & \textbf{Micro-F1} & \textbf{Macro-F1} & \textbf{Micro-F1} & \textbf{Macro-F1} & \textbf{Micro-F1} & \textbf{Macro-F1} & \textbf{Micro-F1} \\
        \midrule
        \textbf{\mymodel} & $70.07 \pm 0.36$ & $70.05 \pm 0.33$ & $96.33 \pm 0.45$ & $96.45 \pm 0.41$ & $62.66 \pm 1.44$ & $65.09 \pm 1.63$ & $73.46 \pm 0.49$ & $74.85 \pm 0.41$ \\
        \midrule
        \textbf{w/o st} & $69.34 \pm 0.42$ & $69.27 \pm 0.39$ & $95.07 \pm 0.25$ & $95.30 \pm 0.22$ & $61.15 \pm 0.92$ & $63.43 \pm 1.04$ & $67.73 \pm 0.42$ & $68.86 \pm 0.38$ \\
        %
        % \textbf{Gain(\%)} & -1.04 & -1.11 & -1.31 & -1.19 & -2.42 & -2.55 & -7.80 & -8.00 \\
        % %
        % \midrule
        %
        \textbf{w/o he} & $67.00 \pm 0.35$ & $67.32 \pm 0.35$ & $76.97 \pm 9.92$ & $83.45 \pm 8.28$ & $51.42 \pm 1.18$ & $54.09 \pm 0.53$ & $38.23 \pm 0.70$ & $46.28 \pm 0.81$ \\
        %
        % \textbf{Gain(\%)} & -4.38 & -3.90 & -20.10 & -13.48 & -17.94 & -16.90 & -47.95 & -38.18 \\
        % %
        % \midrule
        %
        \textbf{w/o cl} & $68.19 \pm 0.40$ & $68.18 \pm 0.37$ & $93.72 \pm 0.38$ & $94.01 \pm 0.38$ & $58.82 \pm 0.68$ & $61.33 \pm 1.21$ & $65.32 \pm 0.24$ & $66.32 \pm 0.51$ \\
        %
        % \textbf{Gain(\%)} & -2.67 & -2.66 & -2.71 & -2.53 & -6.13 & -5.78 & -11.09 & -11.40 \\
        % %
        % \midrule
        %
        \textbf{w/o $\mathbb{B}$} & $60.47 \pm 1.36$ & $61.66 \pm 1.10$ & $88.75 \pm 1.52$ & $88.50 \pm 0.98$ & $46.44 \pm 2.21$ & $47.83 \pm 2.54$ & $55.67 \pm 1.08$ & $57.50 \pm 0.99$ \\
        %
        % \textbf{Gain(\%)} & -13.70 & -11.97 & -7.87 & -8.24 & -25.89 & -26.52 & -24.22 & -23.18 \\
        % %
        % \midrule
        %
        \textbf{w/ cl$^{\prime}$} & $68.18 \pm 0.24$ & $68.23 \pm 0.25$ & $94.24 \pm 0.44$ & $94.53 \pm 0.43$ & $58.11 \pm 2.29$ & $59.84 \pm 2.69$ & $69.08 \pm 0.17$ & $69.81 \pm 0.24$ \\
        %
        % \textbf{Gain(\%)} & -2.70 & -2.60 & -2.17 & -1.99 & -7.26 & -8.07 & -5.97 & -6.73 \\
        %
        \bottomrule
        \end{tabular}
    }
    \label{tab::expr:nc_variant}
    \vspace{-0.5em}
\end{table*}

\begin{table*}[tbp!]
    \centering
    \setlength{\tabcolsep}{3pt}
    \setlength{\abovecaptionskip}{0pt}
    \setlength{\belowcaptionskip}{0pt}
    \caption{The performance of \mymodel and variants for link prediction (\%).}
    \resizebox{\linewidth}{!}{
        \begin{tabular}{c|cc|cc|cc|cc|cc}
        \toprule
        \textbf{Dataset} & \multicolumn{2}{c|}{\textbf{IMDB}} & \multicolumn{2}{c|}{\textbf{DBLP}} & \multicolumn{2}{c|}{\textbf{Alibaba}} & \multicolumn{2}{c|}{\textbf{Alibaba-s}} & \multicolumn{2}{c}{\textbf{Amazon}} \\
        \textbf{Metric} & \textbf{AUC} & \textbf{AP} & \textbf{AUC} & \textbf{AP} & \textbf{AUC} & \textbf{AP} & \textbf{AUC} & \textbf{AP} & \textbf{AUC} & \textbf{AP} \\
        \midrule
        \textbf{\mymodel} & $96.69 \pm 0.44$ & $94.42 \pm 0.52$ & $95.10 \pm 0.08$ & $94.92 \pm 0.08$ & $96.95 \pm 0.45$ & $94.18 \pm 0.69$ & $97.59 \pm 0.17$ & $96.88 \pm 0.34$ & $95.02 \pm 0.60$ & $92.95 \pm 0.44$ \\
        \midrule
        \textbf{w/o st} & $94.51 \pm 0.85$ & $91.73 \pm 1.18$ & $91.24 \pm 0.84$ & $88.92 \pm 1.35$ & $93.88 \pm 0.35$ & $91.79 \pm 0.58$ & $94.72 \pm 0.10$ & $92.17 \pm 0.26$ & $92.94 \pm 0.22$ & $89.54 \pm 0.40$ \\
        %
        % \textbf{Gain(\%)} & -2.26 & -2.85 & -4.05 & -6.32 & -3.17 & -2.54 & -2.94 & -4.86 & -2.19 & -3.66 \\
        % %
        % \midrule
        %
        \textbf{w/o he} & $86.66 \pm 0.94$ & $81.57 \pm 0.91$ & $90.38 \pm 0.53$ & $89.62 \pm 2.06$ & $90.79 \pm 1.16$ & $87.58 \pm 1.67$ & $94.49 \pm 0.67$ & $93.69 \pm 0.65$ & $90.43 \pm 1.38$ & $87.93 \pm 1.31$ \\
        %
        % \textbf{Gain(\%)} & -10.37 & -13.61 & -4.96 & -5.59 & -6.36 & -7.01 & -3.17 & -3.29 & -4.82 & -5.40 \\
        % %
        % \midrule
        %
        \textbf{w/o cl} & $88.46 \pm 1.02$ & $86.00 \pm 0.72$ & $90.30 \pm 0.71$ & $87.67 \pm 1.30$ & $93.26 \pm 0.33$ & $90.90 \pm 0.42$ & $93.15 \pm 0.21$ & $91.65 \pm 0.46$ & $91.43 \pm 0.19$ & $87.70 \pm 0.29$ \\
        %
        % \textbf{Gain(\%)} & -8.52 & -8.92 & -5.05 & -7.64 & -3.81 & -3.49 & -4.55 & -5.40 & -3.78 & -5.64 \\
        % %
        % \midrule
        %
        \textbf{w/o $\mathbb{B}$} & $83.30 \pm 1.47$ & $80.71 \pm 1.31$ & $82.48 \pm 1.94$ & $85.30 \pm 0.82$ & $75.64 \pm 1.45$ & $75.92 \pm 1.79$ & $76.94 \pm 1.24$ & $76.87 \pm 1.45$ & $85.63 \pm 2.67$ & $84.68 \pm 2.66$ \\
        %
        % \textbf{Gain(\%)} & -13.85 & -14.52 & -13.27 & -10.14 & -21.98 & -19.39 & -21.16 & -20.66 & -9.88 & -8.89 \\
        % %
        % \midrule
        %
        \textbf{w/ cl$^{\prime}$} & $94.76 \pm 0.37$ & $92.19 \pm 0.59$ & $93.55 \pm 0.49$ & $93.83 \pm 0.31$ & $92.82 \pm 1.12$ & $89.77 \pm 1.58$ & $94.91 \pm 0.40$ & $93.79 \pm 0.94$ & $92.63 \pm 0.21$ & $90.16 \pm 0.12$ \\
        %
        % \textbf{Gain(\%)} & -2.00 & -2.37 & -1.63 & -1.15 & -4.26 & -4.68 & -2.75 & -3.19 & -2.51 & -2.99 \\
        %
        \bottomrule
        \end{tabular}
    }
    \label{tab::expr:lp_variant}
    \vspace{-1em}
\end{table*}

\subsection{Evaluation of Disentanglement \textup{(\textbf{RQ3})}}  \label{sec::exper:eval_disentangle}

\begin{figure}
    \centering
    \includegraphics[width=0.5\linewidth]{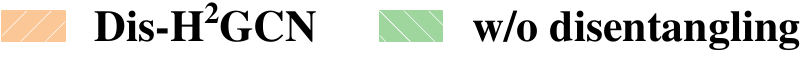} \\
    \subfloat[Node classification.]{
        \includegraphics[width=0.41\linewidth]{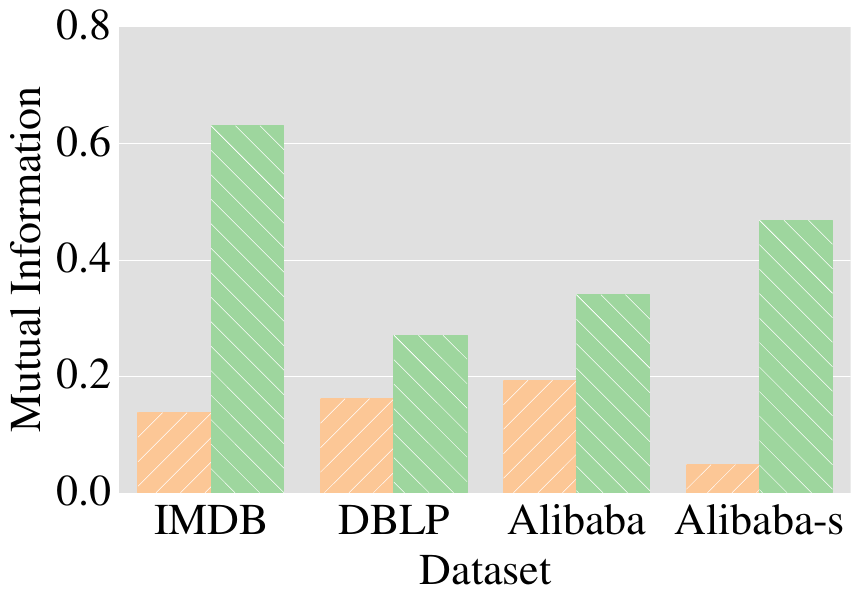}
        \label{fig::expr:mutual_information:nc}
    }
    \subfloat[Link prediction.]{
        \includegraphics[width=0.50\linewidth]{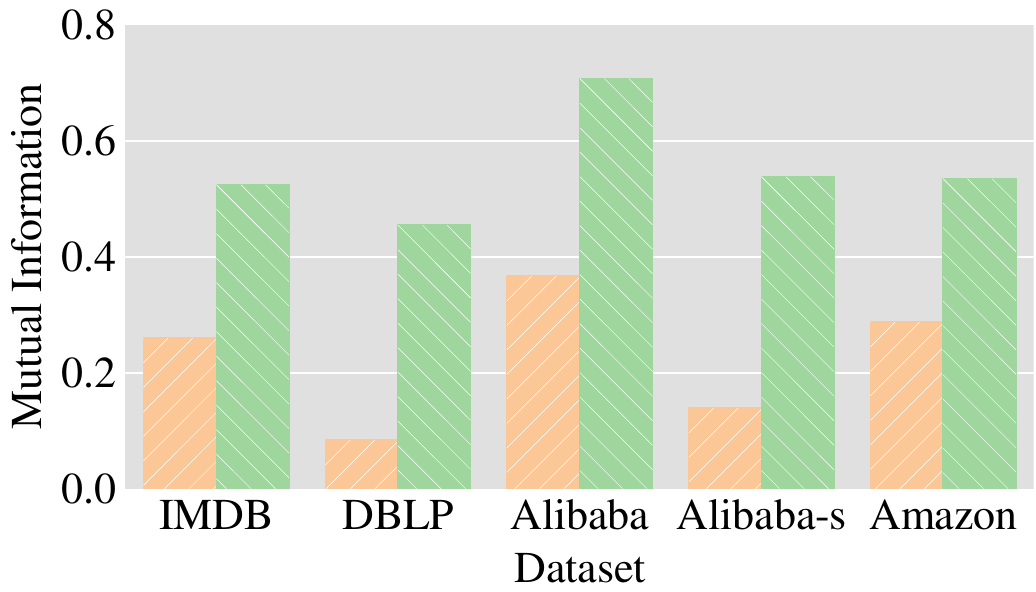}
        \label{fig::expr:mutual_information:lp}
    }
    \caption{Mutual information of our model and -w/o disentangling on both downstream tasks.}
    \label{fig::expr:mutual_information}
    \vspace{-1em}
\end{figure}

In this section, we answer RQ3 via quantitatively evaluating the extent of disentanglement.
A new variant of our \mymodel is constructed by removing disentangling module, named as \textbf{w/o disentangling}.
We calculate the mutual information between the structural embeddings and the semantic embeddings for both \mymodel and w/o disentangling to measure the influence of this module.
As Fig.~\ref{fig::expr:mutual_information} illustrates, the participation of disentangling module markedly decreases the mutual information between structural and semantic embeddings, succeeding in make a higher extent of disentanglement.

We make a more intuitive explanation through further experiments.
Generally, node types are the reflection of semantic information and are supposed to be more related to semantic features in heterogeneous graphs.
To attest to this point, we conduct node type classification tasks on all five datasets by treating the node types as predefined labels, measured with the same metrics as node classification in Section~\ref{sec::exper:task:nc}.
The results are shown in TABLE~\ref{tab::expr:tc}.
Obviously the semantic embeddings have much better performance than structural ones and shows a lot more correlations with semantic information in the graphs.
For the dataset Alibaba-s, the structural embeddings exhibit a huge performance gap between the two metrics Macro-F1 and Micro-F1.
This is because that the two node types in the dataset are extremely unbalanced and classifying all nodes into the major group will lead to a high Micro-F1, but leave the Macro-F1 a value close to 0.5.

\subsection{Benefits of Disentanglement \textup{(\textbf{RQ4})}}  \label{sec::exper:benefit_disentangle}

% \begin{table*}[tbp!]
%     \centering
%     \setlength{\tabcolsep}{8pt}
%     \setlength{\abovecaptionskip}{0pt}
%     \setlength{\belowcaptionskip}{0pt}
%     \caption{The performance of disentangled representations on node type classification (\%).}
%     % \renewcommand{\arraystretch}{1.15}
%     \resizebox{\linewidth}{!}{
%         \begin{tabular}{c|cc|cc|cc|cc|cc}
%         \toprule
%         %
%         \textbf{Dataset} & \multicolumn{2}{c|}{\textbf{IMDB}} & \multicolumn{2}{c|}{\textbf{DBLP}} & \multicolumn{2}{c|}{\textbf{Alibaba}} & \multicolumn{2}{c|}{\textbf{Alibaba-s}} & \multicolumn{2}{c}{\textbf{Amazon}} \\
%         %
%         \textbf{Metric} & \textbf{Macro-F1} & \textbf{Micro-F1} & \textbf{Macro-F1} & \textbf{Micro-F1} & \textbf{Macro-F1} & \textbf{Micro-F1} & \textbf{Macro-F1} & \textbf{Micro-F1} & \textbf{Macro-F1} & \textbf{Micro-F1} \\
%         %
%         \midrule
%         %
%         \textbf{\mymodel} & 91.35 & 93.67 & 70.32 & 97.19 & 83.94 & 79.87 & 79.02 & 94.15 & 94.89 & 95.64 \\
%         %
%         \midrule
%         %
%         \textbf{-structural} & 44.82 & 53.17 & 58.28 & 64.14 & 40.20 & 63.06 & 51.58 & 91.26 & 63.78 & 73.91 \\
%         %
%         \textbf{-semantic} & 96.75 & 98.31 & 72.95 & 98.89 & 82.11 & 79.34 & 80.16 & 96.39 & 92.45 & 94.96 \\
%         %
%         \bottomrule
%         \end{tabular}
%     }
%     \label{tab::expr:tc}
%     \vspace{-1em}
% \end{table*}

\begin{table}[tbp!]
    \centering
    \setlength{\tabcolsep}{2pt}
    \setlength{\abovecaptionskip}{0pt}
    \setlength{\belowcaptionskip}{0pt}
    \caption{The performance of disentangled representations on node type classification (\%).}
    \resizebox{\linewidth}{!}{
        \begin{tabular}{c|cc|cc|cc}
        \toprule
        \multirow{2}{*}{\textbf{Dataset}} & \multicolumn{2}{c|}{\textbf{\mymodel}} & \multicolumn{2}{c|}{\textbf{-structural}} & \multicolumn{2}{c}{\textbf{-semantic}} \\
        & \textbf{Macro-F1} & \textbf{Micro-F1} & \textbf{Macro-F1} & \textbf{Micro-F1} & \textbf{Macro-F1} & \textbf{Micro-F1} \\
        \midrule
        \textbf{IMDB} & 91.35 & 93.67 & 44.82 & 53.17 & 96.75 & 98.31 \\
        \textbf{DBLP} & 70.32 & 97.19 & 58.28 & 64.14 & 72.95 & 98.89 \\
        \textbf{Alibaba} & 83.94 & 79.87 & 40.20 & 63.06 & 82.11 & 79.34 \\
        \textbf{Alibaba-s} & 79.02 & 94.15 & 51.58 & 91.26 & 80.16 & 96.39 \\
        \textbf{Amazon} & 94.89 & 95.64 & 63.78 & 73.91 & 92.45 & 94.96 \\
        \bottomrule
        \end{tabular}
    }
    \label{tab::expr:tc}
    \vspace{-1em}
\end{table}

\begin{figure*}[tbp!]
    \scriptsize
    \centering
    \setlength{\tabcolsep}{2pt}
    \resizebox{\linewidth}{!}{
        \begin{tabular}{ccccc}
        & \multicolumn{4}{c}{\quad \includegraphics[width=0.5\linewidth, align=c]{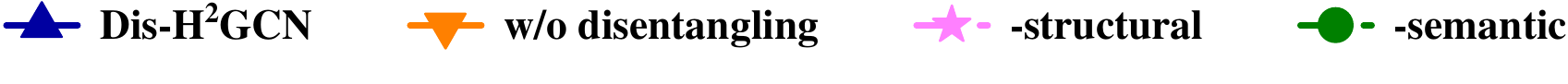}} \\
        & \quad~\quad \textbf{Node classification} & \quad~\quad \textbf{Link prediction} & \quad~\quad \textbf{Node classification} & \quad~\quad \textbf{Link prediction} \\
        & \multicolumn{2}{c}{\quad \textbf{Proportion of training set}} & \multicolumn{2}{c}{\quad \textbf{Number of downstream model layers}} \\
        \rotatebox{90}{\textbf{DBLP}} & \includegraphics[width=0.22\linewidth, align=c]{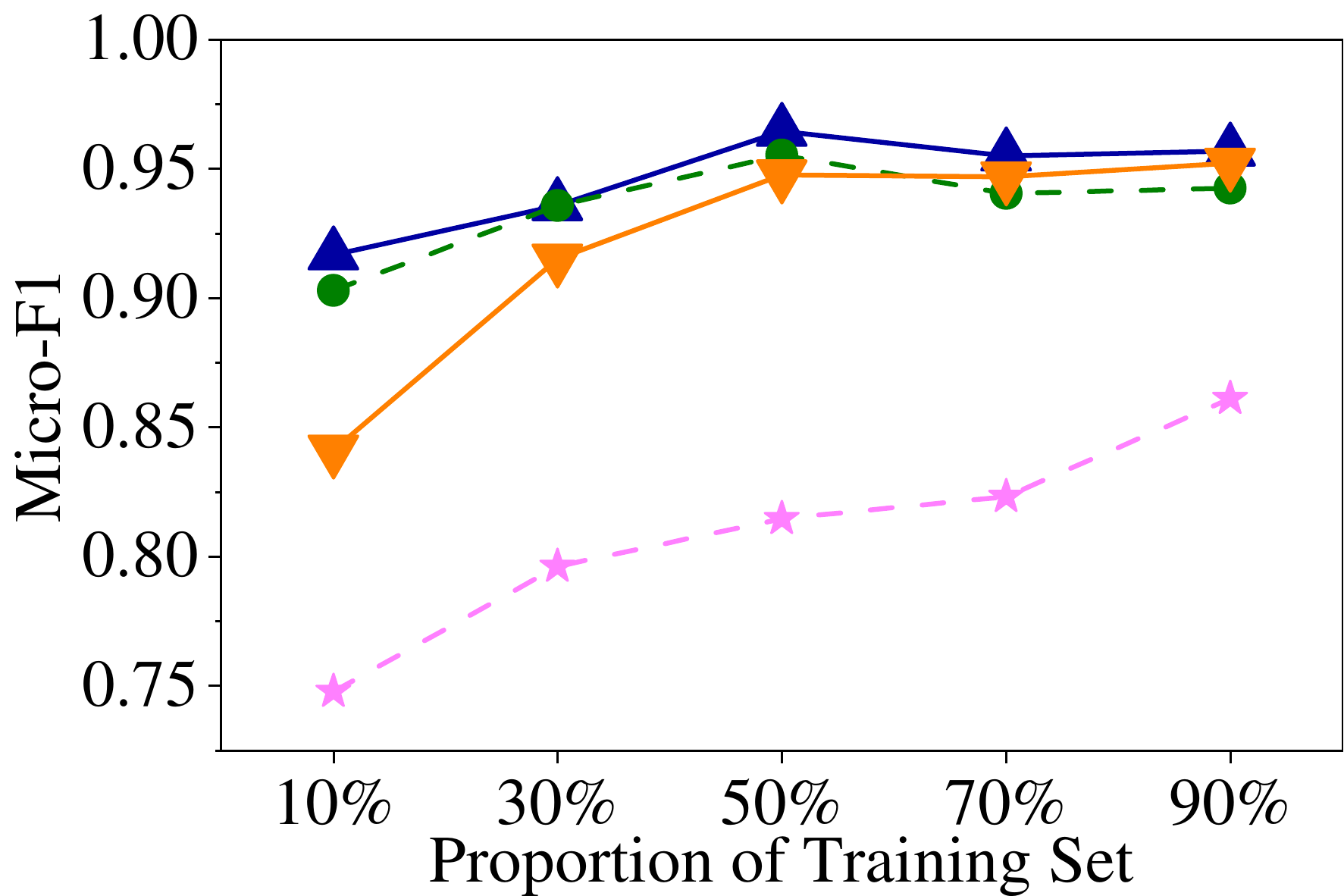} & \includegraphics[width=0.22\linewidth, align=c]{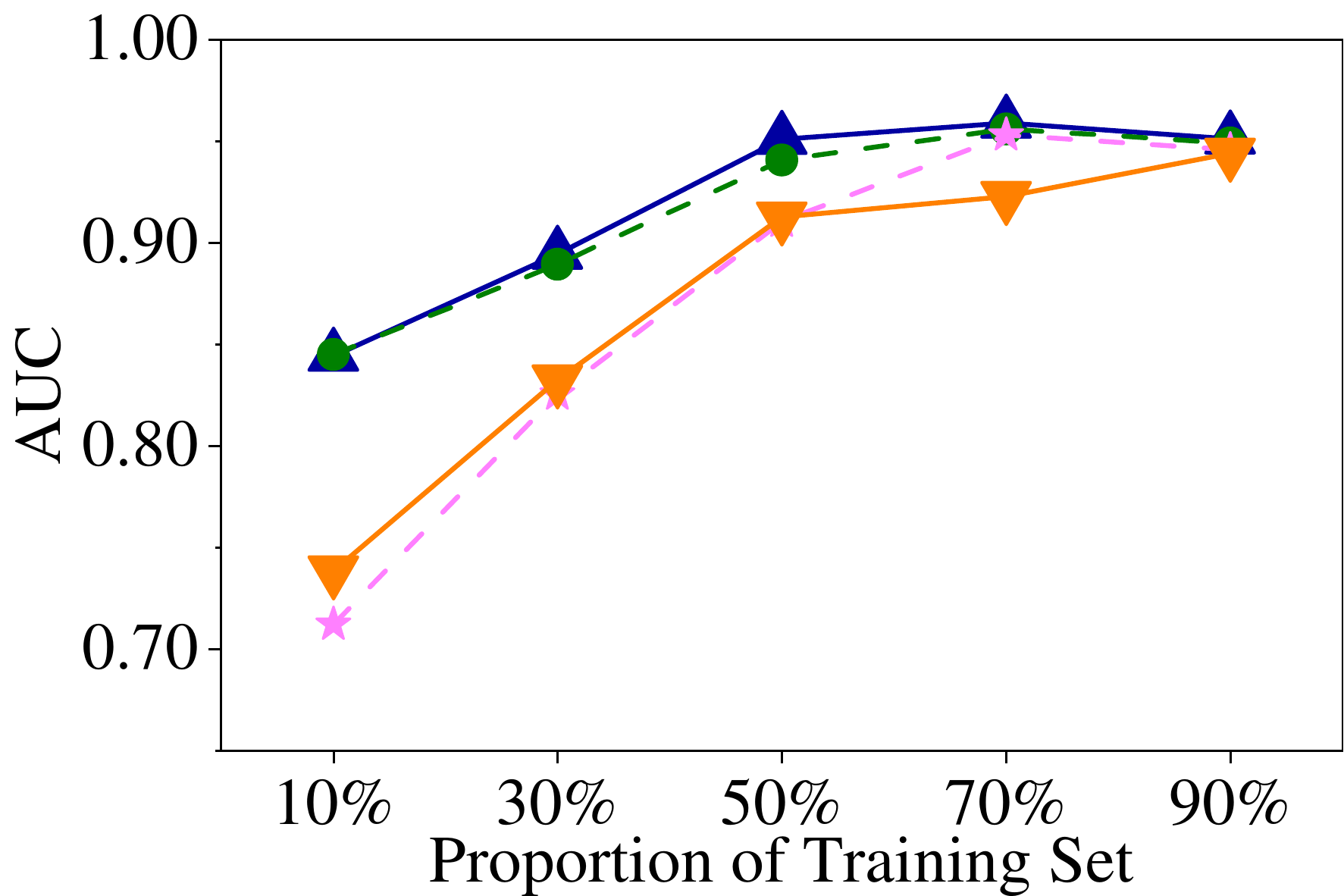} & \includegraphics[width=0.22\linewidth, align=c]{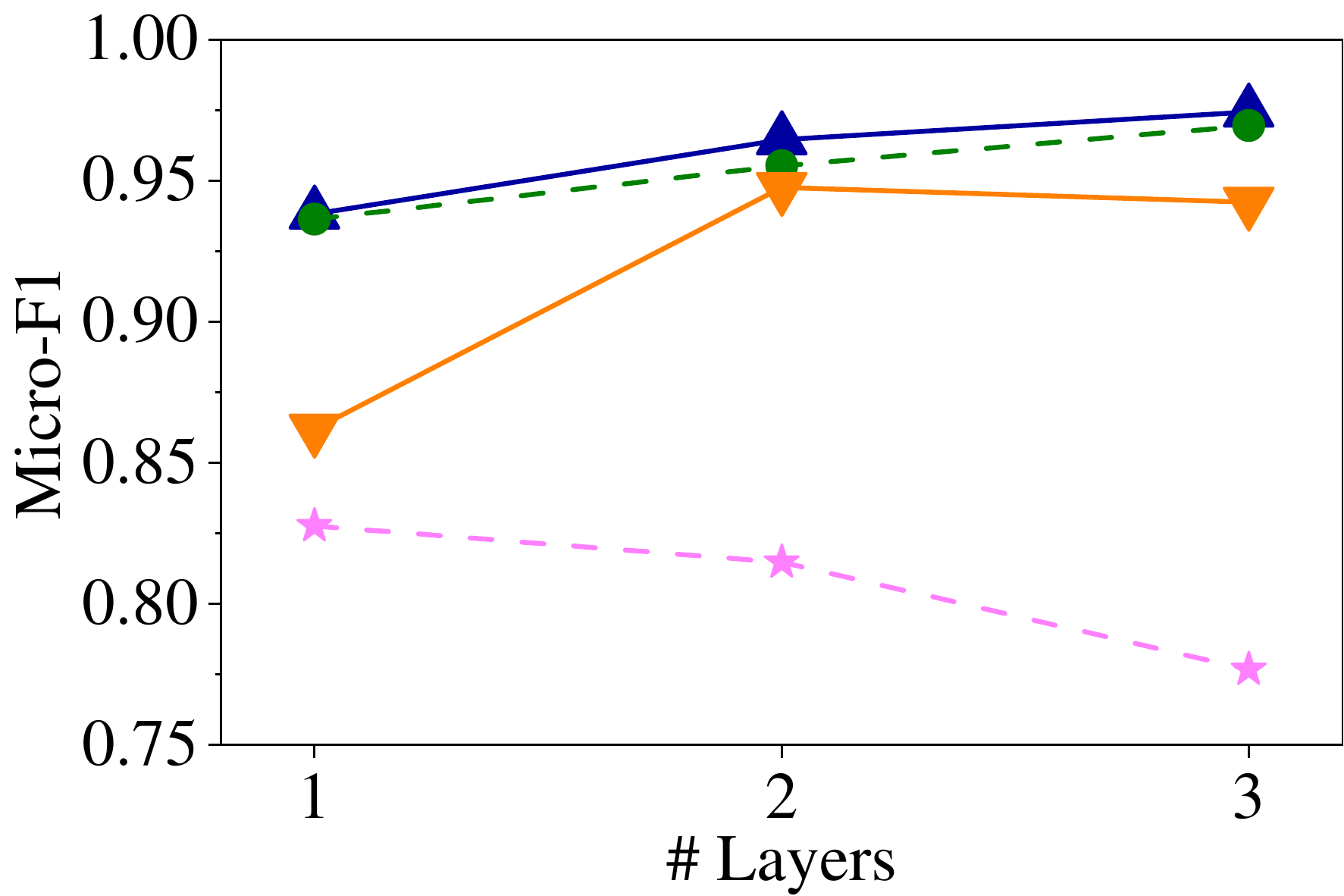} & \includegraphics[width=0.22\linewidth, align=c]{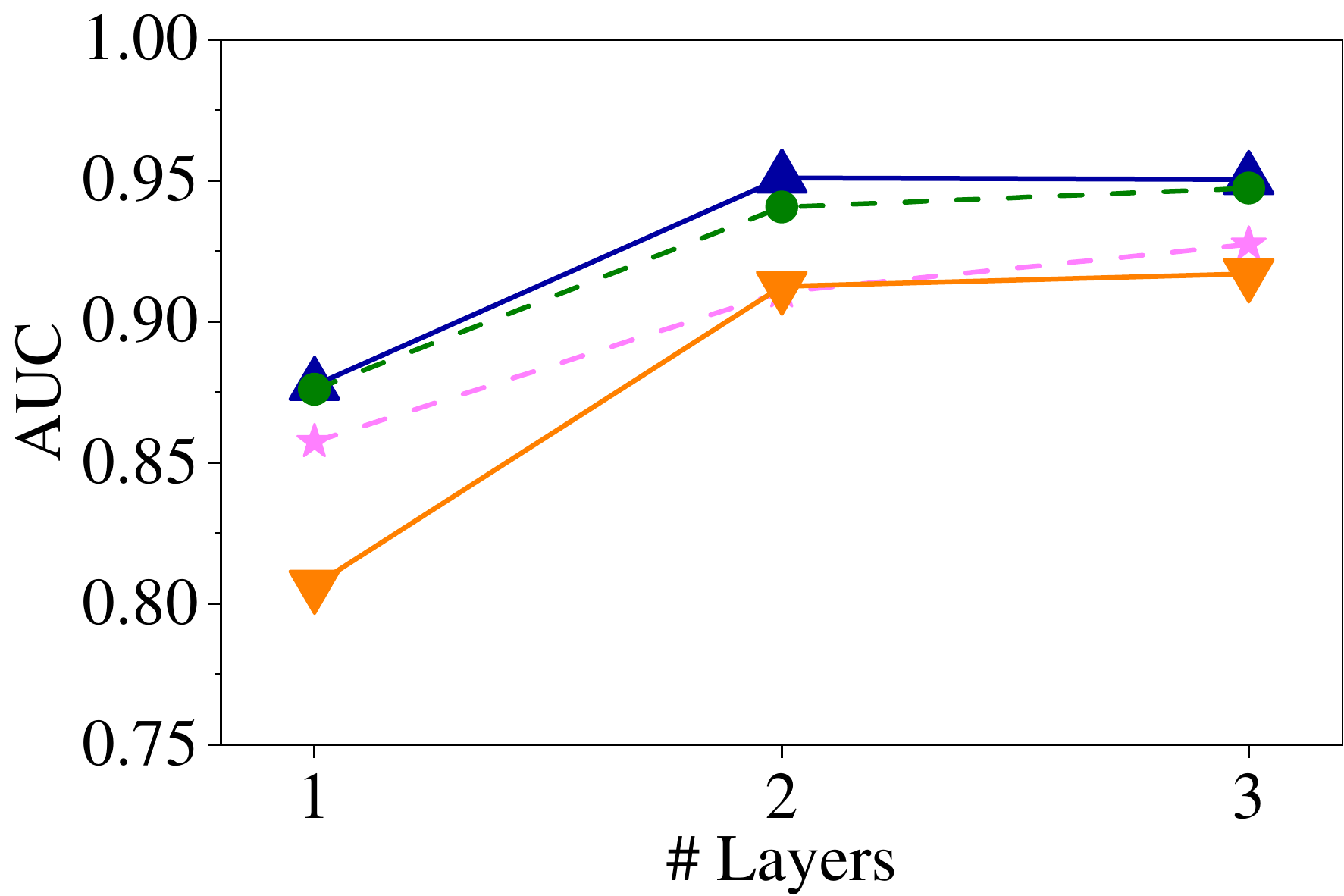} \\
        \rotatebox{90}{\textbf{Alibaba}} & \includegraphics[width=0.22\linewidth, align=c]{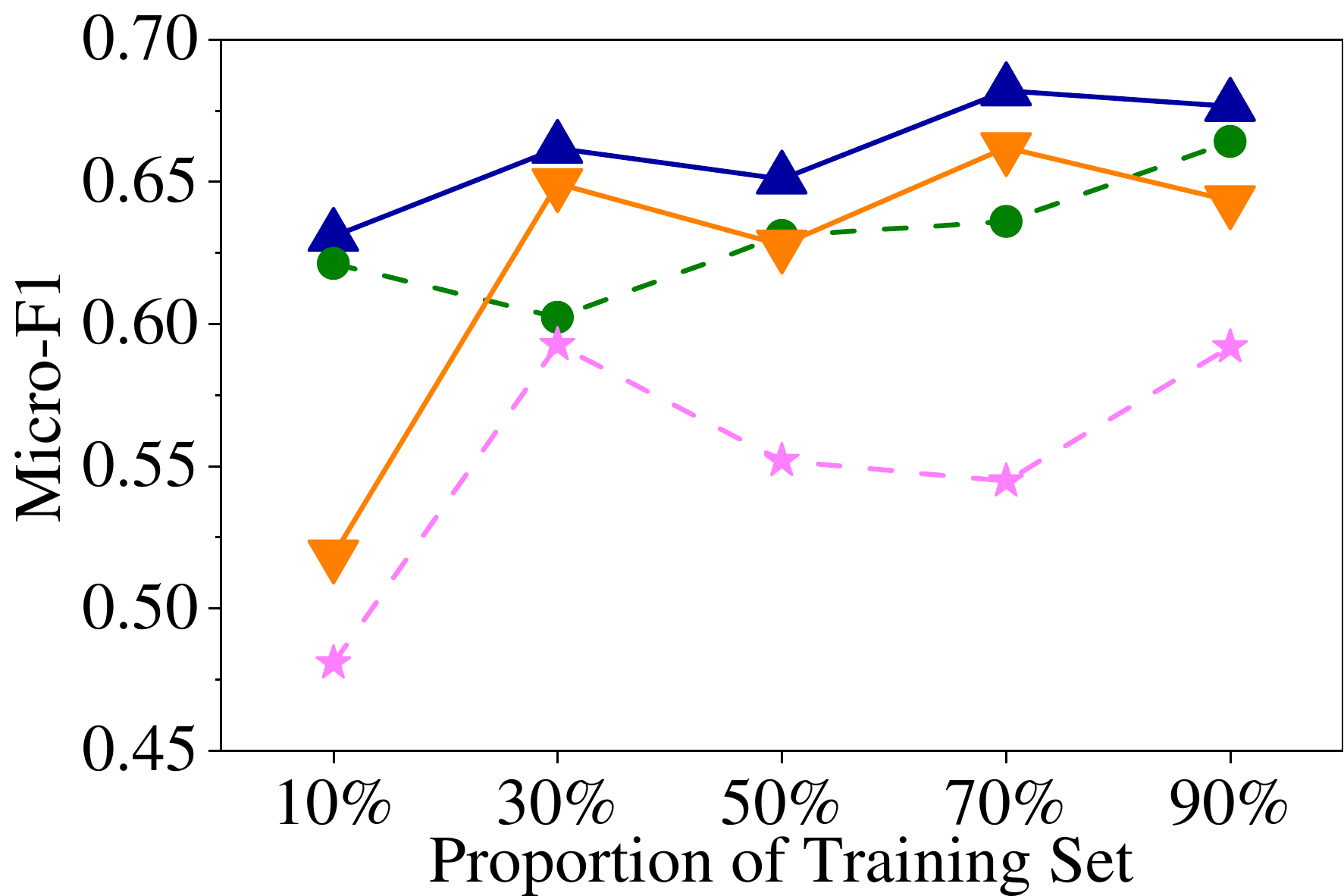} & \includegraphics[width=0.22\linewidth, align=c]{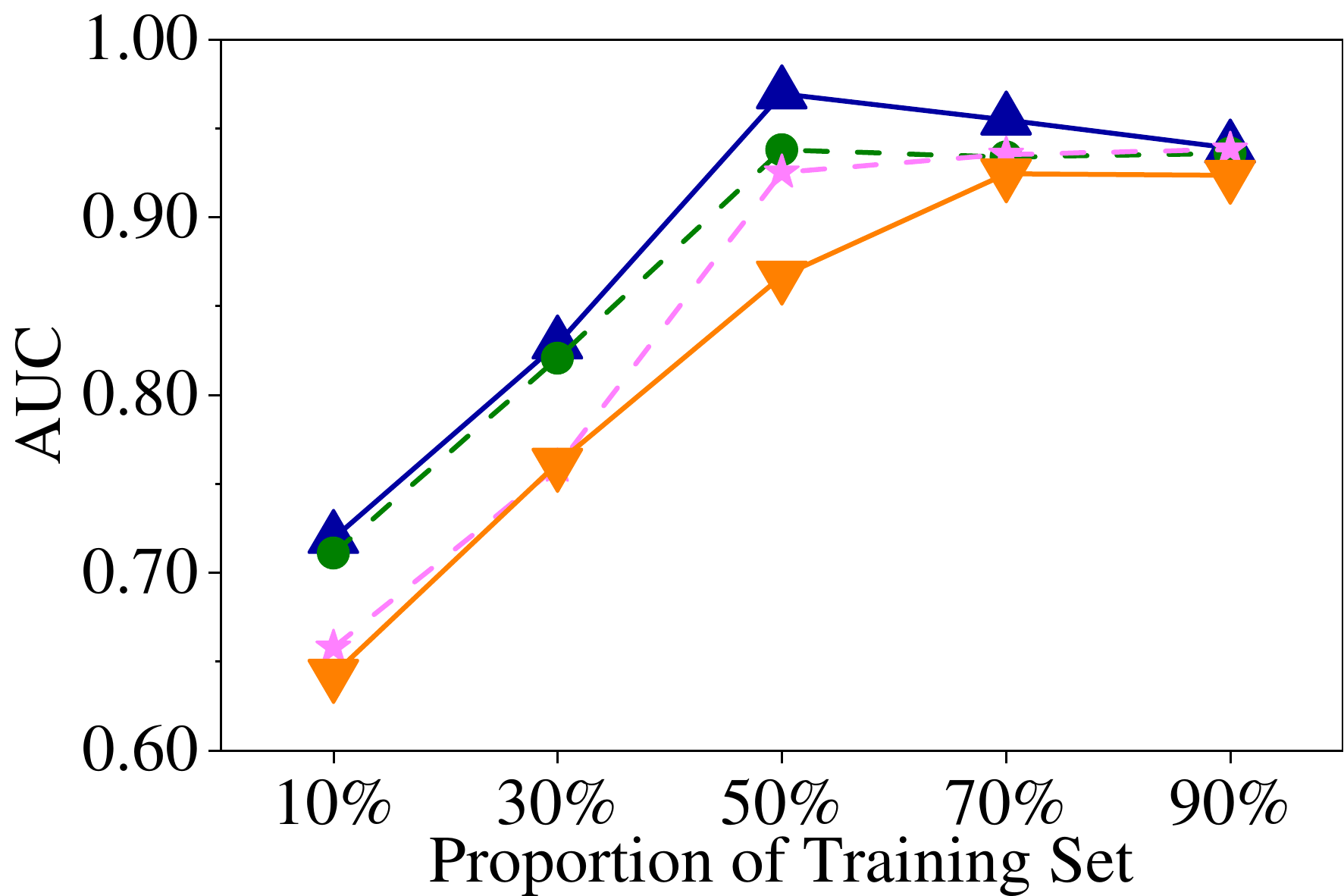} & \includegraphics[width=0.22\linewidth, align=c]{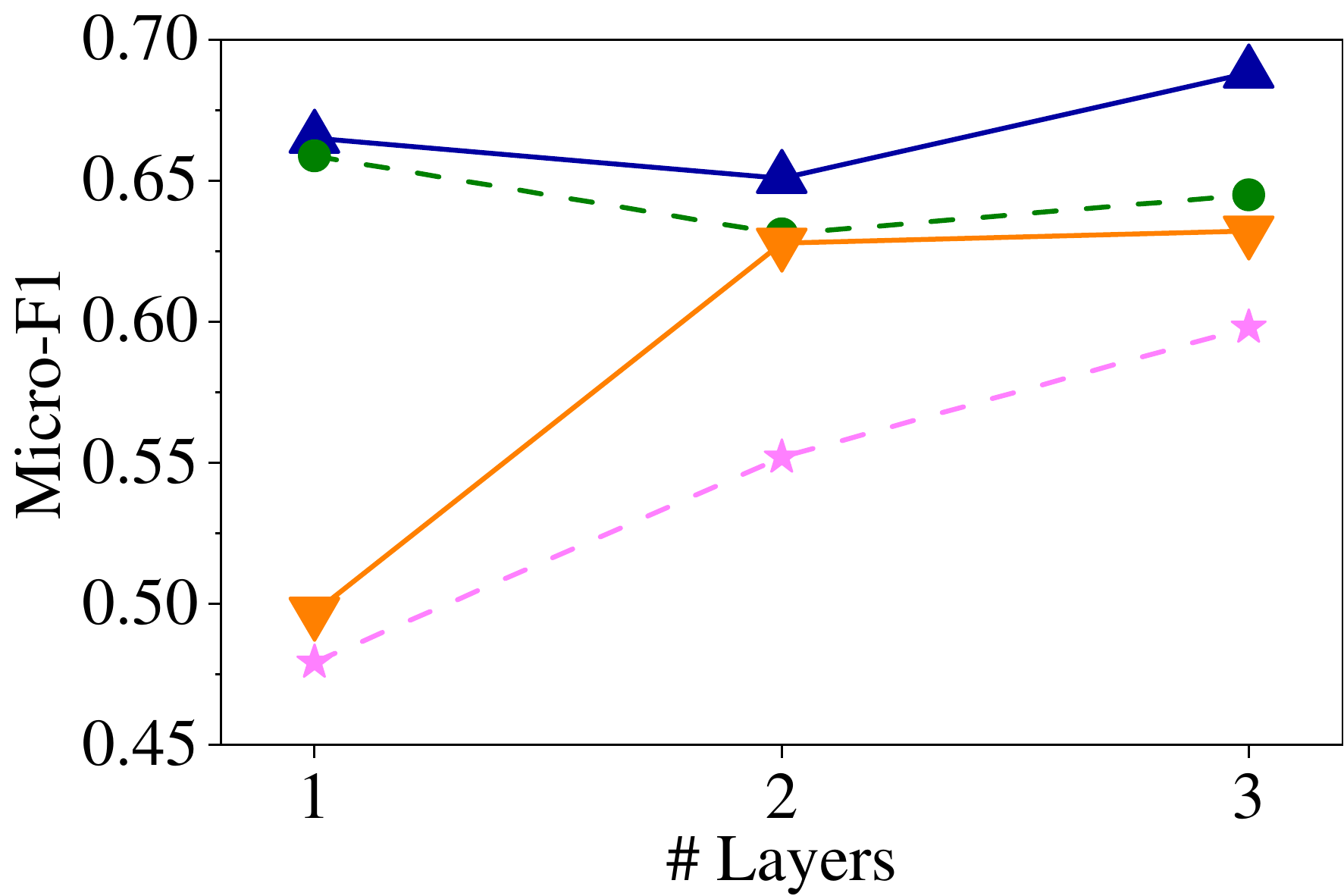} & \includegraphics[width=0.22\linewidth, align=c]{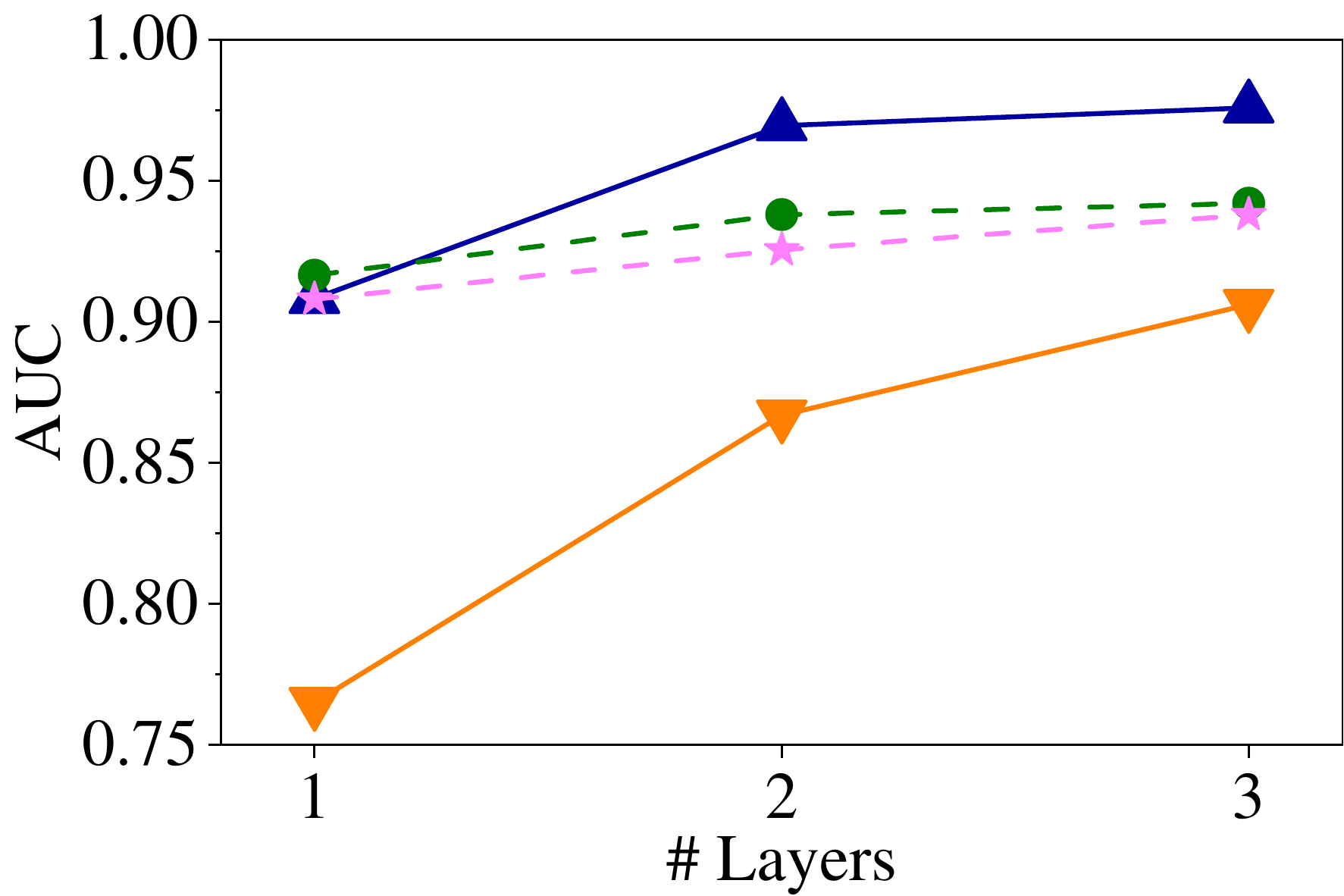} \\
        & & & & \\
        & \multicolumn{2}{c}{\quad \textbf{Proportion of structural perturbation}} & \multicolumn{2}{c}{\quad \textbf{Proportion of semantic perturbation}} \\
        %
        % & \quad~\quad \textbf{Node classification} & \quad~\quad \textbf{Link prediction} & \quad~\quad \textbf{Node classification} & \quad~\quad \textbf{Link prediction} \\
        %
        \rotatebox{90}{\textbf{DBLP}} & \includegraphics[width=0.22\linewidth, align=c]{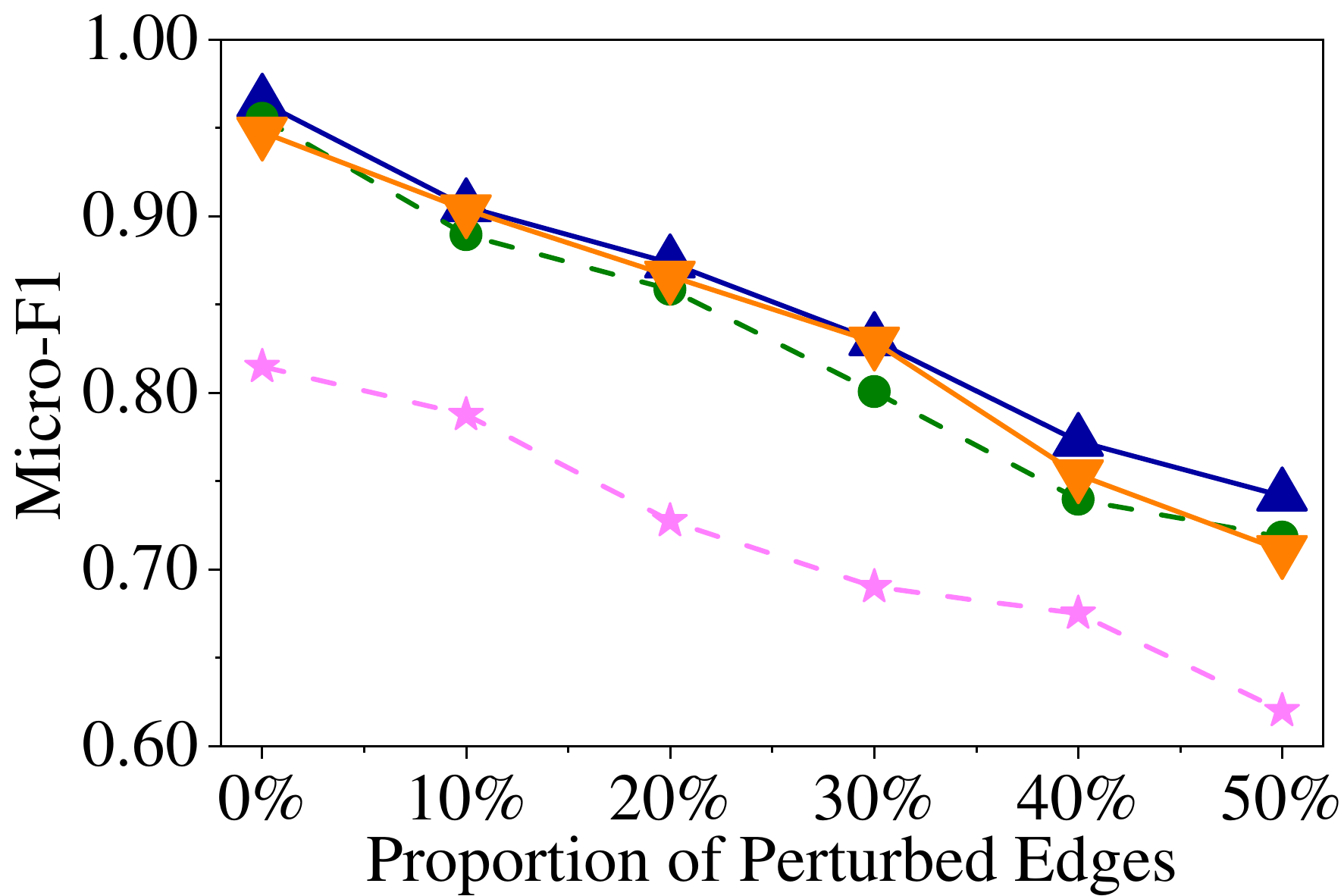} & \includegraphics[width=0.22\linewidth, align=c]{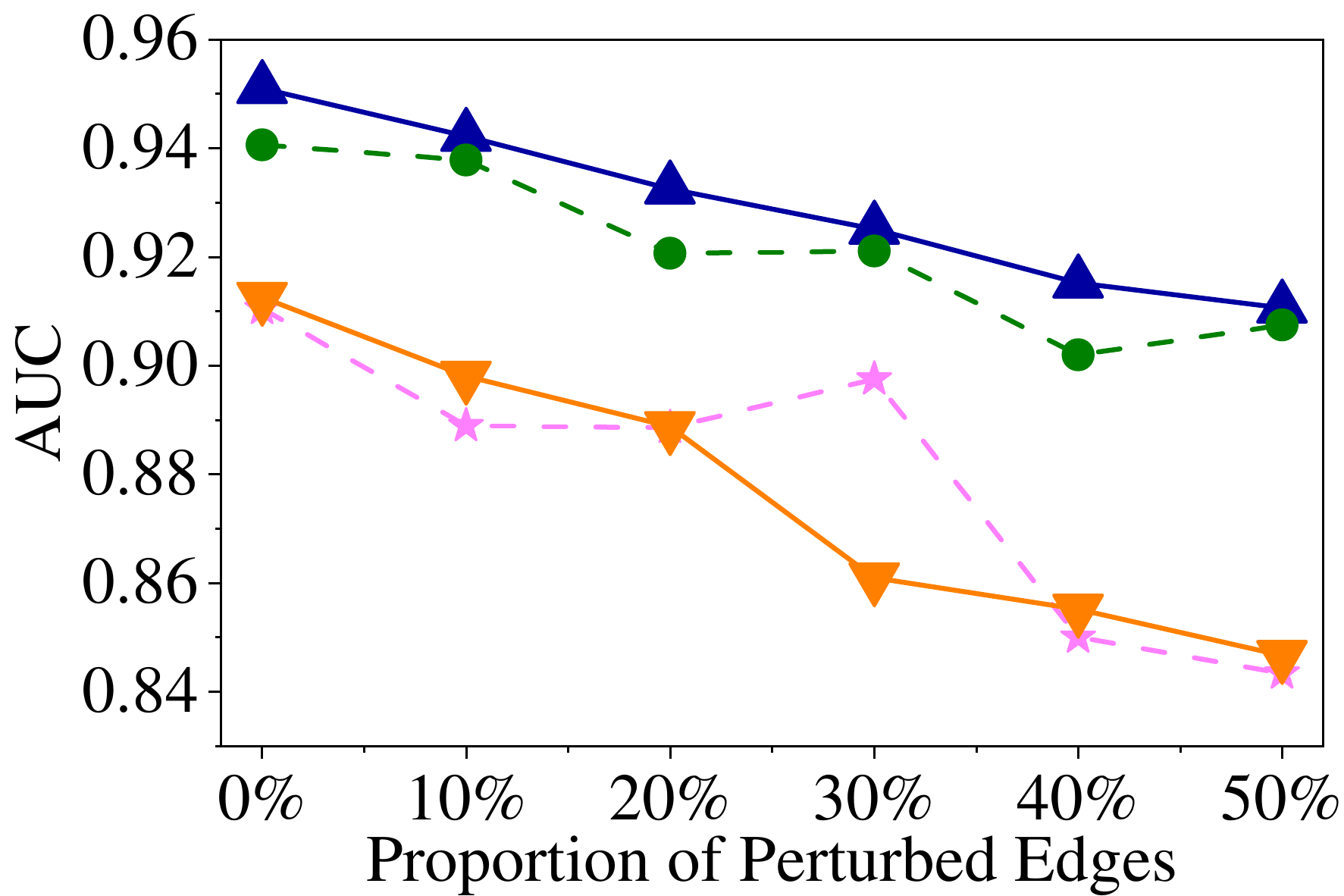} & \includegraphics[width=0.22\linewidth, align=c]{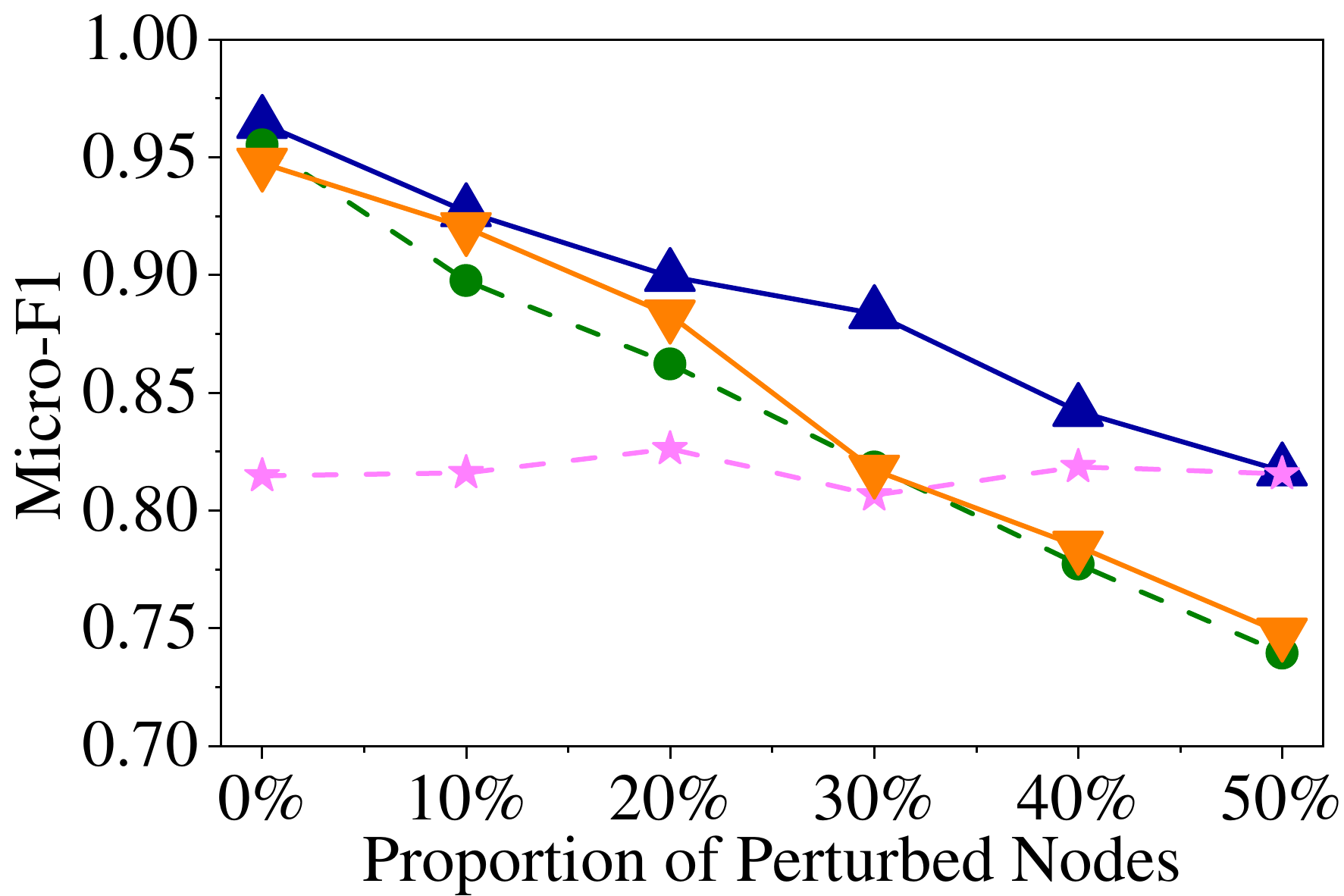} & \includegraphics[width=0.22\linewidth, align=c]{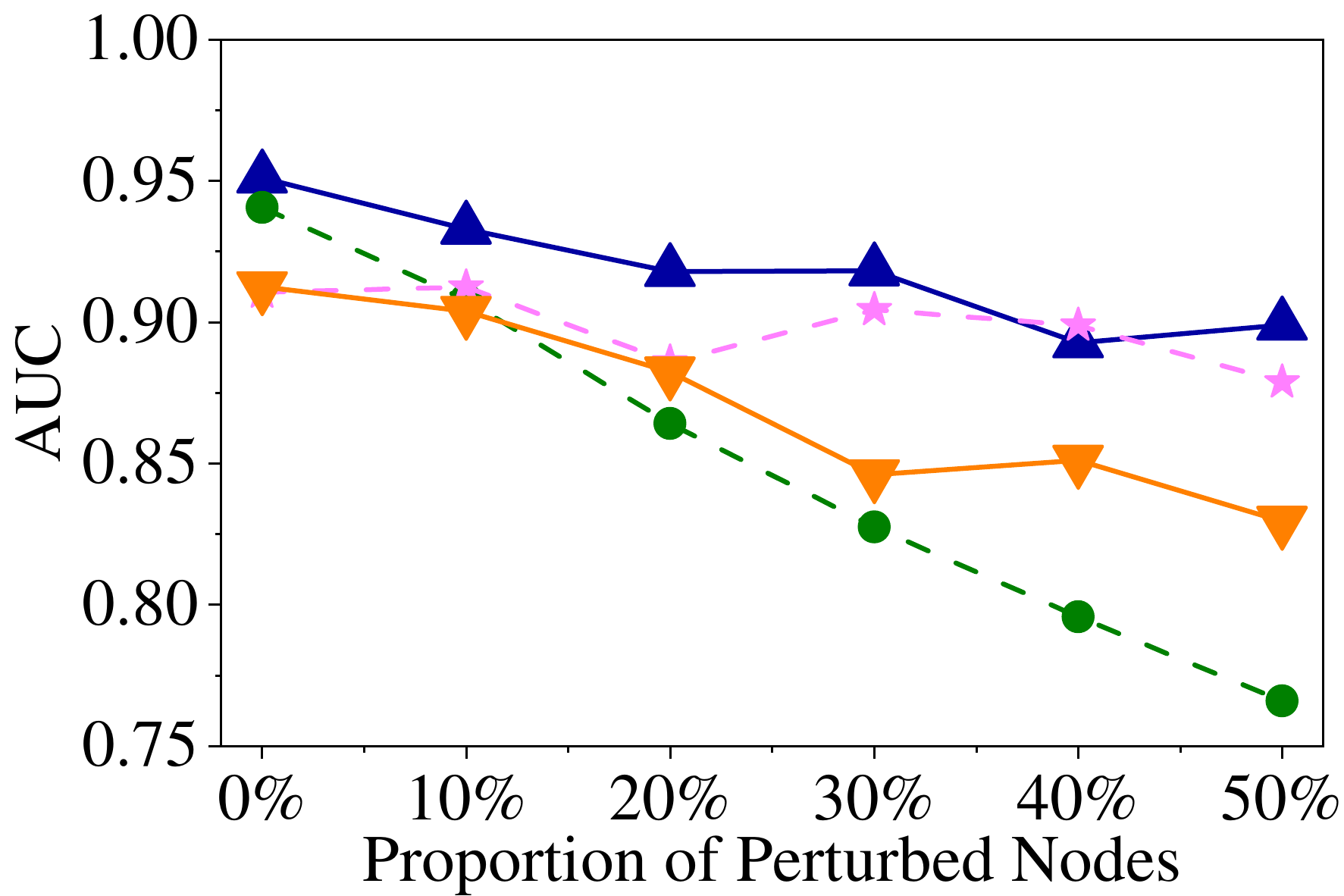} \\
        \rotatebox{90}{\textbf{Alibaba}} & \includegraphics[width=0.22\linewidth, align=c]{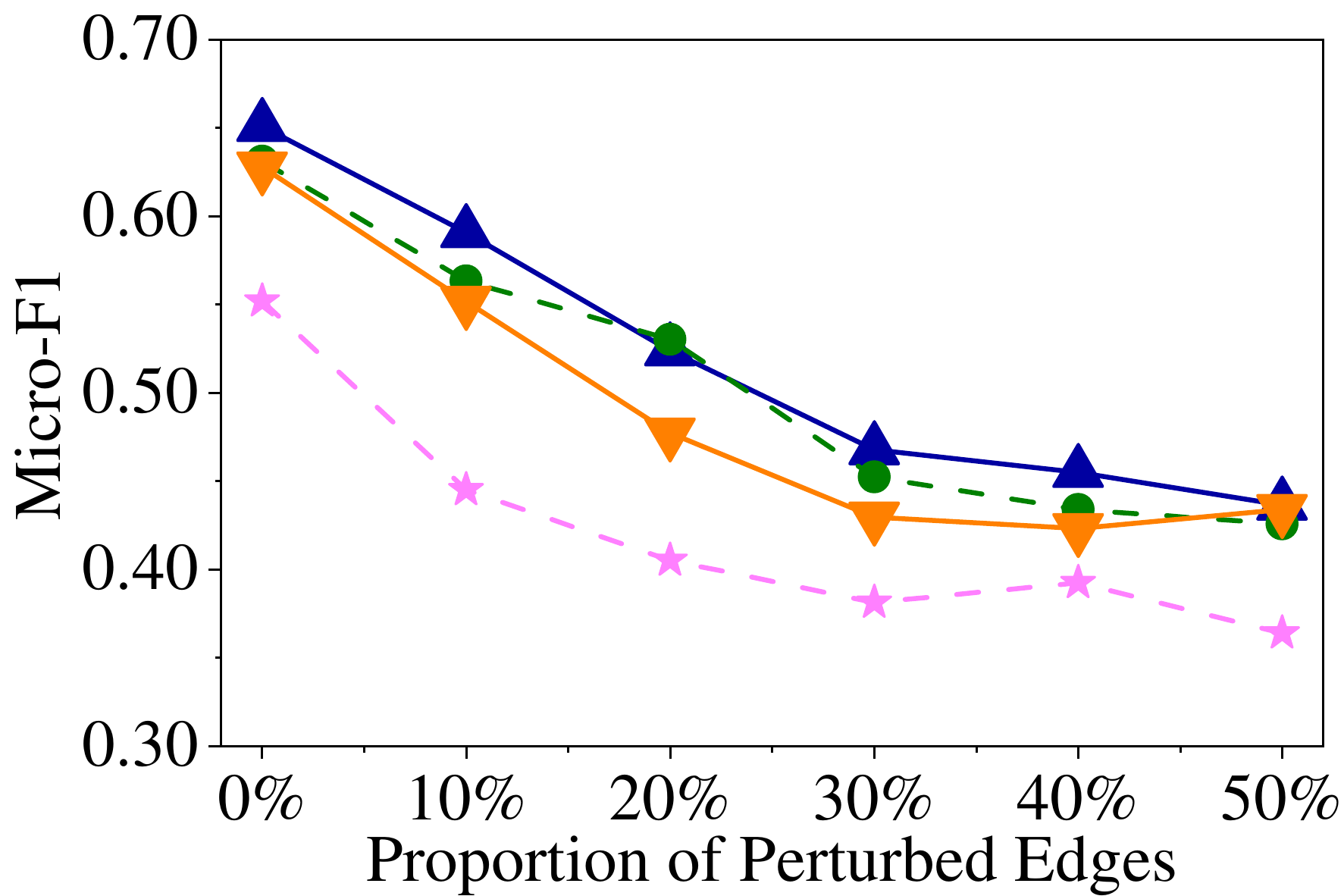} & \includegraphics[width=0.22\linewidth, align=c]{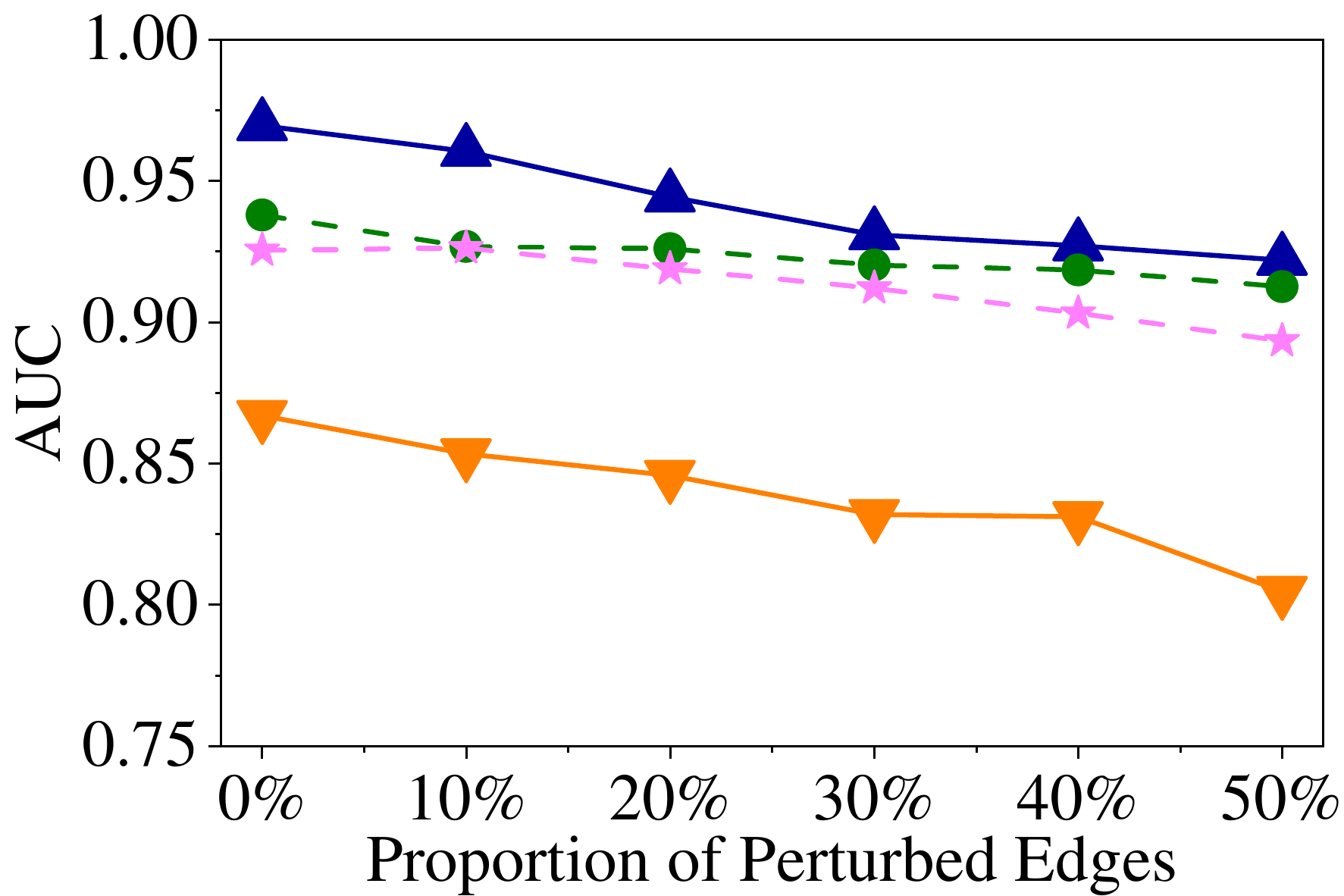} & \includegraphics[width=0.22\linewidth, align=c]{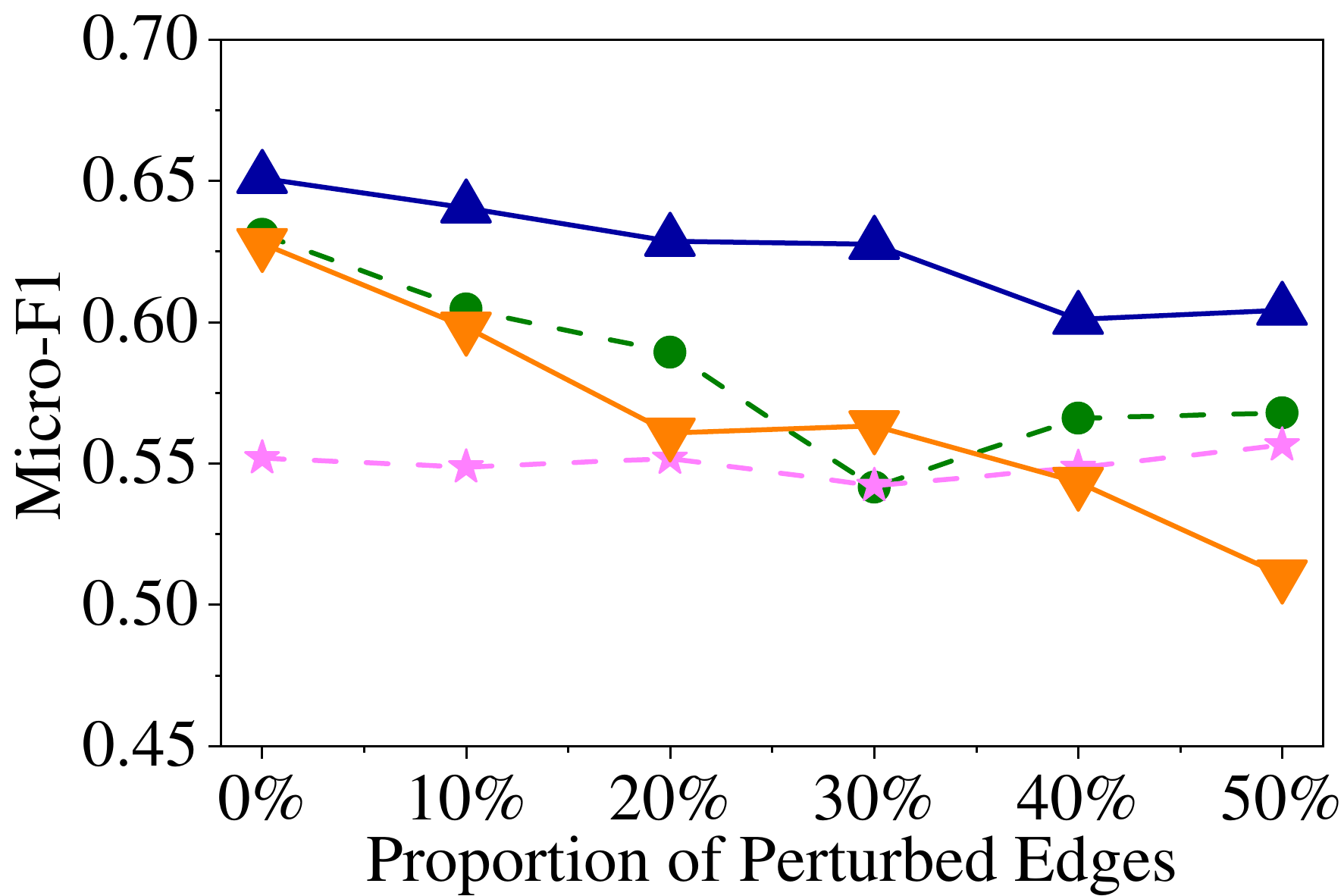} & \includegraphics[width=0.22\linewidth, align=c]{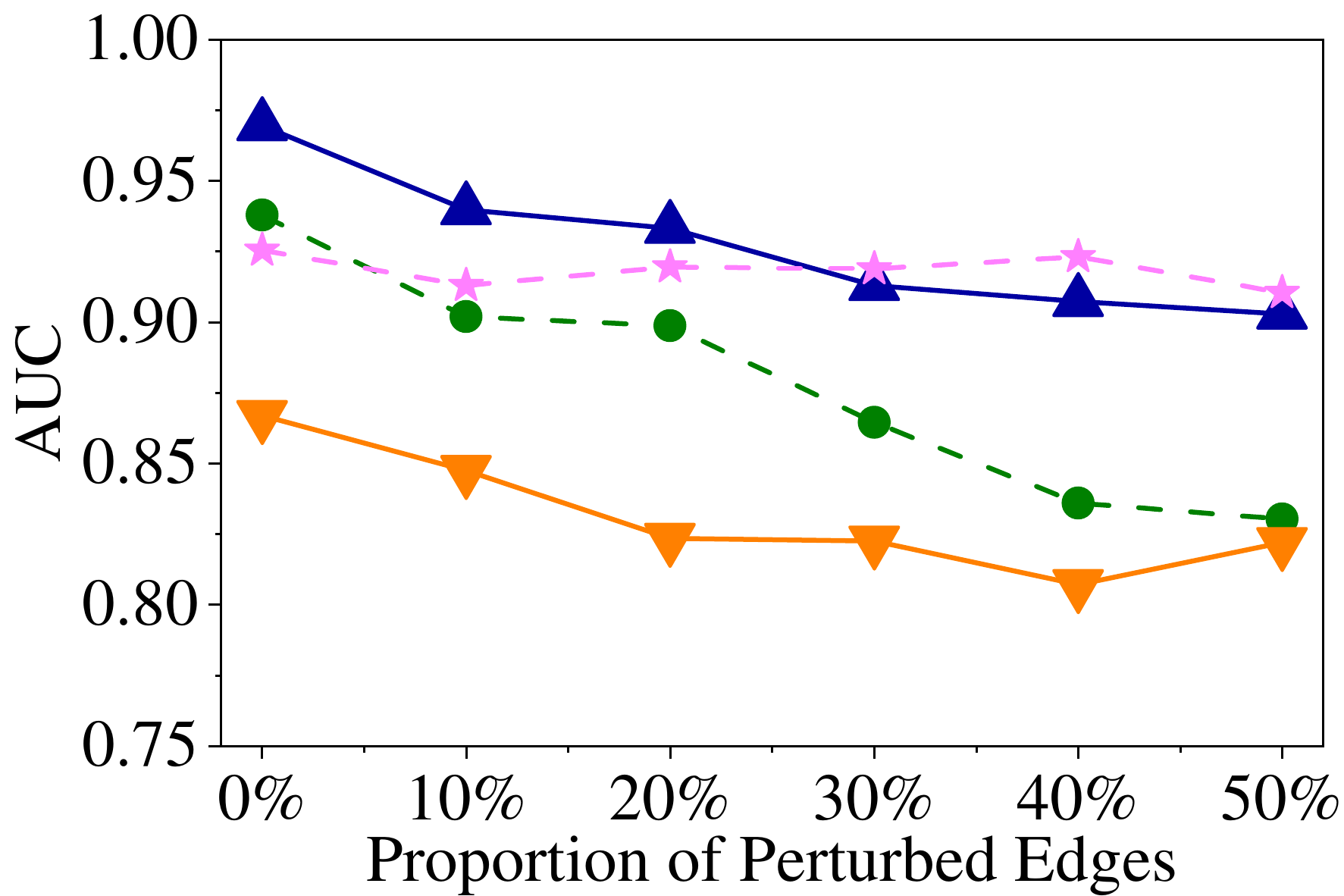} \\
        \end{tabular}
    }
    \caption{The performance with different training resources (proportions of training set and numbers of downstream model layers, the upper two rows) and against different perturbations (proportions of structural and semantic perturbations, the lower two rows) on DBLP and Alibaba datasets.}
    \label{fig::expr:benefit}
    \vspace{-1em}
\end{figure*}

In order to answer RQ4, we study the benefits of disentanglement on DBLP and Alibaba in the following two aspects.

\subsubsection{Training Resources}  \label{sec::exper:benefit_disentangle:resource}

We evaluate the performance variation of \mymodel and w/o disentangling by adjusting two training resources: proportion of training set (from 10\% to 90\% in steps of 20\%) and layer number of downstream models (from 1 to 3).
As the upper two rows of Fig.~\ref{fig::expr:benefit} demonstrate, the model with disentangling module keeps a larger performance lead on a smaller training set or a simpler downstream model than on more training resources in both node classification and link prediction.
The results prove that the disentanglement helps in suffering from the fewer training resources.

\begin{figure}
    \centering
    \setlength{\tabcolsep}{3pt}
    \resizebox{\linewidth}{!}{
        \begin{tabular}{cc}
        \quad~\quad \textbf{Node classification} & \quad~\quad \textbf{Link prediction} \\
        \includegraphics[width=0.66\linewidth]{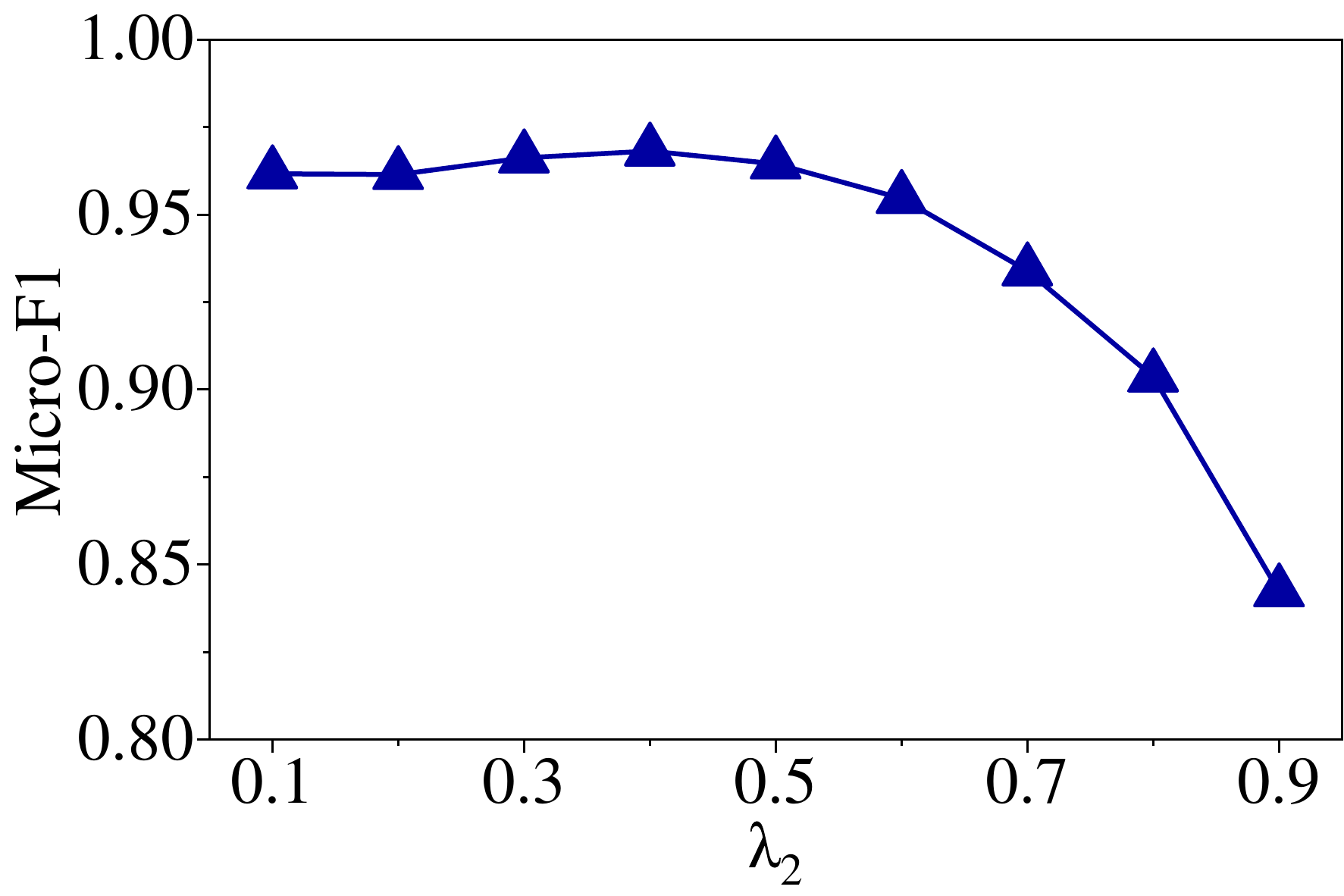} & \includegraphics[width=0.66\linewidth]{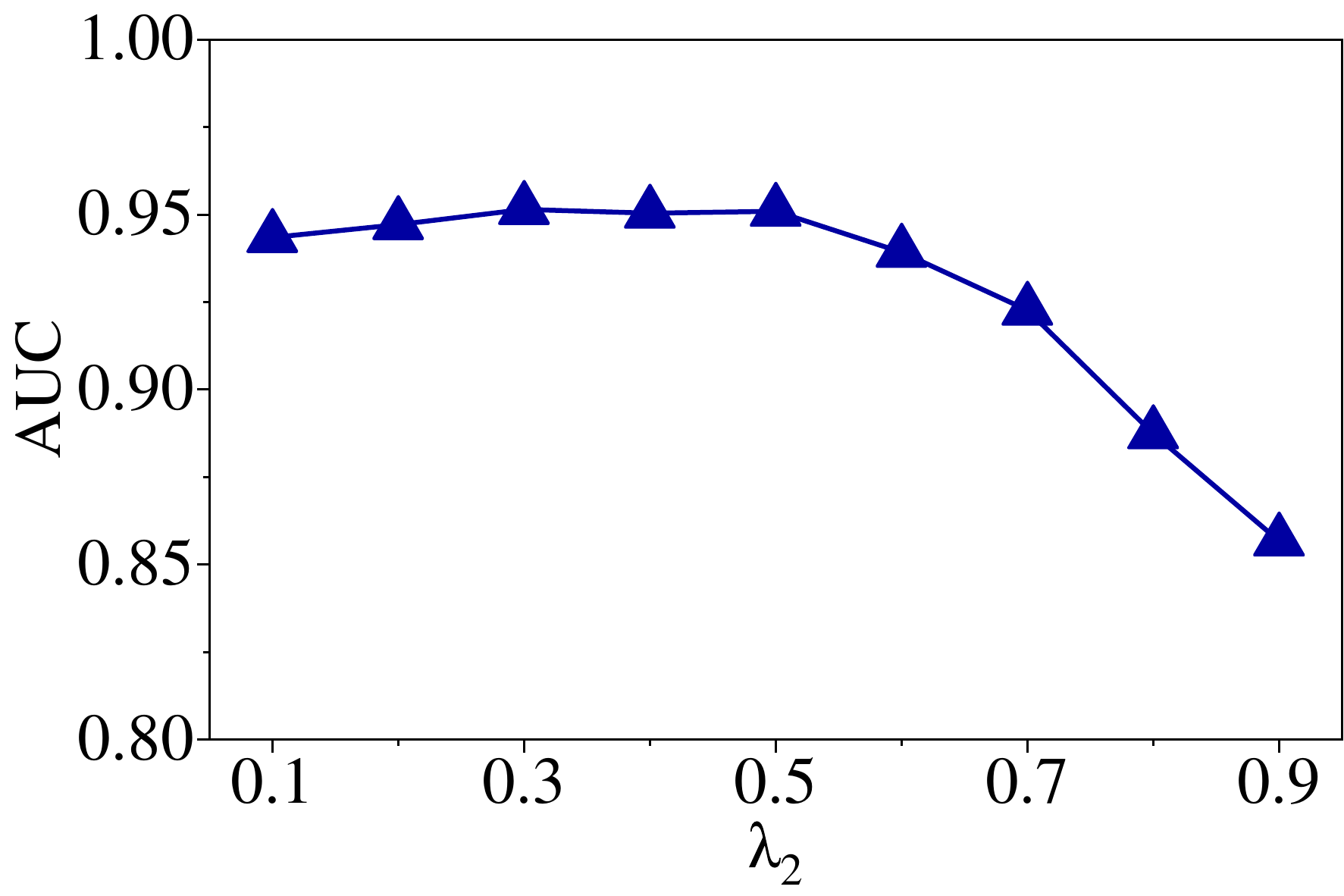} \\ 
        \includegraphics[width=0.66\linewidth]{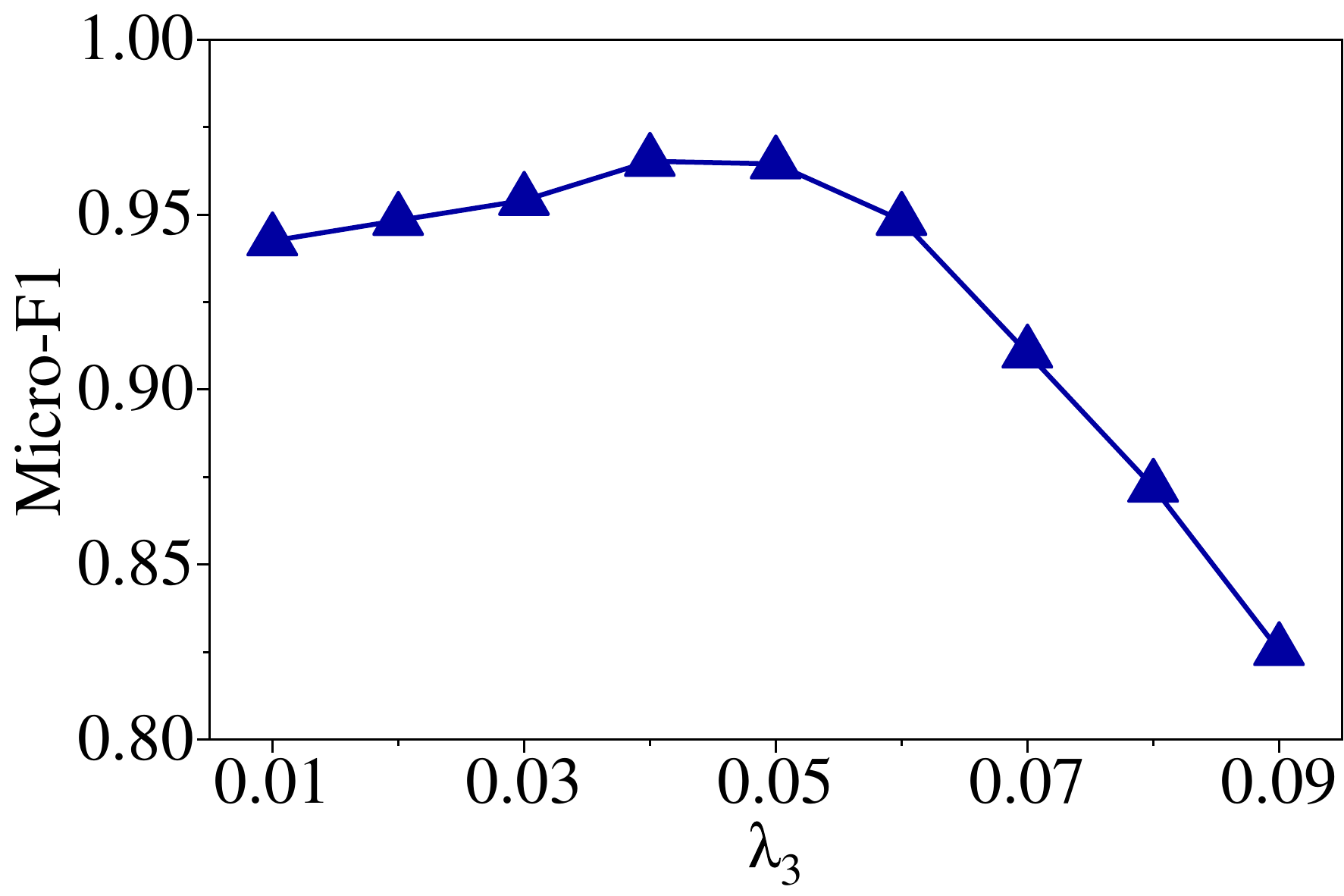} & \includegraphics[width=0.66\linewidth]{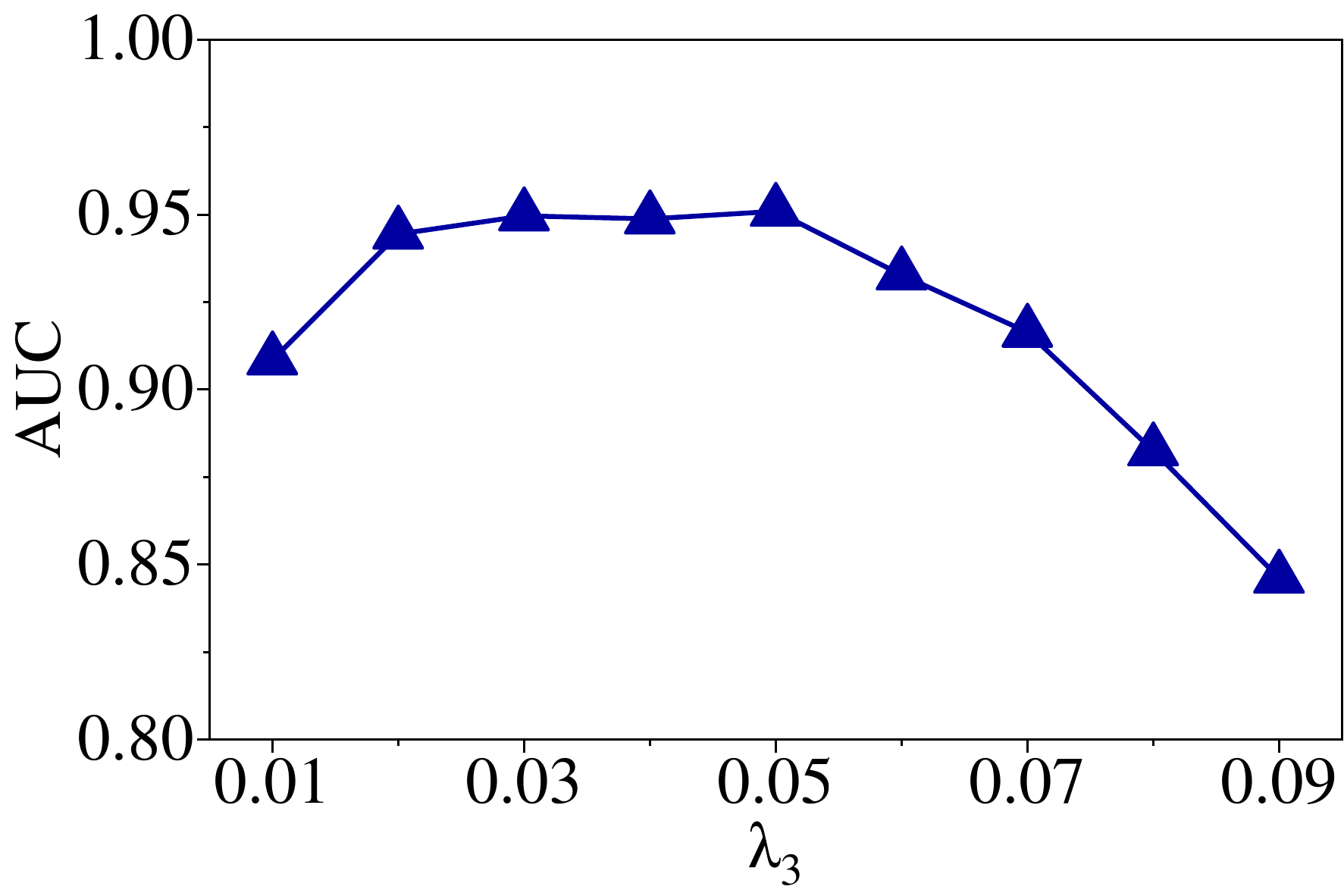} \\
        \end{tabular}
    }
    \caption{The influences of $\lambda_2$ (weight of disentangling module) and $\lambda_3$ (weight of contrastive learning) on DBLP.}
    \label{fig::expr:parameter}
    \vspace{-1em}
\end{figure}

% \begin{figure}
%     \centering
%     \subfloat[Node classification.]{
%         \includegraphics[width=0.35\linewidth]{figs/experiments/parameters/lambda_2_nc.pdf}
%         \label{fig::expr:parameter_lambda_2:nc}
%     } 
%     \subfloat[Link prediction.]{
%         \includegraphics[width=0.35\linewidth]{figs/experiments/parameters/lambda_2_lp.pdf}
%         \label{fig::expr:parameter_lambda_2:lp}
%     }
%     \caption{The influences of $\lambda_2$ (weight of disentangling module) on DBLP.}
%     \label{fig::expr:parameter_lambda_2}
% \end{figure}

% \begin{figure}
%     \centering
%     \subfloat[Node classification.]{
%         \includegraphics[width=0.35\linewidth]{figs/experiments/parameters/lambda_3_nc.pdf}
%         \label{fig::expr:parameter_lambda_3:nc}
%     } 
%     \subfloat[Link prediction.]{
%         \includegraphics[width=0.35\linewidth]{figs/experiments/parameters/lambda_3_lp.pdf}
%         \label{fig::expr:parameter_lambda_3:lp}
%     }
%     \caption{The influences of $\lambda_3$ (weight of contrastive learning) on DBLP.}
%     \label{fig::expr:parameter_lambda_3}
% \end{figure}

\subsubsection{Robustness against Perturbations}  \label{sec::exper:benefit_disentangle:perturbation}

To verify the effect of disentanglement in robustness, we make two kinds of perturbations on training data.
Structural perturbations refer to adding or deleting edges in the input graphs, and semantic perturbations disturb the node types in DBLP and edge types in Alibaba with certain ratios.
The results are shown in the lower two rows of Fig.~\ref{fig::expr:benefit}, revealing that the performance of w/o disentangling drops more rapidly than that of \mymodel with the increase of perturbation proportions.
It can be concluded that disentanglement plays a positive role in robustness against perturbations.
It is also worth noting the stable performance of structural embeddings under semantic perturbations, which illustrates that disentangled structural embeddings are little influenced by the semantics in graphs.

\subsection{Parameters Analysis \textup{(\textbf{RQ5})}}   \label{sec::exper:parameter}

We further analyse the influences of hyper-parameters $\lambda_2$ and $\lambda_3$ for RQ5.
The two hyper-parameters are used to trade-off the weights of disentangling module and contrastive learning in the overall loss function.
The value of $\lambda_2$ is set from 0.1 to 0.9 in steps of 0.1, and $\lambda_3$ ranges from 0.01 to 0.09 in steps of 0.01.
The results of both evaluation tasks with various values of $\lambda_2$ and $\lambda_3$ on DBLP are shown in Fig.~\ref{fig::expr:parameter}.

It can be observed that the results in all cases exhibit a tendency to first rise and then drop.
This is because the effects of both disentangling module and contrastive learning are more likely a regularization in the loss function to constrain the training process bearing on their respective optimization objectives.
A tiny value is not able to impose sufficient constraints, while a too large value may directly affect the optimization for main tasks.
What is worthy noting is that the performance deteriorates rapidly with the continuous increase of $\lambda_2$, and the deterioration is even severer for $\lambda_3$ although its value is smaller.
It suggests that contrastive learning is valid in supervised tasks but its weight need to be set carefully to achieve the best performance.

\section{Conclusion}  \label{sec::conlu}

In this paper, we propose a disentangled hyperbolic representation learning method \mymodel for heterogeneous graphs, addressing the problem of mixed representations and Euclidean spaces not conducing to modeling the rich information therein.
Specifically, we first collapse the heterogeneous graphs into corresponding homogeneous ones for the learning of pure structural information.
Simultaneously the contrastive learning upon a heterogeneous graph convolutional network with the type-based message propagation mechanism is applied to fully mine the heterogeneous information in the graphs.
In the disentangling module, we use the mutual information estimation and a trainable discriminator to constraint the model training.
The entire model is built on the hyperbolic spaces to meet the data distributions.
Extensive experiments on real-world datasets show the superiority of \mymodel and the further analysis proves the effectiveness of disentanglement and hyperbolic geometry.

\bibliographystyle{IEEEtran}
\bibliography{biblist}

% \newpage

 

\begin{IEEEbiography}[{\includegraphics[width=1in,height=1.25in,clip,keepaspectratio]{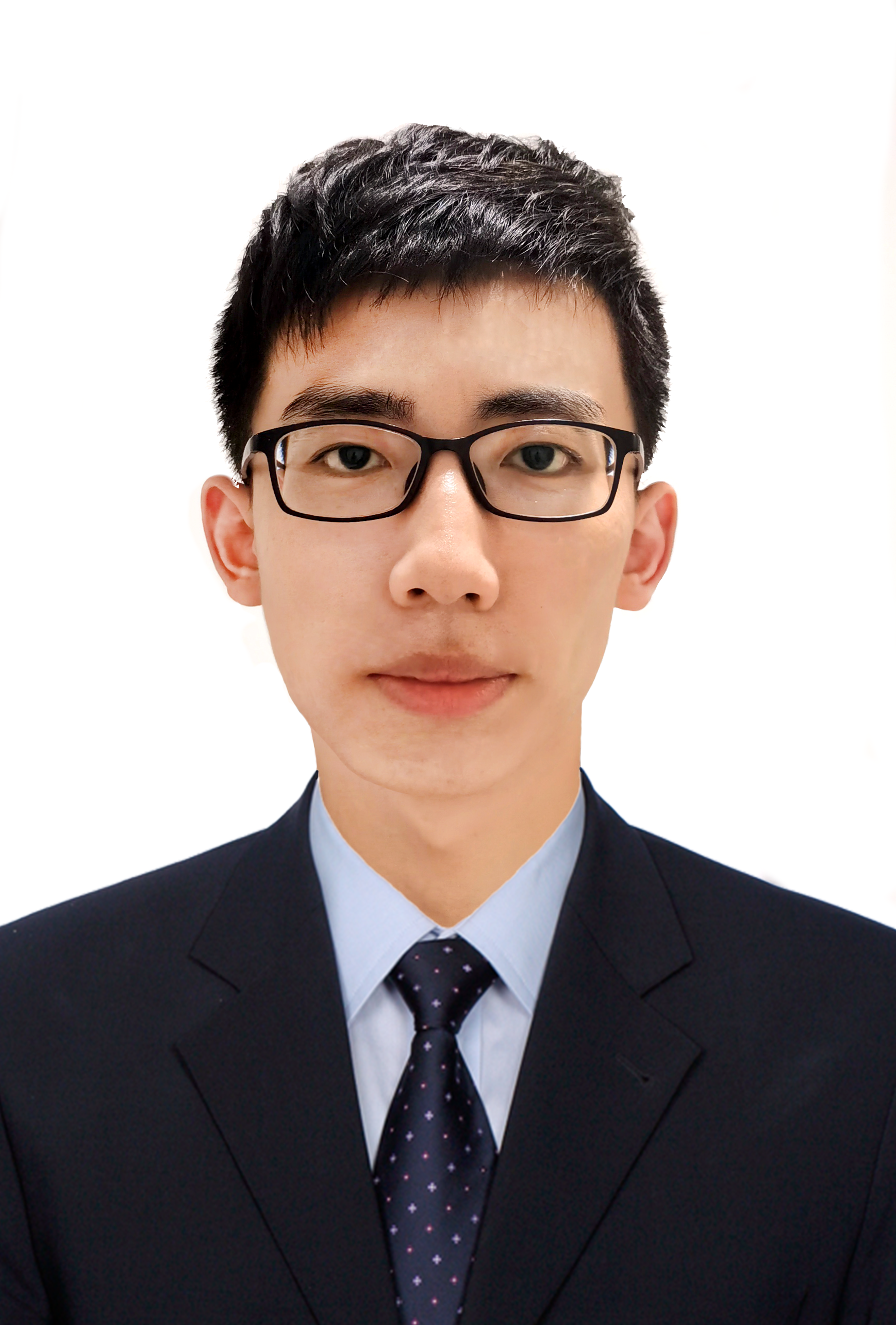}}]{Qijie Bai}
received the B.S. degree from Nankai University, Tianjin, China, in 2020. He is currently a Ph.D. student in Nankai University. His main research interests include graph data, data mining and machine learning.
\end{IEEEbiography}

\begin{IEEEbiography}[{\includegraphics[width=1in,height=1.25in,clip,keepaspectratio]{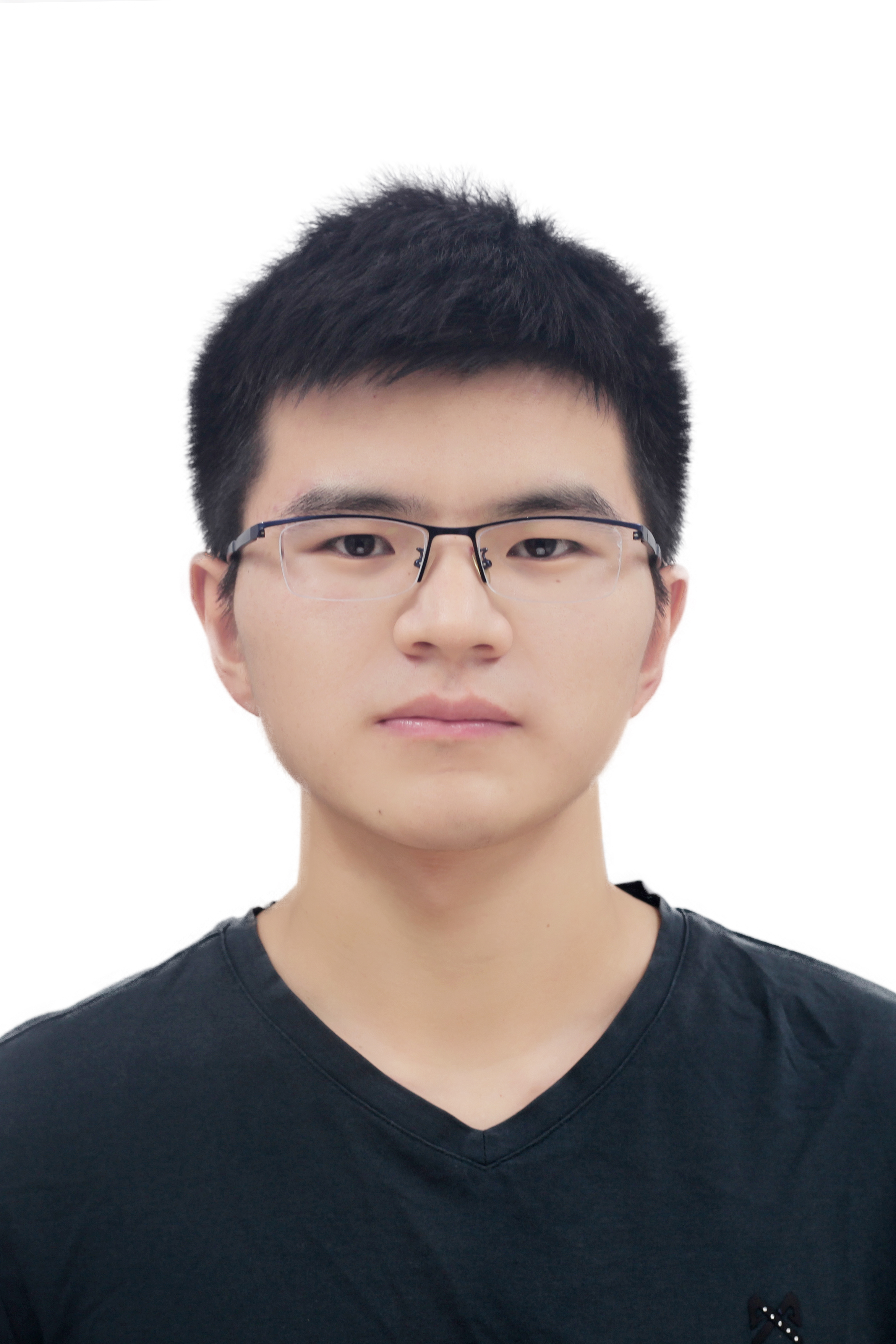}}]{Changli Nie}
received the B.S. degree from Nankai University, Tianjin, China, in 2022. He is currently a master student in Nankai University. His main research interests include graph domain adaption and recommendation system.
\end{IEEEbiography}

\begin{IEEEbiography}[{\includegraphics[width=1in,height=1.25in,clip,keepaspectratio]{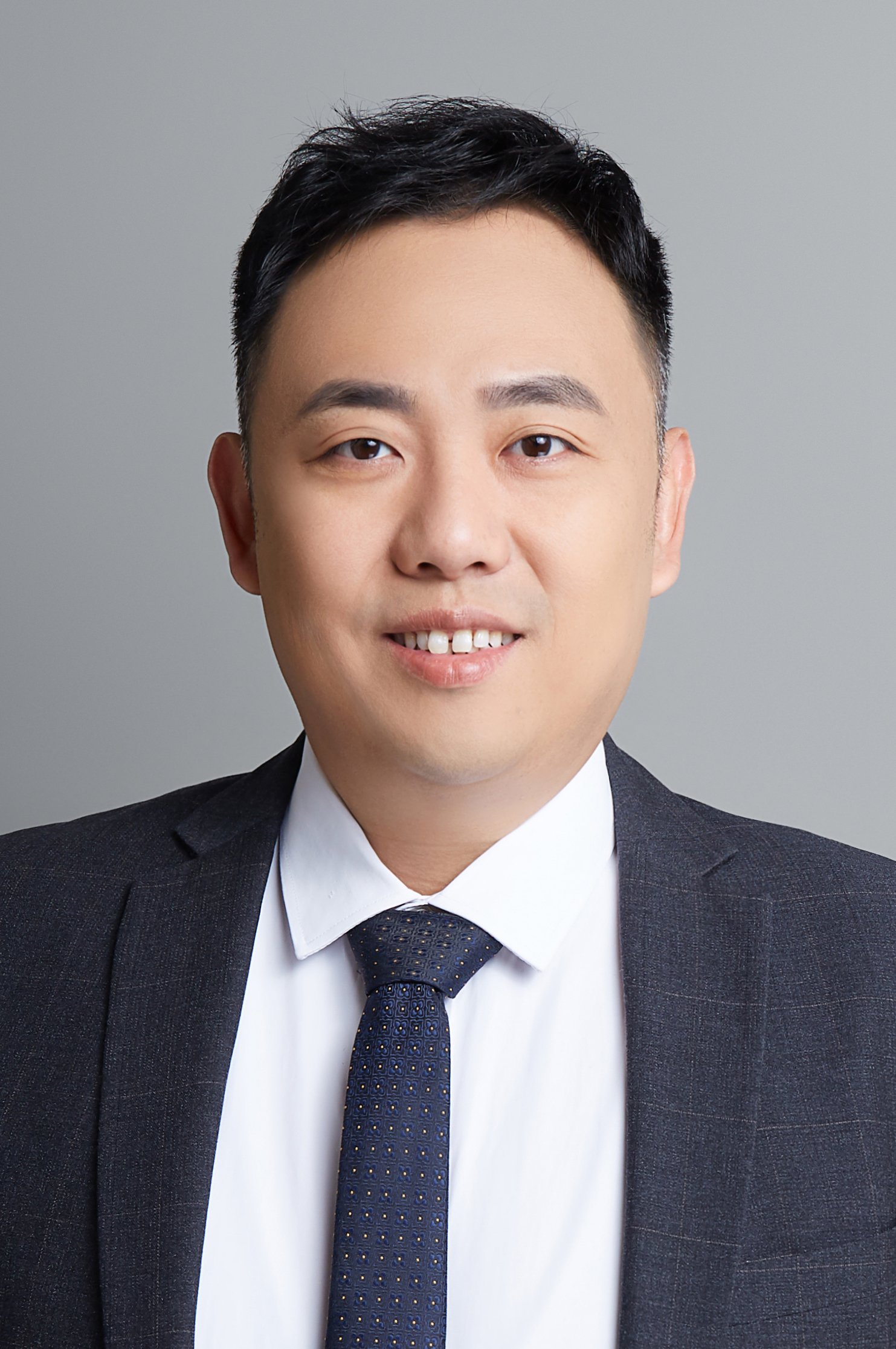}}]{Haiwei Zhang}
received the B.S., M.S. and Ph.D. degrees from Nankai University, Tianjin, China, in 2002, 2005 and 2008, respectively. He is currently an associate professor and master supervisor in the college of Computer Science, Nankai University. His main research interests include graph data, database, data mining and XML data management.
\end{IEEEbiography}

\begin{IEEEbiography}[{\includegraphics[width=1in,height=1.25in,clip,keepaspectratio]{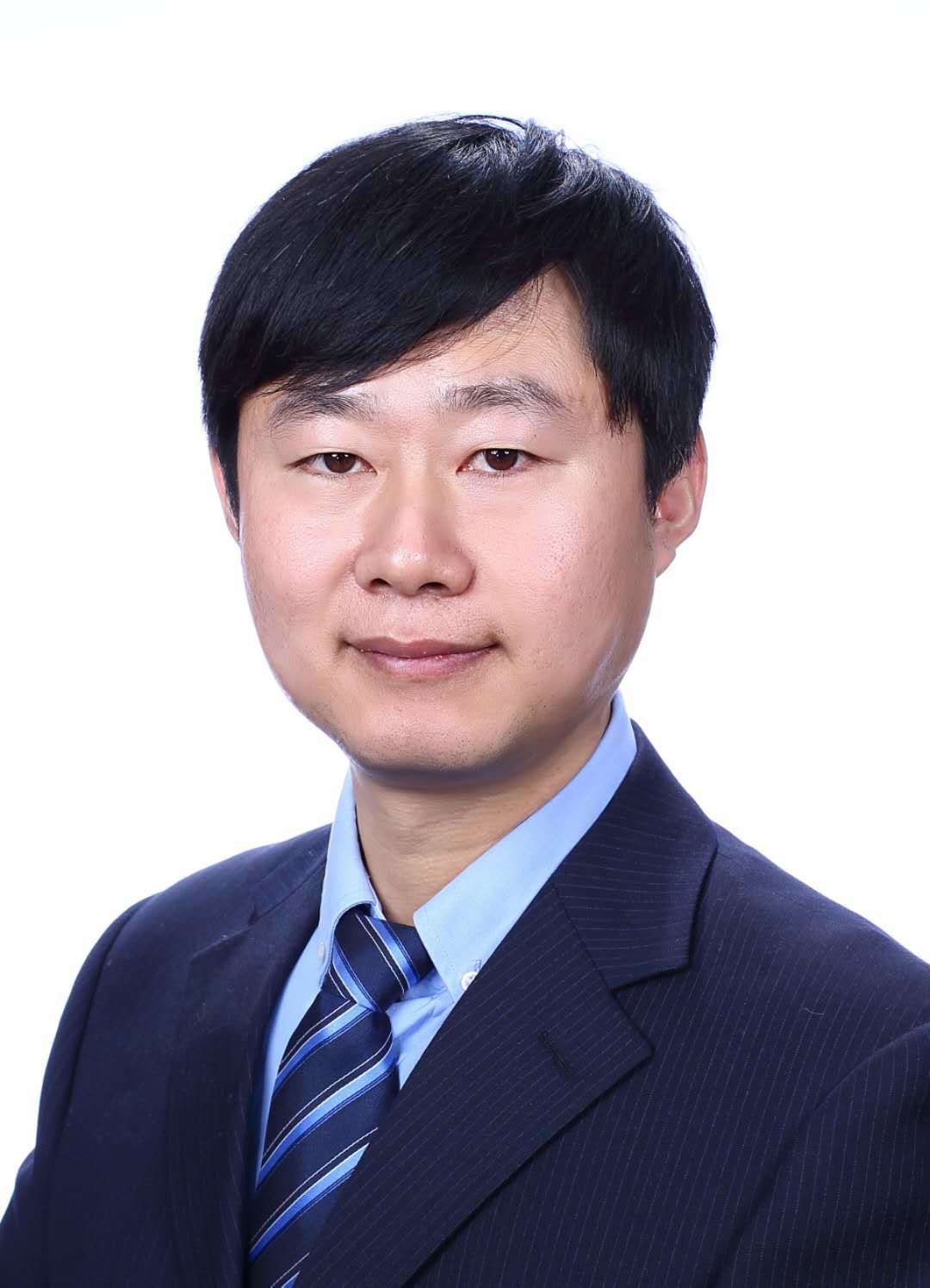}}]{Zhicheng Dou}
(Member, IEEE) received the B.S. and Ph.D. degrees from Nankai University, Tianjin, China, in 2003 and 2008, respectively. He is currently a professor in Gaoling School of Artificial Intelligence, Renmin University of China, Beijing, China. From July 2008 to September 2014, he was with Microsoft Research Asia. His current research interests are information retrieval, natural language processing, and big data analysis. He was the recipient of the Best Paper Runner-Up Award from SIGIR 2013, and Best Paper Award from AIRS 2012. He was the program co-chair of the short paper track for SIGIR 2019. His homepage is \url{http://playbigdata.ruc.edu.cn/dou}.
\end{IEEEbiography}

\begin{IEEEbiography}[{\includegraphics[width=1in,height=1.25in,clip,keepaspectratio]{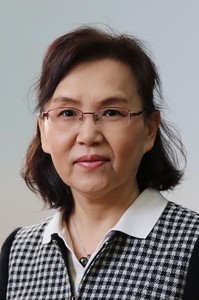}}]{Xiaojie Yuan}
received the B.S., M.S. and Ph.D. degrees from Nankai University, Tianjin, China. She is currently working as a professor of College of Computer Science, Nankai University. She leads a research group working on topics of database, data mining and information retrieval.
\end{IEEEbiography}



% \vfill

\end{document}